\documentclass[magister,archiwum]{helpers/dyplom}
\usepackage[utf8]{inputenc}
\usepackage{hyperref}
\usepackage[noabbrev]{cleveref}
\usepackage[figurename=Figure]{caption}
\usepackage{gensymb}
\usepackage{algorithm}

\usepackage{tabularx}

\usepackage{diagbox}

\usepackage{algpseudocode}
\usepackage[toc]{appendix}

\usepackage{url}
\usepackage{lipsum}
\usepackage{graphicx}
\usepackage{caption}
\usepackage{subcaption}
\usepackage{booktabs}
\usepackage{xurl}

\author{Piotr Gramacki}
\title{Unsupervised embedding and similarity detection of microregions using public transport schedules}
\promotor{dr inż.  Piotr Szymański}
\wydzial{Faculty of Computer Science and Management}
\kierunek{Applied Computer Science}
\specjalnosc{Data Science}
\krotkiestreszczenie{This thesis proposed an unsupervised embedding method for micro-regions in a city based on public transport schedules. It was tested on a wide selection of European cities, which were used to perform an exploratory analysis of an embedding space. Clustering methods were utilized to identify a typology of public transport offer types. The proposed solution found regions with similar public transport availability and allowed to compare public transport quality between cities.}
\slowakluczowe{unsupervised representation learning, spatial data, urban data, public transport schedules, similarity detection}

\begin{document}

\renewcommand*\contentsname{Table of contents}

\maketitle

\pagenumbering{gobble}
\shipout\null

% --- Strona ze streszczeniem i abstraktem ------------------------------------------------------------------
\addtocontents{toc}{\protect\setcounter{tocdepth}{-1}}
\chapter*{Abstract} 

The role of spatial data in tackling city-related tasks has been growing in recent years. 
To use them in machine learning models, it is often necessary to transform them into a vector representation, which has led to the development in the field of spatial data representation learning.
There is also a growing variety of spatial data types for which representation learning methods are proposed.
Public transport timetables have so far not been used in the task of learning representations of regions in a city. 
In this work, a method is developed to embed public transport availability information into vector space.
To conduct experiments on its application, public transport timetables were collected from 48 European cities.
Using the H3 spatial indexing method, they were divided into micro-regions. 
A method was also proposed to identify regions with similar characteristics of public transport offers. 
On its basis, a multi-level typology of public transport offers in the regions was defined. 
This thesis shows that the proposed representation method makes it possible to identify micro-regions with similar public transport characteristics between the cities, and can be used to evaluate the quality of public transport available in a city.

% Kilka sztuczek, żeby:
% - Abstract pojawił się na tej samej stronie co Streszczenie
% - Abstract nie pojawił się w spisie treści
\begingroup
\renewcommand{\cleardoublepage}{}
\renewcommand{\clearpage}{}
\chapter*{Streszczenie} 

Rola danych przestrzennych w rozwiązywaniu zadań związanych z miastem rośnie w ostatnich latach.
Aby możliwe było wykorzystanie ich w modelach uczenia maszynowego często niezbędne jest przekształcenie ich do reprezentacji wektorowej, co spowodowało rozwój dziedziny uczenia reprezentacji danych przestrzennych.
Rośnie również różnorodność typów danych przestrzennych dla których proponowane są metody uczenia reprezentacji. 
Rozkłady jazdy transportu publicznego do tej pory nie były wykorzystywane w zadaniu uczenia reprezentacji regionów w mieście. 
W tej pracy opracowano metodę osadzania informacji o dostępności transportu publicznego w przestrzeni wektorowej.
W celu przeprowadzenia eksperymentów dotyczących jej zastosowania, zebrano informację o rozkładach jazdy w 48 miastach Europy.
Wykorzystując metodę indeksowania przestrzennego H3 podzielono je na mikroregiony.
Zaproponowano również metodę identyfikacji regionów o podobnej charakterystyce oferty transportu publicznego.
Na jej podstawie zdefiniowaną wielopoziomową typologię ofert transportu publicznego w regionach.
W pracy pokazano, że zaproponowana metoda reprezentacji umożliwia zidentyfikowanie regionów o podobnej charakterystyce transportu publicznego pomiędzy miastami oraz może być wykorzystana do oceny jakości dostępnego w mieście transportu publicznego.
\endgroup
\addtocontents{toc}{\protect\setcounter{tocdepth}{2}}
% --- Koniec strony ze streszczeniem i abstraktem -----------------------------------------------------------
\cleardoublepage
\tableofcontents

\pagenumbering{arabic}
\setcounter{page}{0}

\chapter{Introduction}
\label{chapter:intorduction}

Spatial data are available widely for many cities around the world. They are used in solving urban-related tasks such as traffic flow prediction, house price estimation, road network optimization, identification of accident locations, managing transportation issues, and many others. They may also be a valuable source of information in comparing so-called micro-regions in the city. For this, however, it is necessary to have a method of representation of those data. Hence in this section, we present the motivations for using representation learning for spatial data as well as topic analysis with regards to public transport and micro-regions identification. Finally, we formulate the thesis objectives.

\section{Representation learning in spatial data science}

Term representation learning is used to describe a set of techniques for the automated discovery of features from raw data. One of the methods from this area is embedding. It can be defined as a task of finding an optimal function $f: D \rightarrow \rm{I\!R}^{d}$, which transforms data $D$ into $d$-dimensional vector representation.

Learning vector representation of spatial data may increase the performance of models used for solving tasks based on spatial data as was shown in the literature in recent years\cite{cape}\cite{venue2vec}\cite{hrnr}. The main advantage which is gained when translating pure spatial data to latent space is the possibility to define a measure of distance between objects. This makes comparing micro-regions of the city in a certain spatial context possible. Finally, one can perform exploration in such a new space, which may reveal previously unnoticed patterns or features present in the city.

\section{Public transport schedules}

One of the urban types of data which to the best of the author’s knowledge was not deeply explored in the context of various machine learning or data mining tasks is public transport schedules. Even though they are widely available in a unified format for many cities around the world (see \ref{sec:gtfs}). It is also obvious that public transport availability is an important aspect when comparing regions in the city or comparing cities with each other. Hence public transport quality is often one of deciding factors when choosing a district to live in.

Analysis of transport schedules in the sense of their real suitability for city dwellers is not a trivial task. Therefore, providing a method to learn meaningful vector representation of public transport availability for a given micro-region seems to be important for the development of systems with the aforementioned functionality.

Ability to analyze an entire city to e.g find the areas which need an improvement in terms of public transportation would be certainly beneficial for urban planners and city authorities. This may also act as a way to compare current public transport infrastructure between cities. This possibility may be used to compare with cities considered as a "gold standard" in the area of public transport organizations. 

Unfortunately, there is not much research conducted to analyze an existing public transport infrastructure in terms of a static analysis of public transport schedules. Many of them focus mainly on planning optimal routes e.g. between different destinations. If they analyze existing networks they often focus on the prediction of delays or calculation of the estimated time of arrival (ETA). However, exploring similarities between regions of the city or similarities between different cities is still a relatively unexplored area.

\subsection{GTFS format}\label{sec:gtfs}

\begin{figure}[ht]
    \centering
    \includegraphics[width=0.9\textwidth]{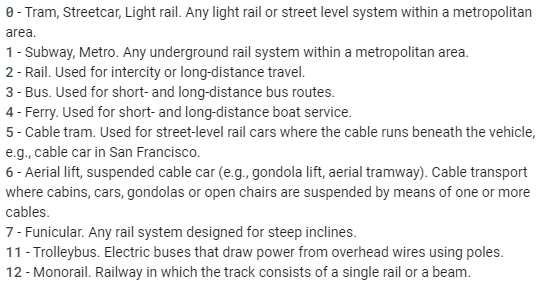}
    \caption{Transportation types valid for GTFS format \cite{gtfs}}
    \label{fig:gtfs-route-types}
\end{figure}

General Transit Feed Specification (GTFS) is a unified format for static public transport schedules introduced by Google \cite{gtfs}. It was first used to incorporate transit data to Google Maps and over the years became de facto a standard format for transit feed data. It contains information about stops, routes, and times or arrival for each stop on a route. In addition to this basic set of information, the GTFS format allows specifying the precise route which the vehicle takes. It can also store additional information about a station such as levels, pathways, or fare information. This format can be used to specify the number of different transportation types, which are presented in Figure \ref{fig:gtfs-route-types}.

The specification of GTFS format consists of obligatory and optional files compressed in a .zip archive. The mandatory files are:

\begin{itemize}
    \item \textit{agency.txt} containing information about agency providing the transportation,
    \item \textit{stops.txt} with the location of stops,
    \item \textit{routes.txt} with description of routes, 
    \item \textit{trips.tx}t describing the trips as a collection of stops on a route,
    \item \textit{stop\_times.txt} which contains times of departure for each stop in a trip,
    \item \textit{calendar.txt} or \textit{calendar\_dates.txt} which specifies the days on which a trip takes place.
\end{itemize}

Those can be expanded with additional data in optional files:

\begin{itemize}
    \item \textit{fare\_attributes.txt} and \textit{fare\_rules.txt} to describe fares which apply for each trip,
    \item \textit{shapes.txt} with roads allignments for vecihle travel paths,
    \item \textit{frequencies.txt} for stops with fixed frequency of departures,
    \item \textit{transfers.txt} with rules for transfer points between routes,
    \item \textit{pathways.txt} with information about pathways between the stops,
    \item \textit{levels.txt} for multi-level stations,
    \item \textit{feed\_info.txt} for additional metadata,
    \item \textit{translations.txt} with agency info translated to other languages
    \item \textit{attributions.txt} with additional attributions for dataset.
\end{itemize}

\subsection{The role of public transport accessibility}

The quality and accessibility of public transport are often one of the main determinants of a standard of living. This can be expressed in different aspects but availability is one of the most important.

The most intuitive definition of public transport quality in a region is the number of different transportation options and a frequency of departures from e.g. the nearest area of residence. Those numbers indicate how easy it is to depart from a given region. Similarly, the variety of accessible destinations is also vital when discussing availability. Those can include either direct routes or routes with a transfer. Such statistics can be created for a single region without considering the whole city. However, the relative availability in comparison with other regions of a city is also important and useful information.

Considering a city as a whole is yet another dimension of public transport analysis. Which parts of a city are well communicated with others? Are there any regions which stand out both in positive and negative aspect? Many such questions can be asked. It is natural that the city center will be more available than suburbs, but isn't the difference too big? Answering such questions may help city authorities to decide about the future development of public transport networks.

Analysis of public transport availability reveals one more aspect - a variety of city types. Circular ones, with the centrally located center, will be different from a line city. This brings up the problem of defining transport availability. It is possibly enough to have a good connection with a city center in a circular city, but in a line city, it is also important to have good connections "in parallel" to the main city line. Cities with two centers should be well connected between two halves as well. 

\begin{figure}[ht]
    \centering
    \includegraphics[width=0.75\textwidth]{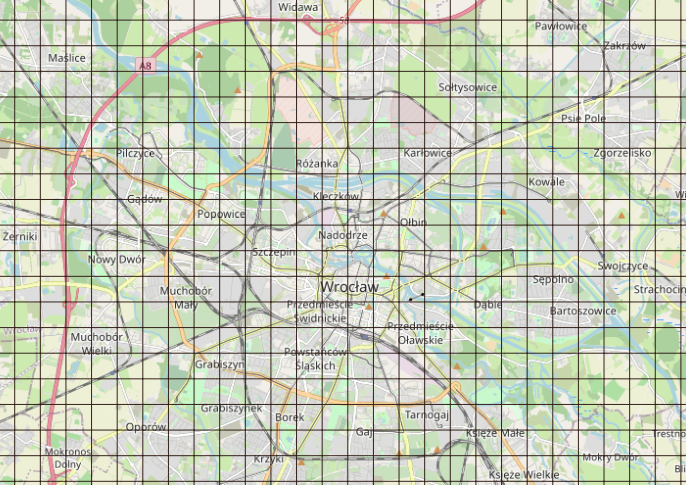}
    \caption{Grid applied on map projection}
    \label{fig:wroclaw-grid}
\end{figure}

\section{Identification of micro regions}

A method of splitting a given area into micro-regions is a crucial aspect of this thesis. There are many ways of achieving this goal. This section will go through the most popular methods and will justify the choice made for this work.

% \subsection{Dividing the map into fragments} 
\subsection{Grid-based division}

The simplest solution is to divide a given area into regular shapes based on a map projection. An example is presented in Figure \ref{fig:wroclaw-grid}. It is simple and intuitive but may result in some inconveniences. Firstly, such division may depend on map scale, and therefore may be inconsistent between different cities. Secondly, it heavily relies on map projection which is inaccurate because the Earth is a sphere that cannot be precisely mapped to a flat surface. Lastly, such division is not easily repeatable as it may depend on the selected area's borders.

\subsection{Manually dividing city}

\begin{figure}[ht]
    \centering
    \includegraphics[width=0.75\textwidth]{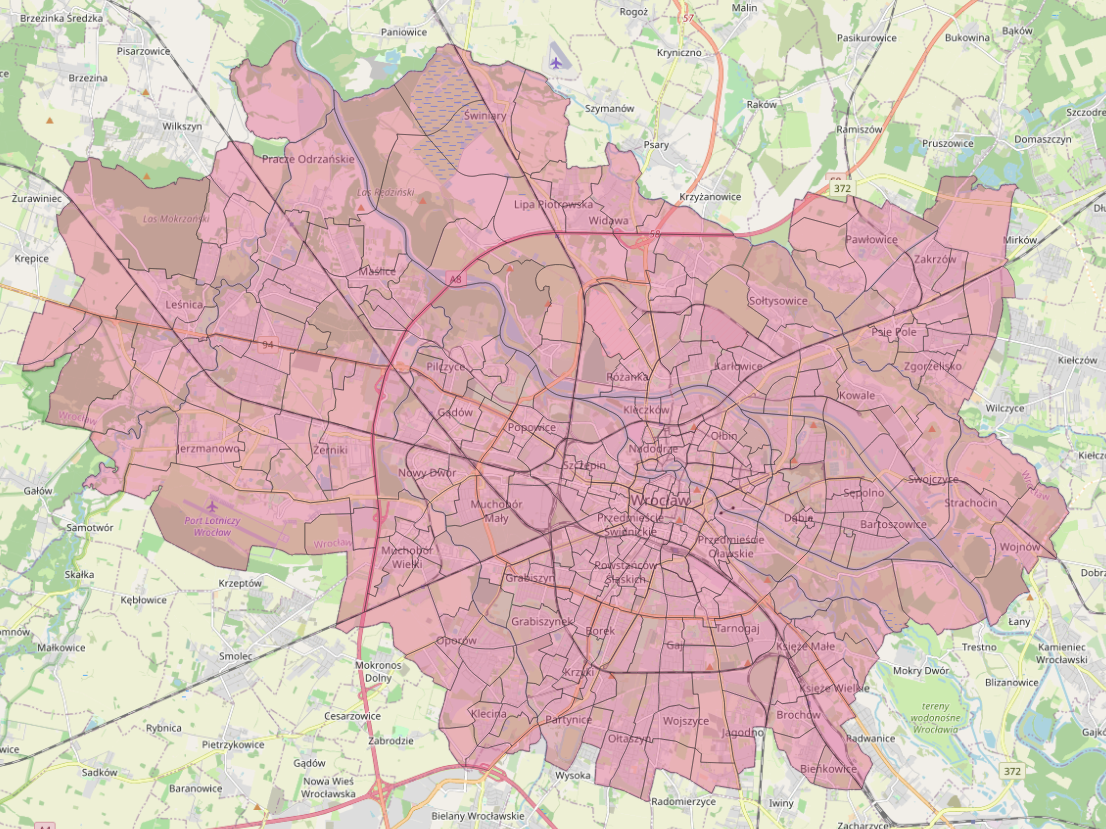}
    \caption{Manual city division created for complex traffic study in Wrocław \cite{kbr2018}}
    \label{fig:wroclaw-kbr}
\end{figure}

Manual pointing of the micro-region is the most accurate solution. It allows to incorporate domain knowledge and creates a division suitable for a given task. An example of such division created for a complex traffic study in Wrocław is presented in Figure \ref{fig:wroclaw-kbr}. It makes it possible to divide a city based on different characteristics e.g. on population density and/or incorporate administrative borders within the city.

An obvious disadvantage of such an approach is that it is labor-intensive and specific to one single area. Therefore it is not suitable for applications that will incorporate different regions of different cities, like the one presented in this thesis. 

\subsection{Spatial indexes}

The spatial index makes it possible to divide a given area into grid cells. Hence each cell is identifiable by its index. Many grid systems work hierarchically, meaning that each cell is composed of smaller cells. Examples of such systems are Google's S2 \cite{s2} based on rectangles and Uber's H3 \cite{h3} which uses hexagons. In this work, the H3 library will be used to partition cities into micro-regions. The reasons behind this decision are presented in the following section.

\subsection{H3 hierarchical spatial index}

\begin{figure}[ht]
    \centering
    \includegraphics[width=0.9\textwidth]{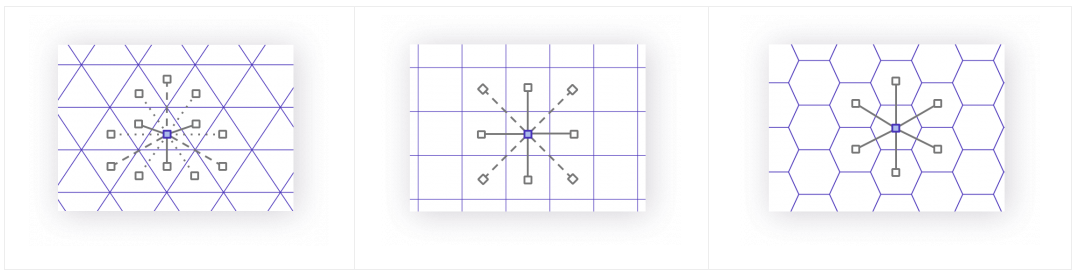}
    \caption{Neighbours and distances for different shapes in space division \cite{h3}}
    \label{fig:h3-rulez}
\end{figure}

Uber's solution uses hexagons which is an effective way of space division. One of its advantage over using rectangles or triangles is that all neighbors are in the same distance. It is illustrated in Figure \ref{fig:h3-rulez}.

\begin{figure}[ht]
    \centering
    \includegraphics[width=0.75\textwidth]{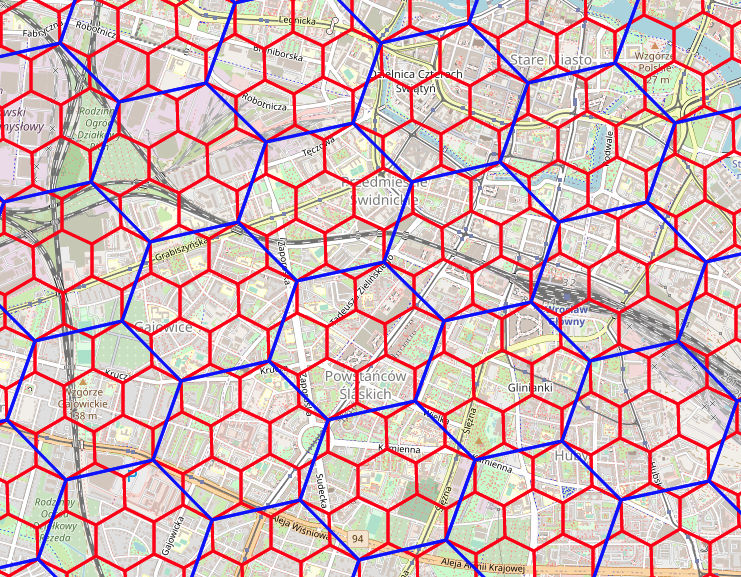}
    \caption{H3 cells - comparison of resolution 8 and 9}
    \label{fig:wroclaw-8-9}
\end{figure}

Another advantage is the hierarchical character of this system. This allows selecting a resolution that is optimal for micro-region identification and makes it possible to consider sub-regions for each micro-region. A visualization of this hierarchical division is presented in Figure \ref{fig:wroclaw-8-9}. 
%Example of city division using H3 spatial index is presented in Figure \ref{fig:wroclaw-h3}.

% \begin{figure}[ht]
%     \centering
%     \includegraphics[width=0.9\textwidth]{images/h3_hierarchical.png}
%     \caption{Hierarchical division of H3 cells \cite{h3}}
%     \label{fig:h3-hierarchy}
% \end{figure}

% \begin{figure}[ht]
%     \centering
%     \includegraphics[width=0.75\textwidth]{images/wroclaw_h3.png}
%     \caption{H3 cells applied to Wrocław at resolution 8}
%     \label{fig:wroclaw-h3}
% \end{figure}

\paragraph{Resolution in H3 spatial index.}

As mentioned above, H3 is a hierarchical index, which is its great advantage. Hierarchy in H3 is implemented as 16 levels of resolution. On level 0 there are 122 base cells covering an entire planet. The smallest hexagons on level 15 have a diameter of length 1m. 

For the task of regions embedding concerning public transport, two seem to be a good fit - 8 and 9. They are respectively about 900m and 350m in diameter. The bigger one covers an area roughly by main roads, whereas resolution 9 is close to a single quarter of buildings. The difference is presented for a small part of Wrocław in Figure \ref{fig:wroclaw-8-9}.

\section{Thesis objective and scope}

The theoretical purpose of this thesis is to propose a method for the unsupervised embedding of micro-regions in the city with the usage of information from public transportation offers. In the acquired latent space, a measure of similarity will be proposed to extract differences in public transportation accessibility between micro-regions. The practical goal of this thesis is to explore this space using data extracted for the city of Wrocław, PL. Once some patterns are found and described in this space, a comparison with different cities will be conducted to find possible similarities between areas across different cities. 

The remainder of this thesis is organized as follows: Chapter 2 contains a review of literature in the areas of representation learning applications for spatial data and public transport analysis, Chapter 3 presents a proposed solution for features selection and embedding model, Chapter 4 reports results of experiments conducted with the exploratory analysis of obtained space, Chapter 5 summarizes the whole thesis and discuses future works.

\chapter{Related works}

This chapter summarizes a literature review that was conducted before starting work on this thesis. It contains a literature review of applications of representation learning in the domain of spatial data. It also pays attention to public transport analysis methods.

\section{Representation learning for spatial data}

This section presents a selection of methods for spatial data embedding. It includes a few selected works with a variety of embedding method inspirations and spatial data types.

\subsection{Loc2Vec (2018)}

Loc2Vec\cite{loc2Vec} proposed a method of spatial data embedding based on the computer vision domain. The goal of this method was to find a metric space in which semantically similar regions are close. Regions were represented by the rasterized data from the OpenStreetMap (OSM). This allowed the processing of unstructured data from OSM using techniques from the image processing domain. To increase the amount of information in a rasterized data, the authors proposed to use a 12-dimensional tensor instead of a 3-dimensional RGB image, where each dimension contains a different type of data (road network, buildings, Points-of-Interest and other).

As an embedding model, the authors used a convolutional neural network trained using a self-supervised approach with a triplet loss. This technique requires two additional samples for each processed image - one positive, to which a distance in embedding space is minimized, and one negative, to which distance is maximized. To obtain a positive samples authors used what is referred to as \textit{Tobler's first law of geography}, which states "Everything is related to everything else, but near things are more related than distant things."\cite{toblers-first-law}. With this assumption authors sampled positive images applying a random rotation and shift them on a map before rasterizing it to a 12-dimensional tensor.

Finally, the authors analyzed such an obtained space and showed that a semantic similarity was reflected in an embedding space by using interpolations, random walks, and vector arithmetic. 

\subsection{CAPE (2018)}

Content-Aware hierarchical POI Embedding (CAPE)\cite{cape} method is inspired by natural language processing. The objective of this work was to propose an embedding method for Points-of-Interest (POI) which was later tested on the task of POI recommendation. The authors constructed a check-in dataset from Instagram, which combined check-in information with textual content. This allowed them to consider not only check-in sequence but also textual information when calculating embedding of POI. 

The authors used a model with two layers to create an embedding: a check-in context layer and a textual context layer. A check-in context layer used POI sequence to train a Skip-gram based model. A fixed number of previously visited POIs served as a context for a target POI. The model was trained using negative sampling. This model captured geographic influence between POIs and was supplemented by a textual context model. It used a word2vec\cite{word2vec} model to learn embeddings of texts from a textual description of check-ins. Here the difference is that an objective was modified to model the probability of a context word conditioned by both target word and associated POI. This layer was also trained using negative sampling. 

The finally proposed model captured both the geographical and textual context of POIs and hence improved POI recommendations.

\subsection{Zone2Vec (2018)}

Zone2Vec\cite{zone2vec} is an embedding method aimed to learn representations of zones in the city. Authors defined a zone as a part of a city separated by main roads. The main source of data used to train an embedding model were trajectories, which were transformed to form a sequence of zones visited on a trip. Additionally, other semantic information was used to define each zone.

As an embedding model, the authors proposed to use a Skip-gram model, first proposed in natural language processing. They treated a whole sequence of zones as a context. Then to utilize semantic information about zones, a doc2vec\cite{doc2vec} method was used, treating each zone as a document, and its semantic information as words. Two resulting embeddings were then concatenated to form a final zone embedding.

The authors tested this solution on both supervised (zone classification) and unsupervised (function discovery) tasks and achieved very promising results.

\subsection{ZE-Mob (2018)}

ZE-Mob model\cite{ze-mob} used representation learning to solve the task of identifying urban functions of zones in the city. A proposed solution was inspired by language processing models. The authors used human mobility from a taxi trips dataset. To identify zones in the city, official zones from US Census Bureau were utilized. 

To calculate zone embeddings, authors utilized pointwise mutual information (PMI) matrix which describes a co-occurrence of zones in mobility patterns. To model differences in attractiveness between zones, authors extended their model with a so-called gravity approach. It is based on a knowledge of how many trips end at a given zone and how long are the trips. 

The authors verified their method against an official land development plan and showed that their method is effective in automatically detecting functional regions in a city. 

\subsection{Venue2Vec (2020)}

Venue2vec\cite{venue2vec} proposed a model for Points-of-Interest (POI) embedding based on intuitions from natural language processing and graph embeddings. Authors used Geo-Social Networks (GSN), which allow users to check in at a given location. Based on such knowledge tasks of next location prediction and future location prediction can be defined. To solve them, the authors proposed a method of representation learning for locations in GSN, which was later used for prediction.

The authors used a DeepWalk method to sample location sequences from the GSN. The random walk probability considers both geographic influences in form of a distance between locations and a sequential relation defined as a frequency of transition between locations. Such sampled sequences were used as sentences to train a Skip-gram model with hierarchical softmax. In addition to embedding learned in this way, a textual embedding of location category (for example \textit{Art \& Entertainment}) is concatenated to form a final location embedding.

With this model, it was shown the improvement of solving both next location and future location prediction tasks.

\subsection{HRNR (2020)}

Hierarchical Road Network Representation model (HRNR)\cite{hrnr} tackled the task of learning meaningful representations of segments in a road network. The authors proposed to represent a road network at three levels: 
\begin{itemize}
    \item road network nodes, which represent uniform segments of a road,
    \item structural regions, which is a set of spatially connected road segments,
    \item functional regions, which consist of structural regions providing certain traffic functionality.
\end{itemize}
Then, they used a combination of Graph Attention Networks (GAT) and Graph Convolutional Networks (GCN) to learn adjacency matrices between levels and vector representations of roads and regions. For functional regions, the authors proposed to utilize trajectories data to extract functional relations inside a road network. 

Authors showed, that using such a hierarchical model for embedding road networks good results can be obtained in a variety of target tasks on multiple real-world datasets.

\subsection{Spatial embeddings summary}

In the process of literature review, a wider selection of methods related to spatial data embeddings was collected. Based on them, a Wikipedia entry on the topic of \textit{Spatial embedding}\cite{wiki:spatial-embeddings} was created in cooperation with other students. It successfully passed a review process and is available on Wikipedia. 

\section{Public transport analysis}

The previous section presented an overview of some selected methods of representation learning in the domain of spatial data. This section will focus on tasks and methods particularly associated with public transport data. I will focus on identifying how this type of data can be used and what methods are utilized. 

\subsection{Delays analysis and prediction}

One of the most commonly tackled tasks related to public transport is delay prediction. Two of the most recent methods are presented in \cite{Shoman2020} and \cite{Barnes2020}. They both utilize deep learning and embeddings to predict delays recorded in real-time data based on public transport schedules. 

The method presented in \cite{Shoman2020} used TabNet model\cite{tabnet} to embed tabular data from GTFS files. Those embeddings were combined with historical data about delays and information about traffic from another data source. The authors showed that their method can work on an entire public transport network and they achieved low error in delay estimation.

Authors of BusTr\cite{Barnes2020} proposed a method for real-time estimation of bus delays. In contrast to other solutions, they inferred bus delays from real-time traffic and historical context for each segment in a public transport network. According to the authors, this allows delay prediction even in areas where real-time public transport data is not available. In their prediction model, they utilized embeddings of time and routes, followed by embeddings of segments of roads and stops. For location embeddings, they used the S2 spatial index\cite{s2}. Current traffic information was obtained from Google Maps. The authors state, that their model outperforms the best alternative for delay prediction and demonstrates very good generalization.

Another task related to public transport delays is an analysis of types and reasons for delays. An example can be found in \cite{Raghothama2016}. The authors collected information about real-time public transport vehicle positions for three months and acquired around 1.5M delays of public transport vehicles. They combined it with information about road infrastructure such as speed limits or the number of lanes. Finally, they applied clustering to identify regions with similar types of delays and their causes. 

\subsection{Public transport network analysis}

As presented above, delay prediction is a frequent task related to public transport. However, the goal of this thesis is to acquire a typology of public transport offers types, which is more of an analytical task related to public transport. 

One of the first examples of an analysis of public transport networks was found in \cite{VonFerber2008}. The authors tried to solve a problem from the perspective of network science. They gathered public transport networks from 14 cities around the world with different characteristics, and for each created their network representation. Then they applied network metrics and network modeling approach to analyze such networks and find similarities between them. 

A different approach to public transport network analysis was presented in \cite{FayyazS.2017}. The authors performed an analysis in terms of the accessibility of public transport. They proposed a method of efficient time travel calculation between stops in the city and performed an analysis of public transport availability based on weighted average time travel (WATT) for each stop. They utilized demographic data - region population - as a weight in this estimation. The authors stated that their method is more effective than others cited in their work and that their method is effective in identifying locations with public transport demand. However, despite providing a significantly more efficient method for time travel calculation, the proposed method still requires a lot of computational time. 

\section{Summary}

This section will summarize a literature review conducted for this thesis. The main observation is the fact that a representation learning approach has been widely applied to spatial data in recent years. Some of the analyzed works focus on proposing an embedding method which is then tested on a variety of tasks \cite{zone2vec},\cite{venue2vec},\cite{hrnr}, while other attempt to solve given tasks and for this purpose they apply representation learning techniques \cite{loc2Vec},\cite{cape},\cite{ze-mob}. They show an improvement, when compared to other methods used for solving such tasks \cite{cape},\cite{zone2vec},\cite{ze-mob},\cite{venue2vec},\cite{hrnr}. Moreover, the obtained embedding spaces are often shown to be interpretable \cite{loc2Vec},\cite{venue2vec}.

Embedding methods for spatial data examined in this review draw inspiration from other areas where the representation learning approach is widely used. The majority of them is based on natural language processing and word2vec\cite{word2vec} method \cite{cape},\cite{zone2vec},\cite{ze-mob}. Others used an existing methods for graph embedding \cite{venue2vec},\cite{hrnr} or methods from the domain of image processing\cite{loc2Vec}. This variety shows that there is a visible and strong trend for applying representation learning for spatial data!

Public transport is picked up in the literature mostly in tasks related to general delay estimation. Most recent methods apply embeddings techniques to improve the quality of delays estimating in public transport\cite{Shoman2020},\cite{Barnes2020}. Delays were also used as a factor when comparing regions around a city\cite{Raghothama2016}. 

The task of analyzing and comparing regions in a city in terms of public transport availability is not covered widely in the literature. Methods from the domain of network science were successfully applied to this task\cite{VonFerber2008}, however, this approach cannot be easily transferred to micro-regions comparison. Utilization of travel times\cite{FayyazS.2017} brings valuable insight into the subject of public transport availability and can be transferred to use micro-regions in place of stops. However, the computation costs of this approach are significant, which opens an opportunity for proposing another method that could be easier to apply to a large number of cities. 

\chapter{Proposed solution}

This chapter provides a detailed view of the proposed solution to the problem of public transport accessibility representation and comparison between cities. It covers a resolution selection of an H3 index which is then used for micro-regions identification. Later it describes a feature engineering process and normalization strategy. After that, the architecture and learning process of an embedding model for micro-regions representations learning is discussed in detail. Finally, methods for the analysis of obtained space are presented.

\section{H3 resolution for micro regions selection}

\begin{figure}[h]
     \centering
     \begin{subfigure}{0.49\textwidth}
         \centering
         \includegraphics[width=\textwidth]{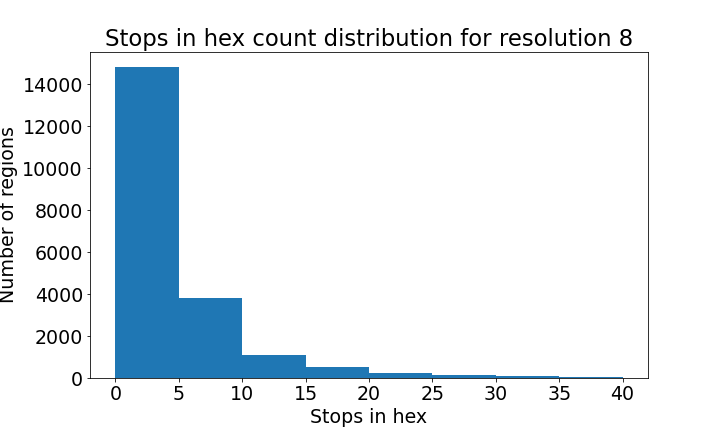}
         \caption{Resolution 8}
     \end{subfigure}
    %  \hfill
     \begin{subfigure}{0.49\textwidth}
         \centering
         \includegraphics[width=\textwidth]{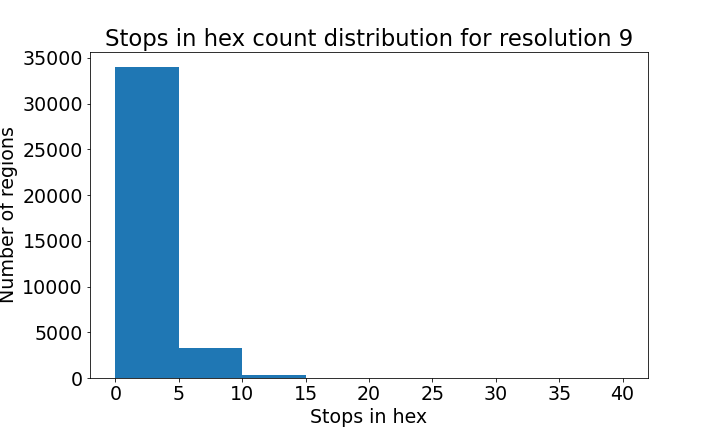}
         \caption{Resolution 9}
     \end{subfigure}
    \caption{Stops count histogram for different H3 resolutions}
    \label{fig:stops-count-histograms}
\end{figure}

As mentioned in Chapter \ref{chapter:intorduction}, H3 spatial index will be used to divide a city into micro-regions. Initially, two resolutions were selected - 8 and 9. To compare them, histograms of the number of bus/tram stops in a hex were drawn for each resolution. Resulting plots are presented on Figure \ref{fig:stops-count-histograms}. A resolution 8 provides more variety in stops number in regions, which seems to be suitable in identifying hubs that consist of multiple stops in close proximity. Furthermore, a number of regions without any stop in it was calculated for both resolutions. Results are presented in Table \ref{tab:empty-regions}. Clearly, at resolution 9 the portion of empty regions raises dramatically. Based on those comparisons resolution 8 was selected for further experiments.

\begin{table}[h]
\centering
\caption{Comparison of regions without any stop for different resolutions}
\label{tab:empty-regions}
    \begin{tabular}{@{}rrrr@{}}
    \toprule
    H3 resolution & Sum of regions & Regions without a stop & \% of empty regions \\ \midrule
    8             & 6478           & 2104                   & 32.48               \\
    9             & 45412          & 33063                  & 72.81               \\ \bottomrule
    \end{tabular}
\end{table}

\section{Features engineering}

As presented in the first chapter, the GTFS format consists of multiple files and is organized around a list of stop times for routes and stops. Such format is impossible to be used directly in any ML/DL model. Therefore, some feature engineering work was necessary to prepare data in the correct format. This section describes this process and justifies the applied solutions. 

From a perspective of a single micro-region, two main public transport accessibility metrics can be defined. Quantity and variety of public transport available. Both were included in the proposed solution.

\subsection{Quantity related features}

To describe the quantity of public transport available for a given micro-region, the sum of trips in a given period was calculated. To perform those calculations the \textit{gtfs-kit} library\cite{gtfs-kit} was used. It provides methods for the calculation of trip counts for each stop at a given time-frequency. Those numbers were later summed up across an entire micro-region. 

Sometimes multiple stops, which form the same route, are aggregated into a single micro-region. As a result, the same trip will be counted multiple times. However, this is an intentional assumption, because those features are not meant to have a meaning of counting vehicles leaving a micro-region. Instead, they describe in how many ways one can leave an area. The intuition behind this is that having more options to get on a bus/tram inside a region is better than just a single stop in the middle of an area.

\subsection{Variety related features}

Another feature which can describe public transport availability is the variety of directions in which one can leave a region. It can be considered on different levels of precision starting from world directions and ending on all regions accessible via public transport with a given number of transfers. For this thesis, an in-between solution was selected. It is described below.

The part of a route description in GTFS format is a \textit{trip\_headsign} column. In practice, this is the destination of a route displayed on a vehicle. They are in most cases either transportation hubs or line ends located most often on the outer ring of the city. This is in a  balance between the generality of world directions and the specificity of each region connected via public transport. Hence, to create features for a region representation, unique head-signs were counted in a given period, resulting in the total number of unique directions available from a region.

\subsection{Time resolution for features and filtering}

Both of the features mentioned above can be calculated for different time windows resulting in a different number of features. For this thesis, one hour window was selected resulting in 24 quantity and 24 variety features for each micro-region. This resolution seems to be a reasonable choice because it presents changes throughout the day and allows to interpret periods for example as rush hours or nighttime. 

After an analysis of timetables for night lines, it was decided to remove them from features of a region. Nighttime transportation availability is significantly different from daytime, as the routes are often longer to minimize the number of vehicles and drivers needed. Moreover, some additional hubs were identified (for example Petrusiewicza street in Wrocław) which are not active during the day. Therefore, only 6-22 hours were included as features for each region. As a result, each micro-region is described with a vector $\mathbf{x}$ of length 34.

Values of quantity- and variety-related feature for each hour is treated as an independent feature, hence a feature vector of length 34. The reason for that being, public transport availability at each hour is treated differently, because of the phenomenon of rush hours. Therefore, \textit{trips\_at\_12} has completely different meaning than \textit{trips\_at\_8}, which describes a morning rush hours. 

\section{Normalization}

\begin{figure}[h]
     \centering
     \begin{subfigure}{0.49\textwidth}
         \centering
         \includegraphics[width=\textwidth]{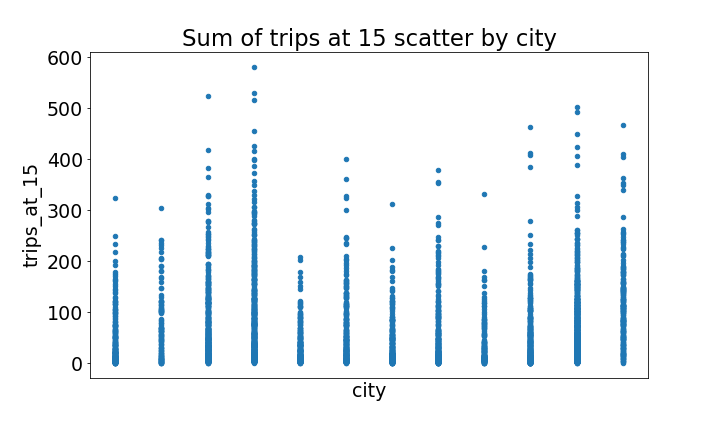}
         \caption{Trips}
     \end{subfigure}
    %  \hfill
     \begin{subfigure}{0.49\textwidth}
         \centering
         \includegraphics[width=\textwidth]{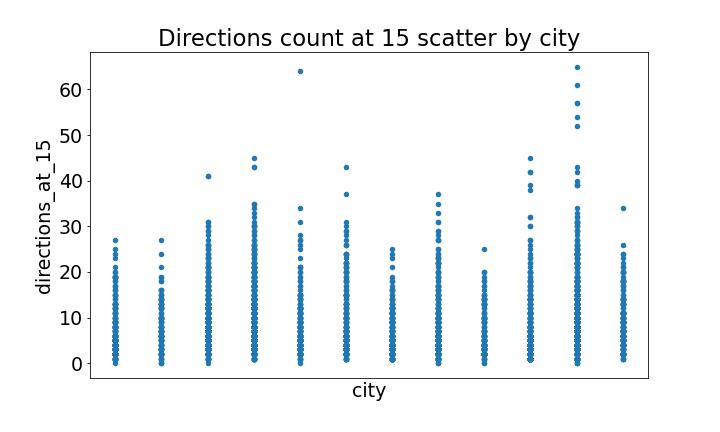}
         \caption{Directions}
     \end{subfigure}
    \caption{Values for sum of trips and directions at 15 hour for different cities}
    \label{fig:trips-directions-scatters}
\end{figure}

The main purpose of this thesis was to propose a method of learning representation of public transport accessibility described by features mentioned in the previous section. To achieve that, the neural network will be used. It will be described in the following section. Here we must note that the above-mentioned features which are the main data in our task should be normalized to be able to serve as an input to the neural network. Figure \ref{fig:trips-directions-scatters} shows a comparison of values for both quantity and variety features (respectively, a sum of trips and destinations at a given hour). The plot was prepared for a selected period during rush hours, which is 15-16. Note that a sum of trips takes the values 10 times bigger than directions count. This may result in a neural network focusing mostly on features with higher values. Another observation is that values vary for different cities. This was also taken into consideration when proposing a method of normalization.

\subsection{Global approach}\label{sec:global-norm}

To achieve the best performance of the embedding model, features will be scaled to the $[0, 1]$ range using \textit{min-max normalization}, which is calculated according to the formula

\begin{equation}
    \mathbf{x}' = \frac{\mathbf{x} - \min(\mathbf{x})}{\max(\mathbf{x}) - \min(\mathbf{x})}.
\end{equation}

As values of each type (quantity and variety) should maintain relative values during the whole day, normalization will be performed twice: once for all columns with a sum of trips and once for all columns with destinations count.

Normalization, which will be referred to as global in the remainder of this thesis, will be conducted for all cities combined. This will ensure that differences in public transport available between cities are maintained. This approach should allow comparing the quality of public transport between cities. 

\subsection{Local approach}\label{sec:local-norm}

The global normalization, which was described above, will be useful in analysis which are meant to incorporate differences in sizes of cities and fleet capabilities. However, in the task of identifying regions with similar role in different cities, one may want to discard those variations. To accommodate for that, another method of normalization is proposed, given by equation:

\begin{equation}
    \mathbf{x}_{city}' = \frac{\mathbf{x}_{city} - \min(\mathbf{x}_{city})}
    {\max(\mathbf{x}_{city}) - \min(\mathbf{x}_{city})},
\end{equation}

where $\mathbf{x}_{city}$ means values for all regions from a single city. 

\begin{figure}[h]
     \centering
     \begin{subfigure}{0.49\textwidth}
         \centering
         \includegraphics[width=\textwidth]{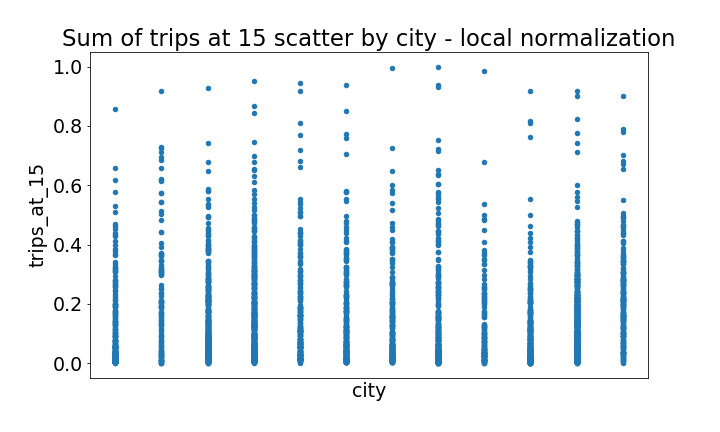}
         \caption{Trips - local normalization}
     \end{subfigure}
    %  \hfill
     \begin{subfigure}{0.49\textwidth}
         \centering
         \includegraphics[width=\textwidth]{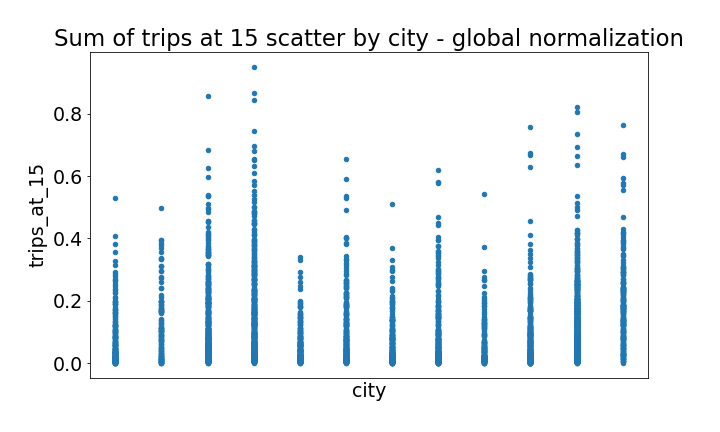}
         \caption{Trips - global normalization}
     \end{subfigure}
     
     \begin{subfigure}{0.49\textwidth}
         \centering
         \includegraphics[width=\textwidth]{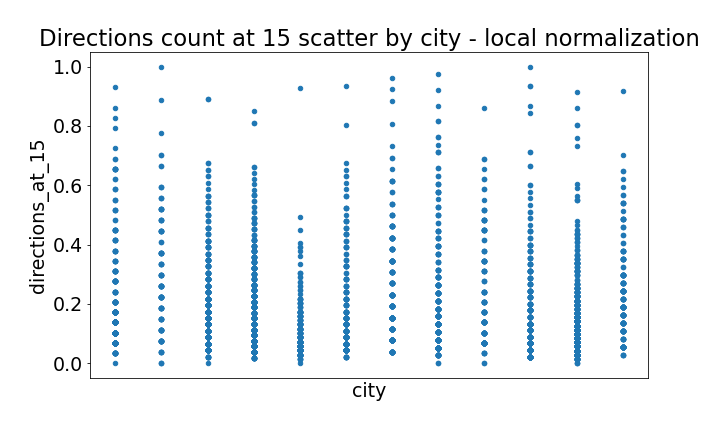}
         \caption{Directions - local normalization}
     \end{subfigure}
    %  \hfill
     \begin{subfigure}{0.49\textwidth}
         \centering
         \includegraphics[width=\textwidth]{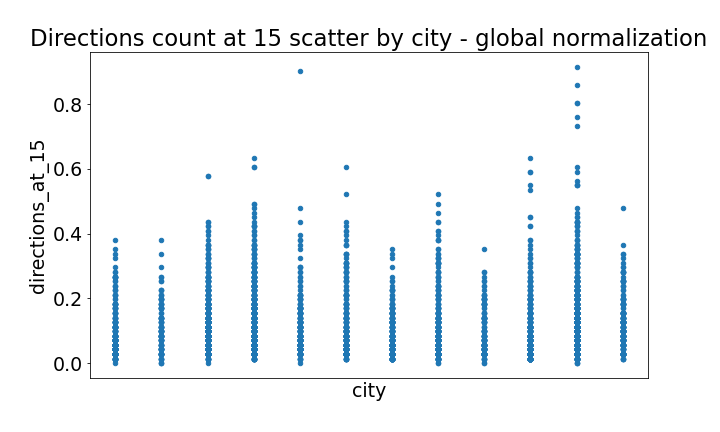}
         \caption{Directions - global normalization}
     \end{subfigure}
    \caption{Comparison of local and global normalization for all cities}
    \label{fig:trips-directions-scatters-normalized}
\end{figure}

The difference between those methods is presented in Figure \ref{fig:trips-directions-scatters-normalized}. As expected, the global normalization kept relative differences between cities, while the local approach removed this variation. 

\section{Embedding method}

This section presents a detailed insight into a model used to learn representations of cities' micro-regions. It will also discuss a neural network architecture used in the experimental section of the thesis. 

\subsection{Autoencoder}\label{sec:autoencoder}

\begin{figure}[ht]
    \centering
    \includegraphics[width=0.75\textwidth]{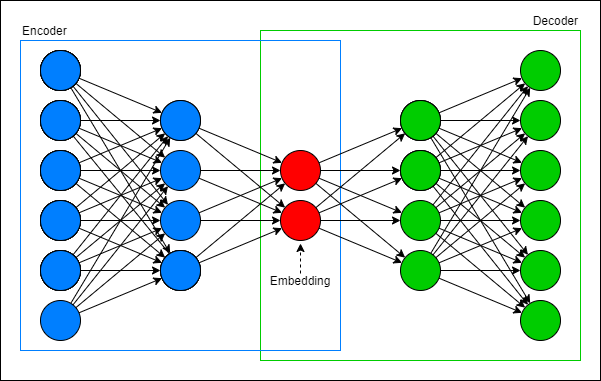}
    \caption{Autoencoder - illustrative block diagram}
    \label{fig:autoencoder-schema}
\end{figure}

Autoencoder is a special type of neural network capable of learning lower- or higher- dimensional embeddings of data. It is trained in an unsupervised manner, which is crucial for this thesis. An overview of an autoencoder architecture is presented in Figure \ref{fig:autoencoder-schema}.

Autoencoder network consists of two networks trained together - Encoder and Decoder. The encoder network can be treated as a function $\phi: \mathcal{X} \rightarrow \mathcal{Z}$, and decoder as $\psi: \mathcal{Z} \rightarrow \mathcal{X}$. Having input vector $\mathbf{x} \in \mathcal{X}$, encoder network produces an embedding $\phi(\mathbf{x}) = \mathbf{z}$, which is then fed into decoder network to obtain reconstructed input $\psi(\mathbf{z}) = \mathbf{x'}$. Both networks are trained together using backpropagation algorithm, with the loss function for a batch of inputs $\mathbf{X}$ defined as:

\begin{equation}
    \mathcal{L} = \frac{1}{N} \sum_{n=1}^{N} (\mathbf{X}_n - \mathbf{X'}_n)^2,
\end{equation}

which is a \textit{mean squared error loss}, also referred as \textit{mse\_loss}. When the training is complete, only the encoder is used for inference.

% \subsection{Dense representation}

Historically first applications of autoencoders focused on compressing an input and discovering nonlinear correlations in data can be found in \cite{Kramer1991}. It is possible as a result of using nonlinear activation functions, and smaller dimensionality of space $\mathcal{Z}$. In this thesis, a 16-dimensional representation is obtained using an autoencoder with an architecture presented below.

\begin{itemize}
    \item Encoder:
    
    \begin{itemize}
        \item $34 \rightarrow 24$ neurons fully connected layer,
        \item ReLU activation function,
        \item $24 \rightarrow 16$ neurons fully connected layer;
    \end{itemize}
    
    \item Decoder:
    
    \begin{itemize}
        \item $16 \rightarrow 24$ neurons fully connected layer,
        \item ReLU activation function,
        \item $24 \rightarrow 34$ neurons fully connected layer;
    \end{itemize}
    
\end{itemize}

This particular architecture was selected somewhat arbitrarily because in a representation learning domain the goal is to extract representations from raw data and only at the stage of exploratory analysis it is possible to determine whether they are good or not. However, several intuitions are backing up this choice of architecture. To begin with, the amount of data, which is available for training is not big. The reason for that is the limited availability of timetables data. A city is divided into approximately 300-500 micro-regions, which means that there is not enough data to train a very deep model. Moreover, as 34 features are representing each region, an embedding dimension should be smaller. Otherwise, it will not extract higher-level features from data and therefore will just \textit{remember} all examples. Finally, the selection of too few dimensions could mean that it will less effective when defining more types in a typology. Having considered that, a 16-dimensional embedding layer and one hidden layer seem like a reasonable choice.

\section{Clustering}

Having an embedding model trained and embeddings calculated for all micro-regions, the next step was to analyze the results and finally try to extract valuable information while comparing the cities. This step will use unsupervised methods like distance calculations in embedding space later used in a clustering task. Since the author has good knowledge of how public transport is organized in the city of Wrocław, this city will be primarily used to verify the correctness of obtained results. As the task solved in this thesis is purely unsupervised in nature, there is no other way to evaluate results other than analyzing obtained space by an expert. 

\subsection{Distance metrics}

In general clustering methods require a distance metric to be defined in an obtained embedding space. Following a common practice in the literature, an euclidean distance (Eq \ref{eq:euclidean-distance}) and cosine distance (Eq \ref{eq:cosine-diatance}) will be used in representations analysis. 

\begin{equation}
    \label{eq:euclidean-distance}
    d_{e}(\mathbf{x}, \mathbf{y}) = || \mathbf{x} - \mathbf{y} ||_2
\end{equation}

\begin{equation}
    \label{eq:cosine-diatance}
    d_c(\mathbf{x}, \mathbf{y}) = 1 - \frac{ \mathbf{x} \cdot \mathbf{y} }{ ||\mathbf{x}||_2 ||\mathbf{y}||_2 }
\end{equation}

\subsection{Hierarchical approach}

The last element of the proposed solution for exploratory analysis performed on a given embedding space is clustering. One of the main goals of this analysis is to propose a multi-level classification in terms of public transport accessibility and its quality. This was a reason for choosing a hierarchical clustering approach in a form of \textit{agglomerative clustering}. 

This is an iterative method, which starts by placing all samples in separate clusters. Then, at each step, two nearest clusters are merged, until a given number of clusters is obtained or another stop criterion is met. To select clusters to merge, multiple methods were proposed and in this thesis, the following two will be used:

\begin{itemize}
    \item \textit{ward}
    
    Clusters are merged to minimize the in-cluster variance. This method is based on a euclidean distance \cite{ward}.
    
    \item \textit{average}
    
    Clusters are merged based on average distance between elements in clusters. The distance between clusters is defined as follows:
    
    \begin{equation}
        d_{avg}(u, v) = \sum_{ij} \frac{d_c(u_i, v_j)}{|u| * |v|},
    \end{equation}
    
    where $|u|, |v|$ means cluster size also known as cardinality.
\end{itemize}

\section{Exploratory analysis and typology identification}

A final goal of an analysis is to receive a typology of public transport offer types across multiple cities. 
% A final goal of an analysis is to detect differences in the transport offer in different cities.

To achieve that, a clustering method described above will be launched multiple times, increasing the number of clusters at each step. The results (clusters) thus obtained at every step will be analyzed to determine the number of clusters that form a kind of an interpretable typology. The desired outcome would consist of multiple levels of typology, which can be characterized and later used for comparisons between cities. The last step will be to describe an obtained typology.
% In the last step, crucial from a practical point of view, a qualitative description of the obtained typology will be made. 

\chapter{Experiments}

This chapter presents the results of experiments, which were conducted as a part of this thesis. A process of typology identification is described, which includes normalization and selection of clustering methods. Then, experiments conducted in this thesis are described with the formulation of research questions and a description of the dataset. Finally, obtained typology is described and further analyzed.

The experiments code was implemented in Python using libraries \textit{scikit-learn}\cite{scikit-learn}, \textit{PyTorch}\cite{pytorch} and \textit{PyTorch Lightning}\cite{pytorch-lightning}.

\section{Typology identification methodology}\label{sec:typology-identification-methodology}

The goal of this thesis is to identify a typology of types of public transport offers. To achieve that, all analyzed cities were divided into hexagons using Uber's H3 spatial index\cite{h3}, which represent regions in a city. Each region was then embedded using the autoencoder described in \ref{sec:autoencoder}. This section describes exploratory analysis done on the obtained embedding space in order to search typology of types of public transport offers.

\subsection{Selection of normalization method}

To obtain a typology that is universal for all cities used in this research, a global method of normalization was used as described in \ref{sec:global-norm}. This method preserves the relative differences between cities, which is essential for this typology, as its goal is to maintain differences between cities at various stages of development. 

A local approach to normalization (see \ref{sec:local-norm}) may be useful when defining a typology of functions of regions in the city. This idea was described in detail in \ref{sec:future}, but was not explored in this thesis, since defining and analyzing another typology is out of scope for this work. 

\subsection{Method for grouping regions}

To group similar regions a clustering was utilized. Since an explainable typology is the main goal, an agglomerative clustering approach was selected. This enabled visualization in the form of a dendrogram which helped to understand splits.
Furthermore, during stepping up with a number of clusters only one of the existing clusters is divided. This provides a better understanding of the clustering process. 

\begin{figure}[h]
     \centering
     \begin{subfigure}{0.49\textwidth}
         \centering
         \includegraphics[width=\textwidth]{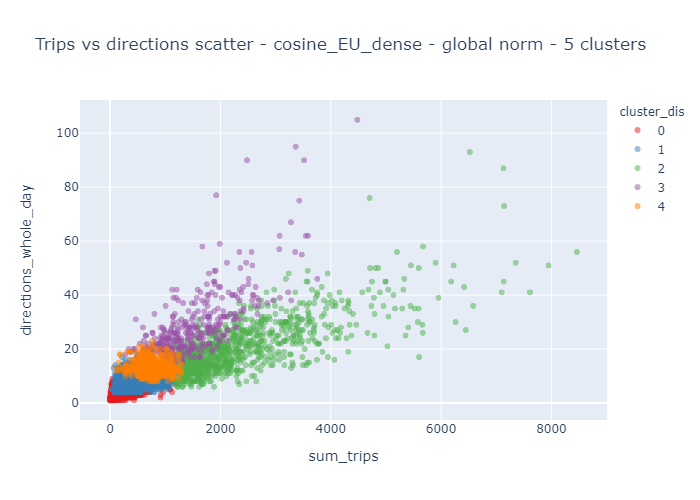}
         \caption{Cosine}
     \end{subfigure}
    %  \hfill
     \begin{subfigure}{0.49\textwidth}
         \centering
         \includegraphics[width=\textwidth]{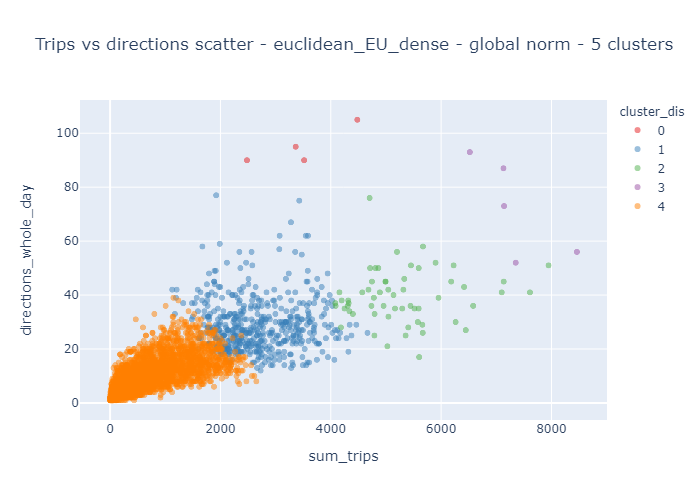}
         \caption{Euclidean}
     \end{subfigure}
    \caption{Clustering using average linkage with cosine and euclidean metric}
    \label{fig:metrics-comparison}
\end{figure}

As a distance metric, cosine and euclidean metrics were considered. A cosine metric finds similarities in \textit{direction} from the origin, which would discard quantity differences between values and focus on a general trend. This would work great with a local normalization-based solution, which aims to find similarities between regions in terms of public transport usage type because it discards differences for example in the number of trips. However, in this thesis, a global approach is utilized, which means that those differences are important. Therefore, a euclidean distance is used, since it extracts actual distances between samples. Those differences between methods are noticeable in Figure \ref{fig:metrics-comparison}. The plots are based on aggregated features, which are discussed in detail in \ref{sec:aggregated-scatter-plots}.

\begin{figure}[h]
    \centering
    \includegraphics[width=0.49\textwidth]{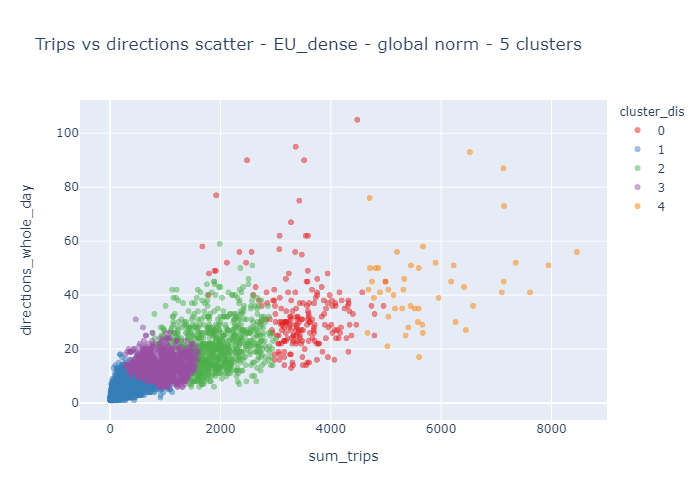}
    \caption{Clustering using Ward's method}
    \label{fig:metrics-comparison-ward}
\end{figure}

Agglomerative clustering requires a method for selecting clusters to combine, also known as the linkage method. In this thesis, Ward's method\cite{ward} was used, because it minimizes intra-cluster variance. This feature is beneficial when identifying a well-differentiated typology. An example of how Ward's method works are presented in Figure \ref{fig:metrics-comparison-ward}. It provides a good differentiation on regions with both low and high volume of public transport.

\subsection{Clustering evaluation}\label{sec:aggregated-scatter-plots}

To evaluate clustering results concerning the goal of defining a differentiated typology several methods of cluster analysis were utilized. For example, a dendrogram for clustering on different levels of hierarchy may be useful for understanding splits. Such a plot is independent of the number of clusters and therefore is a good reference when changing the number of clusters in an agglomerative clustering approach.

More methods of analysis are introduced for particular clustering variants. Here we describe one. Evaluation for a given number of clusters uses two aggregating features:
\begin{itemize}
    \item \textit{sum\_trips} which represent the sum of trips from an entire day (within 6-22 limits),
    \item \textit{directrions\_whole\_day} which represent all directions available during a day (from 6 to 22).
\end{itemize}
They represent public transport availability in a region from a whole day perspective. This is a good measure to justify differences between clusters from a distance. Those features will be used to prepare scatter plots like the ones in Figures \ref{fig:metrics-comparison} and \ref{fig:metrics-comparison-ward}. 

Scatter plots of aggregated features as described above may be insufficient with the increased number of clusters. In this case, to find differences in per-hour features, values of all features for samples from a single cluster will be presented in the form of a boxplot. This will ensure, that discrepancy on for example rush hours will be taken into consideration when proposing a typology. 

\subsection{Share of particular clusters in each city}

To provide more insight on clusters meaning, a share of all clusters for each city will be calculated. This will allow, to analyze a clustering quality, since it will ensure that clusters are inter-city. Moreover, this will be useful when analyzing differences between cities and their public transport availability.

\subsection{Visual evaluation with maps}

As the last stage of clustering analysis, clusters will be visualized on a map, to allow visual assessment of correctness. Since analyzing all twelve cities may be challenging, only four were selected for this process, based on their characteristics. Selected cities are:

\begin{itemize}
    \item Berlin - the biggest city available with good quality of public transport.
    \item Wroclaw - a city of medium size, with a single established center and circular layout.
    \item Tricity (Gdansk, Gdynia, Sopot) - three cities with shared public transport. The whole metropolis forms a line city.
    \item Barcelona - a grid-based city with good quality public transport.
\end{itemize}

\section{Experiment 1: 12 cities}

\subsection{Research questions}

A methodology described in \ref{sec:typology-identification-methodology} will be used to answer research questions, which are defined below.

\paragraph{RQ1}: Can regions in a city be differentiated based on public transport timetables?

\paragraph{RQ2}: Is it possible to define a typology based on those differences?

\paragraph{RQ3}: Will usage of hierarchical clustering result in a multi-level typology?

\paragraph{RQ4}: Can this typology differentiate between whole cities and big parts of cities? 

\paragraph{RQ5}: Will this typology be able to extract some types of regions which are repeated across the cities?

\subsection{Data}

This section describes datasets used in experiments and the process of obtaining this data. Cities selected for experiments will be alongside their characteristics and motivations for selecting them. 

The source of GTFS files for cities was OpenMobilityData archive\cite{openmobilitydata}. It is an open-source archive with over 1200 GTFS feeds from 50 countries, which are available for both city and country-level (such as railway schedules for a whole country). 

When selecting cities for analysis, the goal was to collect a diverse set to allow for a better typology definition. After comparing a list of wanted cities with available feeds 6 polish and 6 European cities were selected.

\paragraph{Poland}

This part presents a list of cities from Poland selected for analysis, alongside with reasons why each city was selected. The criteria when selecting cities was that a city should have more than one kind of public transport.

\begin{itemize}
    \item Wrocław - It is one of the major cities in Poland in which the author lives and studies, which gives him the best insight on how public transport is organized. Moreover, it is a very symmetrical, circular city, with an established city center. 
    
    \item Warszawa - A capital city of Poland and the only Polish city with a subway network. It also has a suburban railway as part of public transport. 
    
    \item Kraków - One of the biggest cities in Poland, with bus and tram.
    
    \item Poznań - Also one of the main cities in Poland, with good quality, bus, and tram network. 
    
    \item Bydgoszcz - Smaller city, but with tram network alongside buses. It is also the capital of its voivodeship. 
    
    \item Tricity (Gdańsk, Gdynia, Sopot) - All three cities are connected with a suburban railway, which is one of the most used in Poland. This metropolis is also the only line city from all selected cities, which makes it great for comparison with other, more circular cities. 
\end{itemize}

\paragraph{Europe}

In this part, selected European cities are listed and their selection is justified. 

\begin{itemize}
    \item Barcelona - A grid-based city, which is another type of city significantly different than others.
    
    \item Berlin - One of the biggest cities with public transport feeds available in Europe. It has very good public transport, which consists of buses, trams, subway and suburban railway.
    
    \item Brussels - Another big European capital city. The whole Brussels Capital Region was used, as those cities share public transport feed. This feature matches well with a selection of Tricity in Poland. 
    
    \item Leipzig - City in Germany with good quality of public transport and character similar to Wrocław. Here the only available GTFS file was outdated (it was from 2019).
    
    \item Prague - Big city, with subway alongside buses and trams. It also incorporates ferries as an additional public transport type. 
    
    \item Vilnius - A city east from Poland to differ from other western cities.
\end{itemize}

\paragraph{Dividing cities into micro regions}

As a method of dividing each city into micro-regions, Uber's H3 spatial index\cite{h3} with resolution 8 was applied. The resulting number of hexagons containing at least one stop in each city was presented in Table \ref{tab:hex-count}.

\begin{table}[ht]
    \caption{Number of regions in each city}
    \label{tab:hex-count}
    \centering
    \begin{tabular}{rrrrrr}
    \toprule
     Bydgoszcz &  Tricity &  Kraków &  Poznań &  Warszawa &  Wrocław \\
    \midrule
           157 &     374 &     349 &     254 &       561 &      271 \\
    \bottomrule
    \toprule
     Barcelona &  Berlin &  Brussels &  Liepzig &  Prague &  Vilnius \\
    \midrule
           120 &     974 &       214 &    252 &    501 &    336 \\
    \bottomrule
    \end{tabular}
\end{table}

\subsection{Clustering results analysis}

This section presents results from an exploratory analysis of obtained representation space and typology identification through hierarchical clustering approach, following the blueprint from Section \ref{sec:typology-identification-methodology}.

\begin{figure}[ht]
    \centering
    \includegraphics[width=0.70\textwidth]{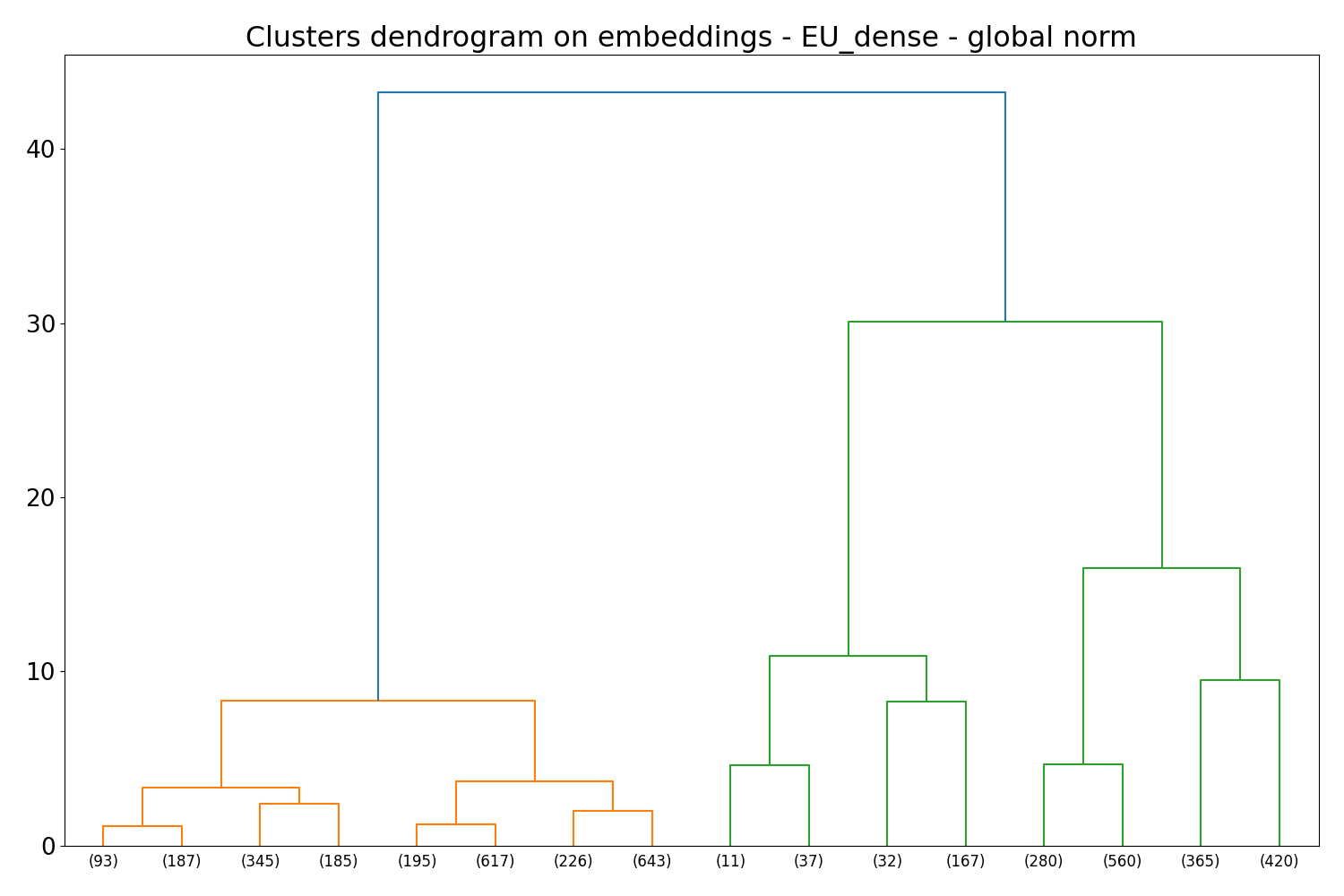}
    \caption{Dendrogram from agglomerative clustering process}
    \label{fig:dendrogram}
\end{figure}

Figure \ref{fig:dendrogram} presents a dendrogram from an agglomerative clustering process. The dendrogram is imbalanced. Cluster splits on the right are significantly higher which suggests that samples on the left are significantly different from those on the right but still with relatively small variance. Analysis in \ref{sec:2-clusters} will show that they differentiate city centers from suburban areas. The suburbs remain grouped in a single cluster for a couple of steps going down the dendrogram and increasing the number of clusters. 

\paragraph{2 clusters}\label{sec:2-clusters}

To make an analysis of clusters in data easier, a top-down approach was proposed. Then, by analyzing where did new clusters occurred and how was a new cluster formed, it was possible to characterize each level and obtained clusters.

\begin{figure}[H]
     \centering
     \begin{subfigure}{0.49\textwidth}
         \centering
         \includegraphics[width=\textwidth]{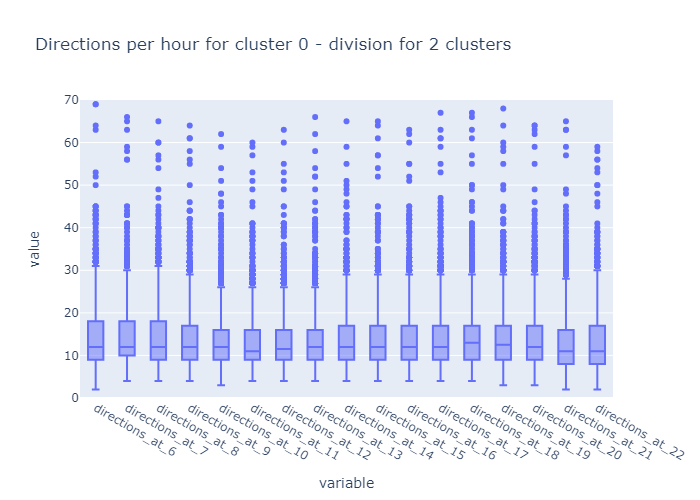}
         \caption{Directions - Cluster 0}
     \end{subfigure}
     \begin{subfigure}{0.49\textwidth}
         \centering
         \includegraphics[width=\textwidth]{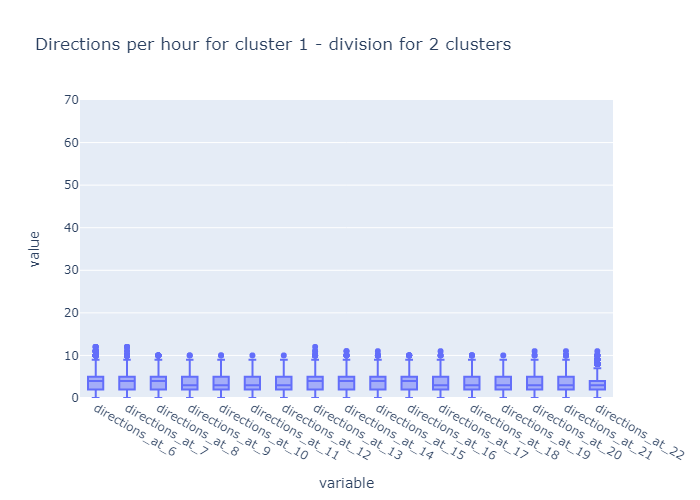}
         \caption{Directions - Cluster 1}
     \end{subfigure}
     \begin{subfigure}{0.49\textwidth}
         \centering
         \includegraphics[width=\textwidth]{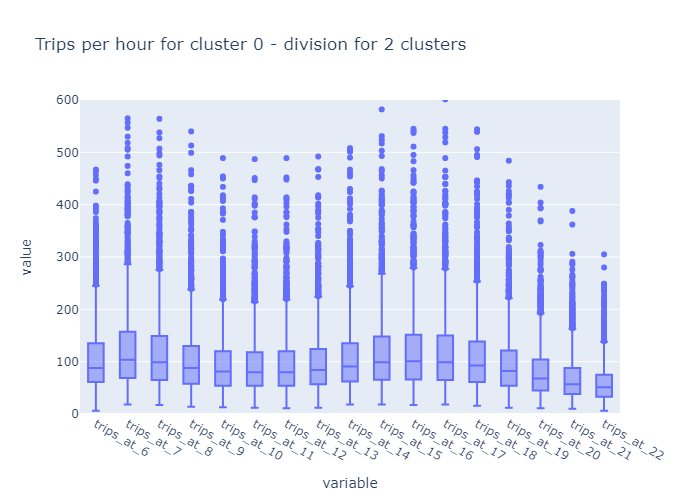}
         \caption{Trips - Cluster 0}
     \end{subfigure}
     \begin{subfigure}{0.49\textwidth}
         \centering
         \includegraphics[width=\textwidth]{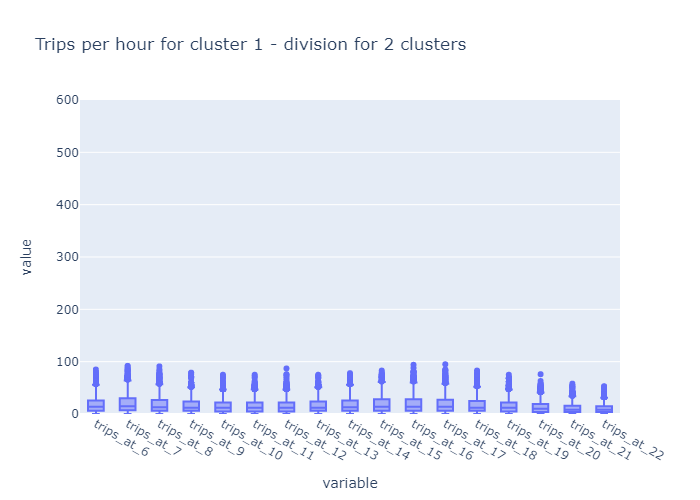}
         \caption{Trips - Cluster 1}
     \end{subfigure}
    \caption{Trips and directions during a day for each cluster - division for 2 clusters}
    \label{fig:2-clusters:directions-trips}
\end{figure}

\begin{figure}[H]
    \centering
    \includegraphics[width=0.75\textwidth]{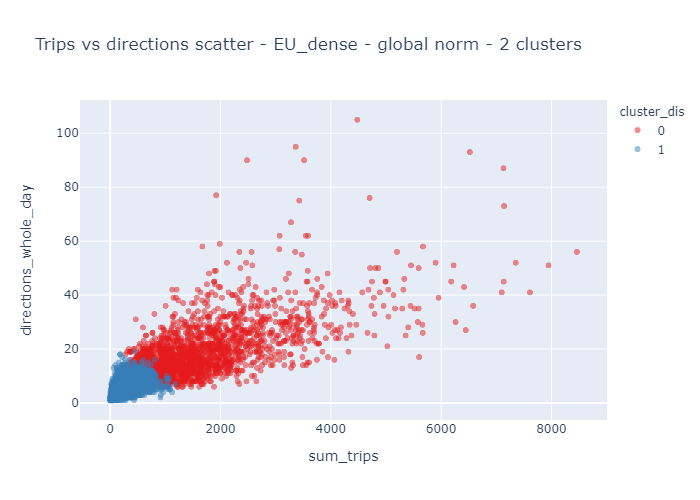}
    \caption{Aggregated features scatter plot - division for 2 clusters}
    \label{fig:2-clusters:aggregated}
\end{figure}

First, a scatter plot of aggregated features was prepared and is presented in Figure \ref{fig:2-clusters:aggregated}. It shows, that a dense area close to point $(0, 0)$ was separated from other points. To verify the variance in those two clusters, boxplots for each cluster were created and included in Figure \ref{fig:2-clusters:directions-trips}. Firstly, with the sum of trips from a region, the difference is clear, as cluster \textbf{1} has significantly smaller values through an entire day. However, both show a substantial number of outliers, which may suggest that some further divisions are possible without making clusters noisy. On the other hand, features related to directions available in cluster \textbf{1} show small variance, and a limited number of outliers. The same cannot be said about values in cluster \textbf{0}. This fact suggests, that it was cluster \textbf{1} which was separated from other points, at this level. 

\begin{figure}[H]
    \centering
    \begin{subfigure}{0.40\textwidth}
        \centering
        \includegraphics[width=\textwidth]{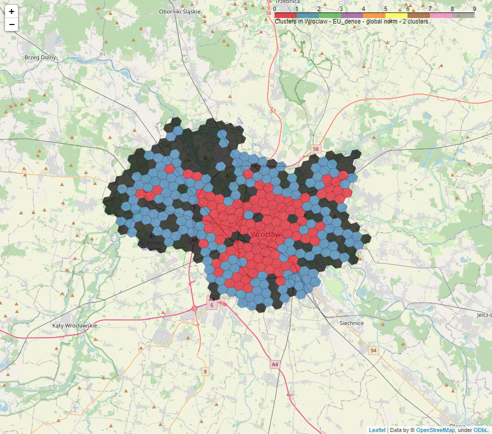}
        \caption{Wroclaw}
    \end{subfigure}
    \hspace{0.15\textwidth}%
    \begin{subfigure}{0.40\textwidth}
        \centering
        \includegraphics[width=\textwidth]{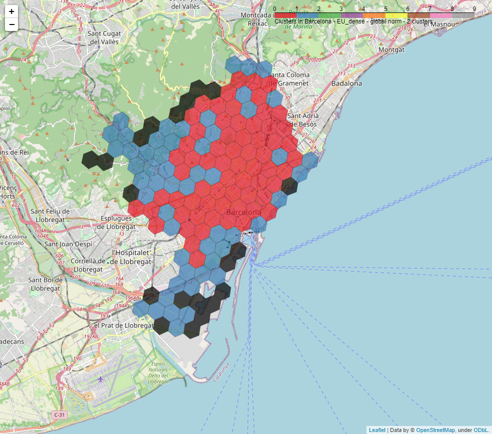}
        \caption{Barcelona}
    \end{subfigure}
    \begin{subfigure}{0.40\textwidth}
        \centering
        \includegraphics[width=\textwidth]{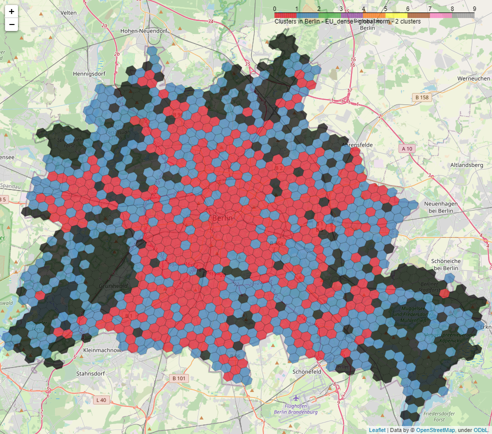}
        \caption{Berlin}
    \end{subfigure}
    \hspace{0.15\textwidth}%
    \begin{subfigure}{0.40\textwidth}
         \centering
         \includegraphics[width=\textwidth]{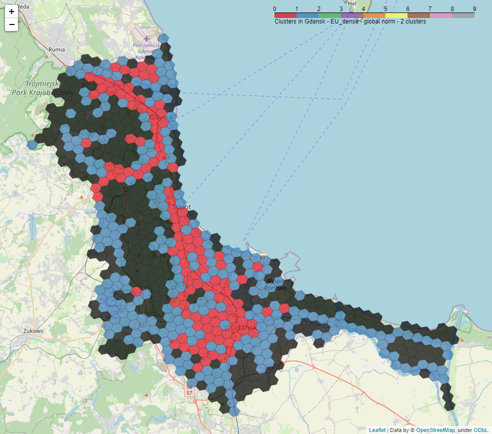}
         \caption{Tricity (Gdansk, Gdynia, Sopot)}
    \end{subfigure}
    \caption{Clusters visualization on maps - division for 2 clusters}
    \label{fig:2-clusters:maps}
\end{figure}

An analysis of values of features in clusters suggests that at this level suburban areas, which are only connected with single lines with a city center are aggregated in cluster \textbf{1}, leaving cluster \textbf{0} with regions with higher diversity of public transport. This can be further justified by analyzing visualizations on maps, which are presented in Figure \ref{fig:2-clusters:maps}. Another conclusion, drawn from the fact that for cluster \textbf{1} trips are more diversified than directions, is that the main characteristic of this cluster is a limited variety of public transport, but how often, a connection to the city center it is available can vary depending on a city.

Table \ref{tab:2-clusters:portion} presents a share of each cluster across all cities. In most cases, it is quite balanced or is close to a 1:2 ratio. Most of the cities have more regions that belong to suburban areas.

\begin{table}[H]
\centering
\caption{Percentage of regions in each cluster for all cities - division for 2 clusters}
\label{tab:2-clusters:portion}
\resizebox{0.9\columnwidth}{!}{%
\begin{tabular}{@{}rrrrrrrrrrrrr@{}}
\toprule
 & \rotatebox{90}{Barcelona} & \rotatebox{90}{Berlin} & \rotatebox{90}{Brussels} & \rotatebox{90}{Bydgoszcz} & \rotatebox{90}{Gdansk} 
& \rotatebox{90}{Krakow} & \rotatebox{90}{Liepzig} & \rotatebox{90}{Poznan} & \rotatebox{90}{Prague} & \rotatebox{90}{Warsaw} 
& \rotatebox{90}{Vilnius} & \rotatebox{90}{Wroclaw} \\ \midrule
0    & 55.00     & 46.20  & 69.16    & 31.21     & 31.82  & 35.82  & 34.13 & 33.07  & 50.50 & 54.01    & 23.81 & 40.22   \\
1    & 45.00     & 53.80  & 30.84    & 68.79     & 68.18  & 64.18  & 65.87 & 66.93  & 49.50 & 45.99    & 76.19 & 59.78   \\ \bottomrule
\end{tabular}
}
\end{table}

\paragraph{3 clusters}

\begin{figure}[H]
    \centering
    \includegraphics[width=0.75\textwidth]{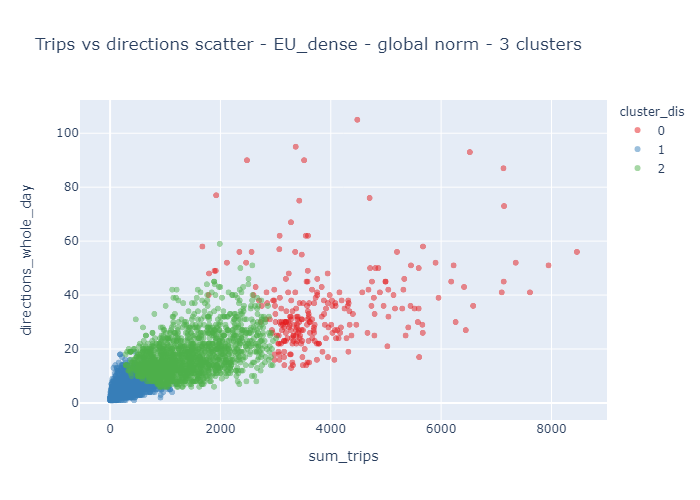}
    \caption{Aggregated features scatter plot - division for 3 clusters}
    \label{fig:3-clusters:aggregated}
\end{figure}

Moving down in the clusters' dendrogram, the opposite end is separated as a cluster. As expected, cluster 1 from the previous step, which contained suburban regions, remained unchanged. Figure \ref{fig:3-clusters:aggregated}, suggest that a red cluster - \textbf{0} - which was formed, groups regions with high volume of public transport available. It is a logical step, to separate transportation hubs, which proves, that the proposed method provides good quality grouping results. 

\begin{figure}[H]
     \centering
     \begin{subfigure}{0.49\textwidth}
         \centering
         \includegraphics[width=\textwidth]{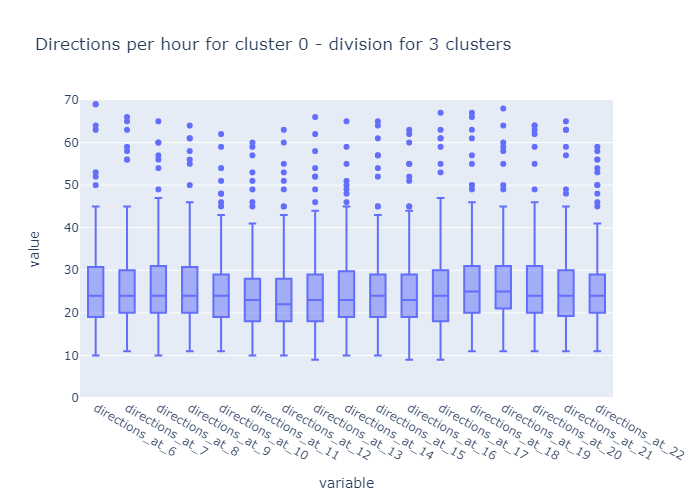}
         \caption{Directions - Cluster 0}
     \end{subfigure}
     \begin{subfigure}{0.49\textwidth}
         \centering
         \includegraphics[width=\textwidth]{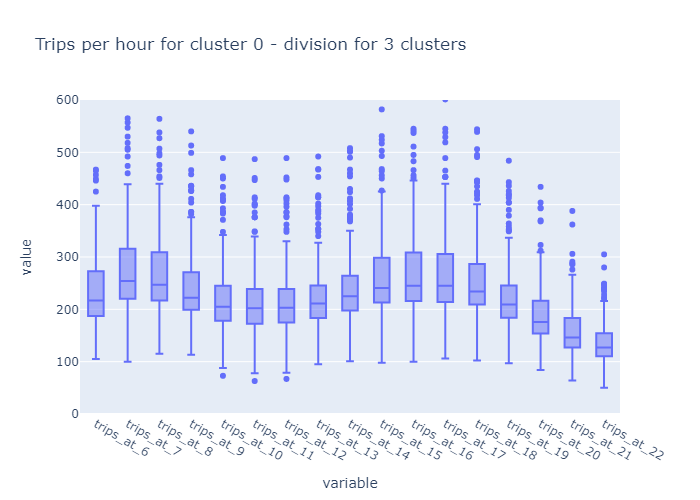}
         \caption{Trips - Cluster 0}
     \end{subfigure}
     \begin{subfigure}{0.49\textwidth}
         \centering
         \includegraphics[width=\textwidth]{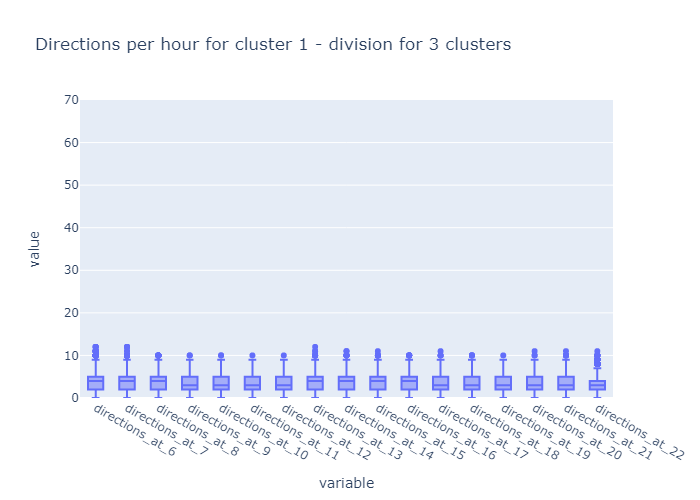}
         \caption{Directions - Cluster 1}
     \end{subfigure}
     \begin{subfigure}{0.49\textwidth}
         \centering
         \includegraphics[width=\textwidth]{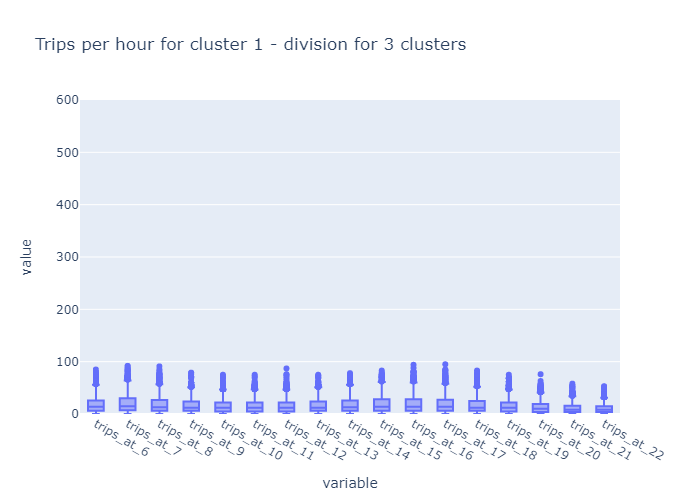}
         \caption{Trips - Cluster 1}
     \end{subfigure}
     \begin{subfigure}{0.49\textwidth}
         \centering
         \includegraphics[width=\textwidth]{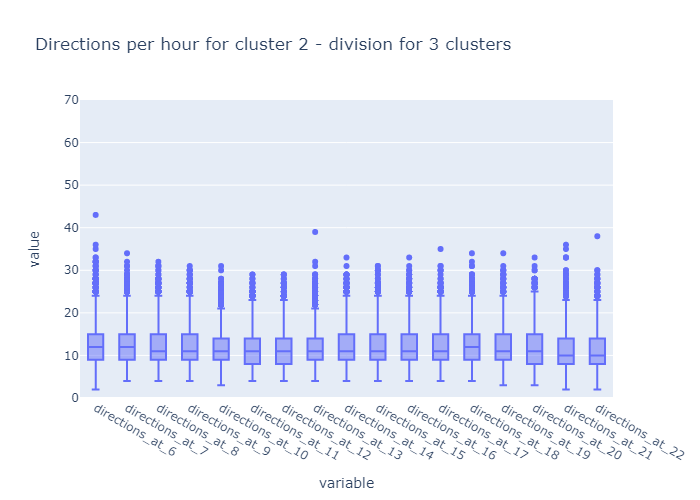}
         \caption{Directions - Cluster 2}
     \end{subfigure}
     \begin{subfigure}{0.49\textwidth}
         \centering
         \includegraphics[width=\textwidth]{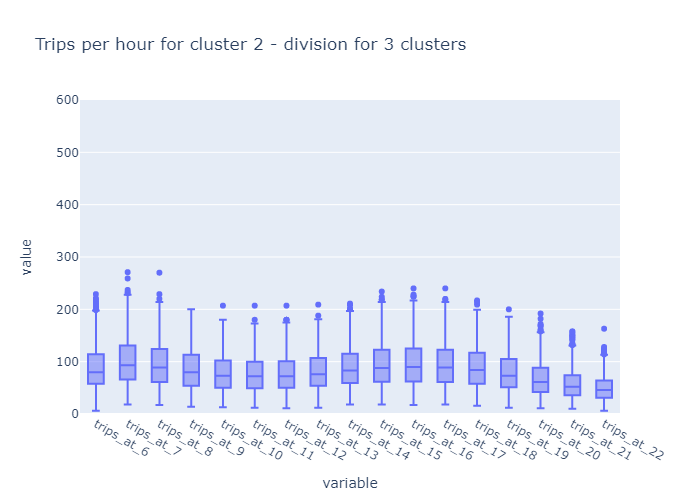}
         \caption{Trips - Cluster 2}
     \end{subfigure}
    \caption{Directions and trips during a day for each cluster - division for 3 clusters}
    \label{fig:3-clusters:directions-trips}
\end{figure}

Figure \ref{fig:3-clusters:directions-trips} contains box-plots with hour-features in both directions and sum of trips from each region. Since cluster \textbf{1} remained unchanged and was characterized in a previous step, an analysis will focus on differences between clusters \textbf{0} and \textbf{2}. 

Cluster \textbf{0}, as mentioned earlier, seems to group transportation hubs across the cities. It has more than 2-3 times higher average values for all features compared to cluster \textbf{2}. It has also higher variance and more outliers in both types of features, which may mean that similarly to cluster \textbf{1}, the transportation availability in hubs is different across the cities. This is an obvious observation since some cities like Berlin have more transportation options and better public transport. However, the fact that across the cities, most of the transportation hubs land in this cluster is another argument that the proposed solution may bring valuable insight into public transport availability. 

\begin{table}[t]
\centering
\caption{Percentage of regions in each cluster for all cities - division for 3 clusters}
\label{tab:3-clusters:portion}
\resizebox{\columnwidth}{!}{%
\begin{tabular}{@{}rrrrrrrrrrrrr@{}}
\toprule
 & \rotatebox{90}{Barcelona} & \rotatebox{90}{Berlin} & \rotatebox{90}{Brussels} & \rotatebox{90}{Bydgoszcz} & \rotatebox{90}{Gdansk} 
& \rotatebox{90}{Krakow} & \rotatebox{90}{Liepzig} & \rotatebox{90}{Poznan} & \rotatebox{90}{Prague} & \rotatebox{90}{Warsaw} 
& \rotatebox{90}{Vilnius} & \rotatebox{90}{Wroclaw} \\ \midrule
0 & 10.00     & 5.03   & 12.15    & 1.27      & 2.14   & 5.44   & 1.98  & 1.18   & 7.58  & 11.94    & 0.89  & 5.54    \\
1 & 45.00     & 53.80  & 30.84    & 68.79     & 68.18  & 64.18  & 65.87 & 66.93  & 49.50 & 45.99    & 76.19 & 59.78   \\
2 & 45.00     & 41.17  & 57.01    & 29.94     & 29.68  & 30.37  & 32.14 & 31.89  & 42.91 & 42.07    & 22.92 & 34.69   \\ \bottomrule
\end{tabular}
}
\end{table}

Table \ref{tab:3-clusters:portion} presents a percentage of regions which were placed in each cluster. At this level, an analysis of the share of regions from cluster \textbf{0}, which represents hubs, shows that there are significant differences between cities. However, some unexpected similarities were identified, like a comparable share of hubs in Wrocław and Berlin. 

\begin{figure}[H]
    \centering
    \begin{subfigure}{0.40\textwidth}
         \centering
         \includegraphics[width=\textwidth]{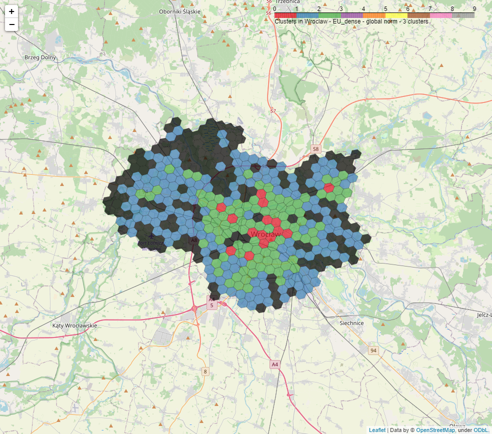}
         \caption{Wroclaw}
    \end{subfigure}
    \hspace{0.15\textwidth}%
    \begin{subfigure}{0.40\textwidth}
         \centering
         \includegraphics[width=\textwidth]{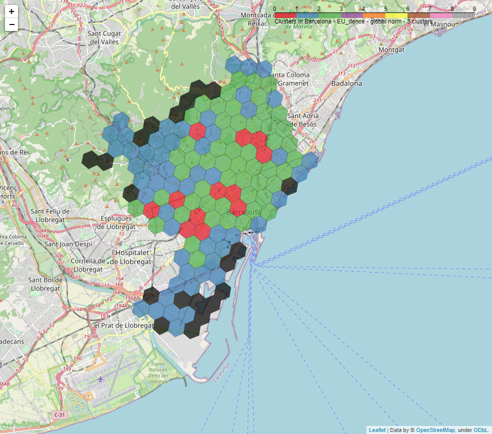}
         \caption{Barcelona}
    \end{subfigure}
    \begin{subfigure}{0.40\textwidth}
         \centering
         \includegraphics[width=\textwidth]{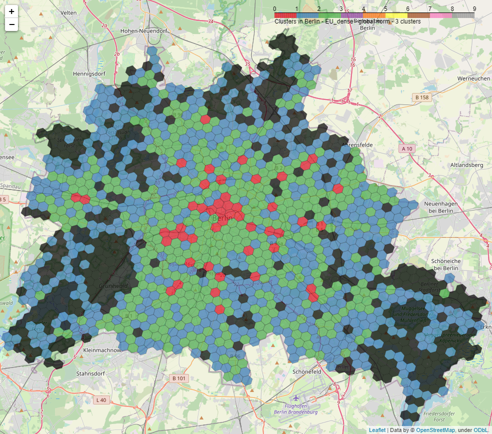}
         \caption{Berlin}
     \end{subfigure}
        \hspace{0.15\textwidth}%
     \begin{subfigure}{0.40\textwidth}
         \centering
         \includegraphics[width=\textwidth]{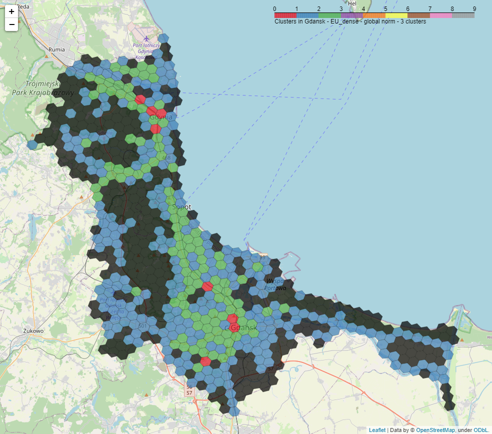}
         \caption{Tricity (Gdansk, Gdynia, Sopot)}
     \end{subfigure}
    \caption{Clusters visualization on maps - division for 3 clusters}
    \label{fig:3-clusters:maps}
\end{figure}

Lastly, a visual analysis with maps on Figure \ref{fig:3-clusters:maps} shows, that points from cluster \textbf{0} match with localizations of transportation hubs in each city. It also shows differences between cities which is represented in a different number of red regions in different cities. 

\paragraph{4 clusters}

Addition of one more cluster resulted in division of cluster \textbf{2} which contained regions from city centers, which were not transportation hubs into cluster \textbf{2} with higher volume of public transport and cluster \textbf{3} with less public transport available. Suburban areas (cluster \textbf{1}) and transportation hubs (cluster \textbf{0}) remained unchanged. 

\begin{figure}[H]
    \centering
    \includegraphics[width=0.75\textwidth]{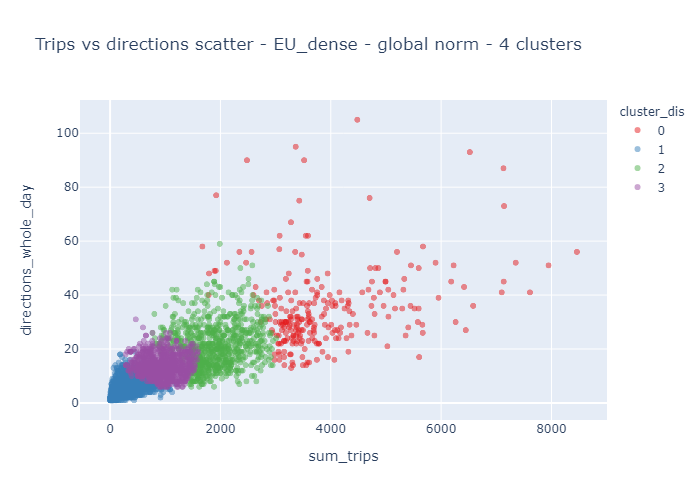}
    \caption{Aggregated features scatter plot - division for 4 clusters}
    \label{fig:4-clusters:aggregated}
\end{figure}

Figure \ref{fig:4-clusters:aggregated} presets clusters on scatter plot of aggregated features. There is no clear vertical or horizontal division between newly formed clusters, which suggests that the difference between them is in the general volume of public transport available. 

\begin{figure}[H]
    \centering
    \begin{subfigure}{0.45\textwidth}
        \centering
        \includegraphics[width=\textwidth]{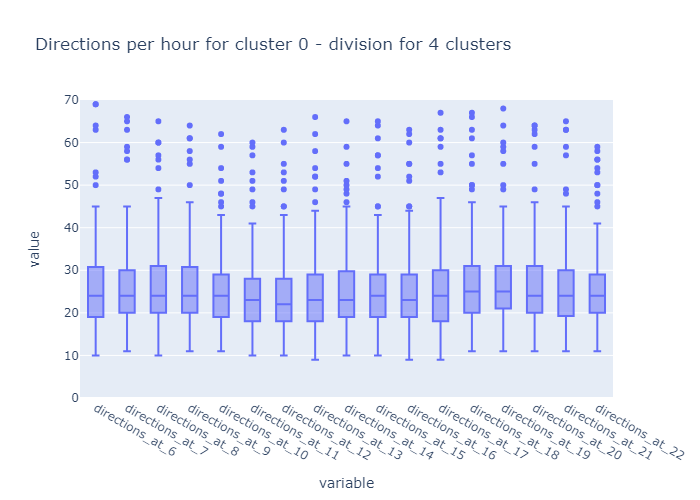}
        \caption{Directions - Cluster 0}
    \end{subfigure}
    \begin{subfigure}{0.45\textwidth}
        \centering
        \includegraphics[width=\textwidth]{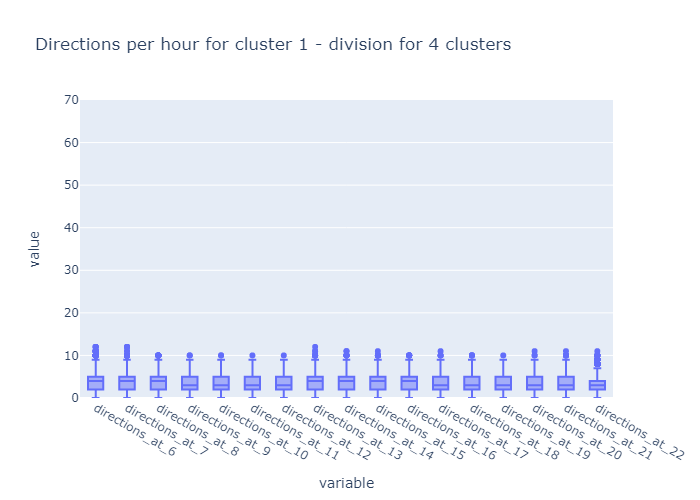}
        \caption{Directions - Cluster 1}
        \label{cluster5:1-directions}
        \label{cluster6:1-directions}
    \end{subfigure}
    \begin{subfigure}{0.45\textwidth}
        \centering
        \includegraphics[width=\textwidth]{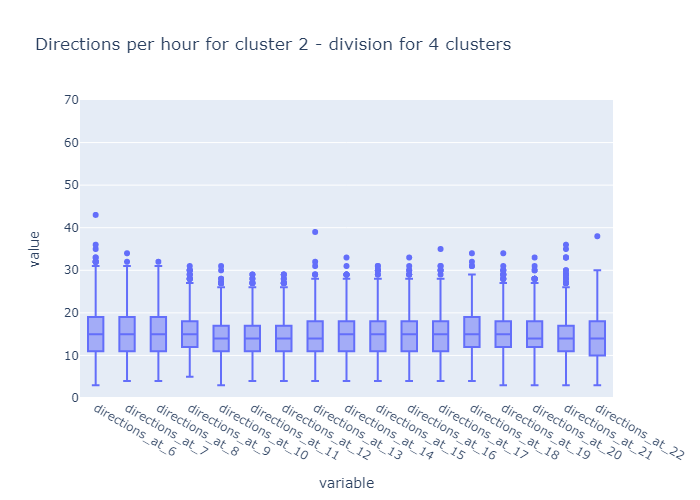}
        \caption{Directions - Cluster 2}
        \label{cluster5:2-directions}
    \end{subfigure}
    \begin{subfigure}{0.45\textwidth}
        \centering
        \includegraphics[width=\textwidth]{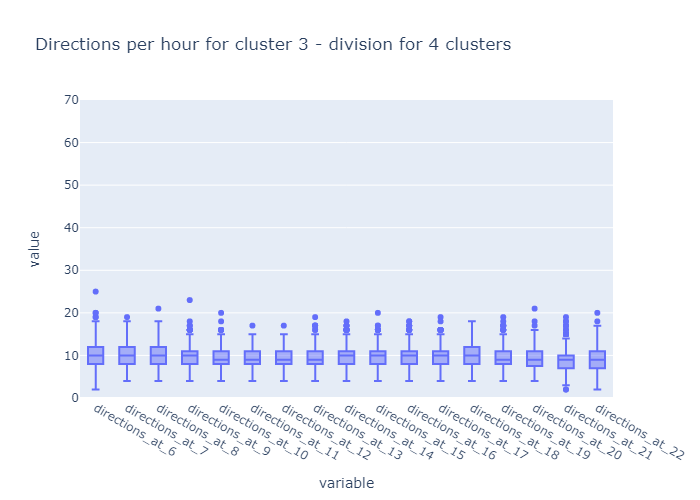}
        \caption{Directions - Cluster 3}
        \label{cluster5:3-directions}
        \label{cluster6:3-directions}
        \label{cluster7:3-directions}
        \label{cluster8:3-directions}
        \label{cluster9:3-directions}
    \end{subfigure}
    \begin{subfigure}{0.45\textwidth}
        \centering
        \includegraphics[width=\textwidth]{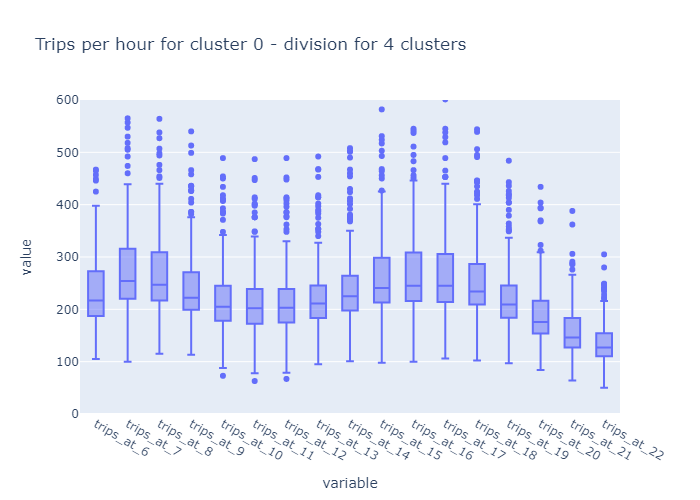}
        \caption{Trips - Cluster 0}
    \end{subfigure}
    \begin{subfigure}{0.45\textwidth}
        \centering
        \includegraphics[width=\textwidth]{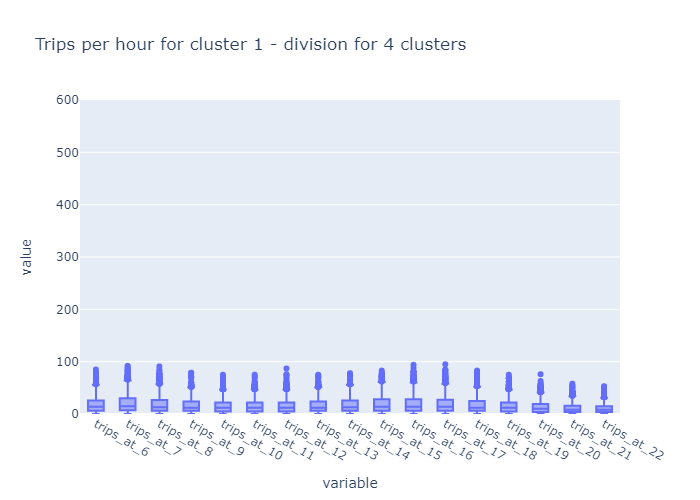}
        \caption{Trips - Cluster 1}
        \label{cluster5:1-trips}
        \label{cluster6:1-trips}
    \end{subfigure}
    \begin{subfigure}{0.45\textwidth}
        \centering
        \includegraphics[width=\textwidth]{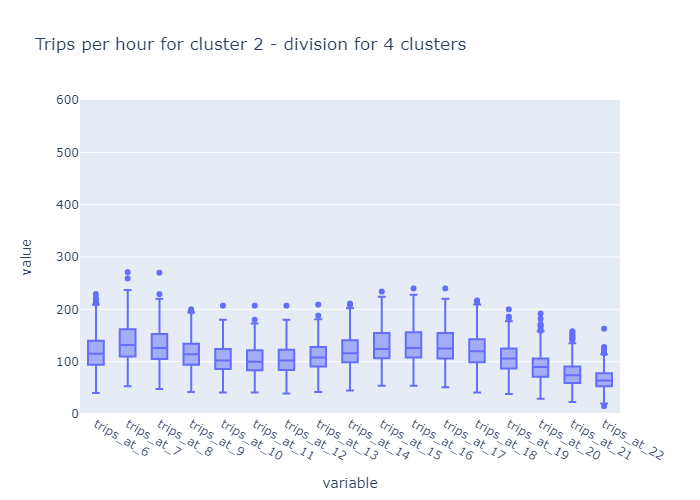}
        \caption{Trips - Cluster 2}
        \label{cluster5:2-trips}
    \end{subfigure}
    \begin{subfigure}{0.45\textwidth}
        \centering
        \includegraphics[width=\textwidth]{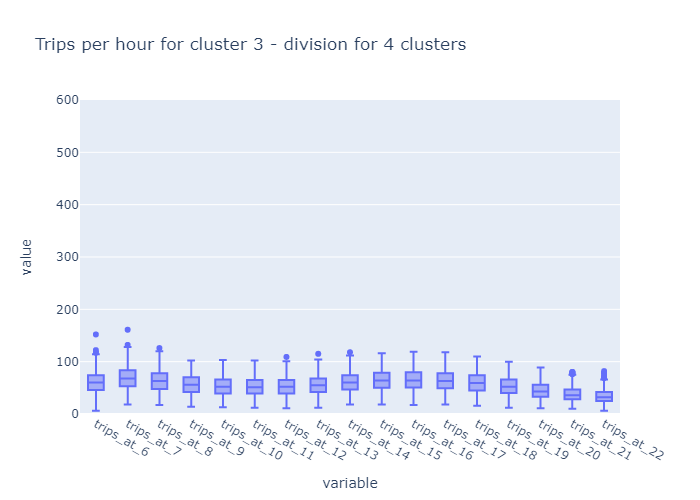}
        \caption{Trips - Cluster 3}
        \label{cluster5:3-trips}
        \label{cluster6:3-trips}
        \label{cluster7:3-trips}
        \label{cluster8:3-trips}
        \label{cluster9:3-trips}
    \end{subfigure}
    \caption{Directions and trips during a day for each cluster - division for 4 clusters}
    \label{fig:4-clusters:directions-trips}
\end{figure}

Analysis of box plots for per-hour features, which is included in Figure \ref{fig:4-clusters:directions-trips}, brings more insight into how newly formed clusters differ. Previously, directions in cluster combining tho newly formed contained a lot of outliers. This is not the case for new clusters \textbf{2} and \textbf{3}. Trips numbers accordingly have smaller variance. It suggests, that this division resulted in good quality clustering with small intra-cluster variance, which was the desired outcome. 

Table \ref{tab:4-clusters:portion} contains a number of regions placed in each cluster across all cities. In most cases, a predominant number of regions falls into cluster \textbf{3}, which contains regions with worse public transport availability. Only cities, which have it the opposite way, are the same which have significantly more hubs than others. Among the remaining cities, there are clear differences in the share of clusters \textbf{2} and \textbf{3}. This suggests that the proposed method is able to differentiate cities and not only find regions of similar characteristics in different cities, which was a motivation when applying a global normalization strategy. 

\begin{figure}[H]
    \centering
    \begin{subfigure}{0.40\textwidth}
         \centering
         \includegraphics[width=\textwidth]{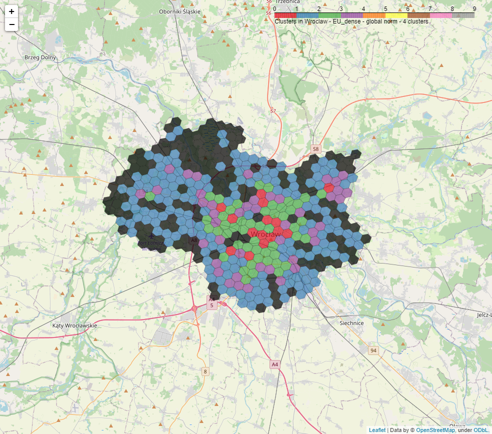}
         \caption{Wroclaw}
    \end{subfigure}
    \hspace{0.15\textwidth}%
    \begin{subfigure}{0.40\textwidth}
         \centering
         \includegraphics[width=\textwidth]{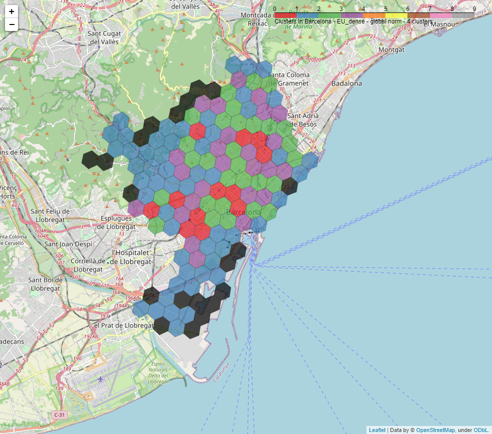}
         \caption{Barcelona}
    \end{subfigure}
    \begin{subfigure}{0.40\textwidth}
         \centering
         \includegraphics[width=\textwidth]{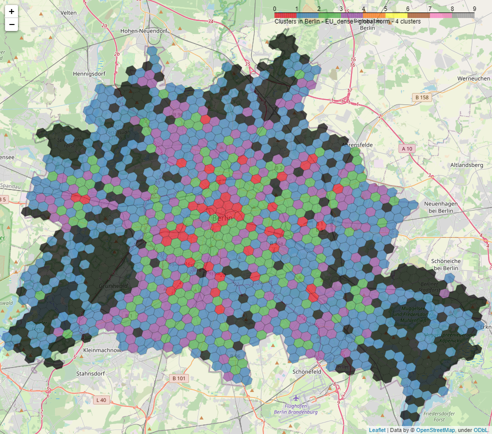}
         \caption{Berlin}
    \end{subfigure}
    \hspace{0.15\textwidth}%
    \begin{subfigure}{0.40\textwidth}
        \centering
        \includegraphics[width=\textwidth]{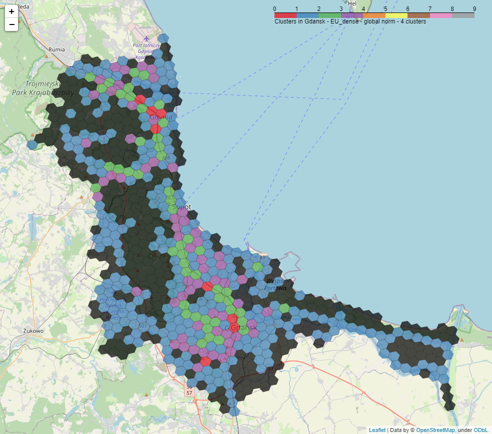}
        \caption{Tricity (Gdansk, Gdynia, Sopot)}
    \end{subfigure}
    \caption{Clusters visualization on maps - division for 4 clusters}
    \label{fig:4-clusters:maps}
\end{figure}

For visual analysis, clusters distribution on city maps is presented in Figure \ref{fig:4-clusters:maps}. The difference between a circular city (Wrocław) and a line city (Tricity) is clear, which is a good sign because there was only one line city in a dataset.

Based on those observations, this number of clusters seems to be a strong candidate to create a level in a typology of public transport availability across the cities. 

\begin{table}[h]
\centering
\caption{Percentage of regions in each cluster for all cities - division for 4 clusters}
\label{tab:4-clusters:portion}
\resizebox{\columnwidth}{!}{%
\begin{tabular}{@{}rrrrrrrrrrrrr@{}}
\toprule
 & \rotatebox{90}{Barcelona} & \rotatebox{90}{Berlin} & \rotatebox{90}{Brussels} & \rotatebox{90}{Bydgoszcz} & \rotatebox{90}{Gdansk} 
& \rotatebox{90}{Krakow} & \rotatebox{90}{Liepzig} & \rotatebox{90}{Poznan} & \rotatebox{90}{Prague} & \rotatebox{90}{Warsaw} 
& \rotatebox{90}{Vilnius} & \rotatebox{90}{Wroclaw} \\ \midrule
0    & 10.00     & 5.03   & 12.15    & 1.27      & 2.14   & 5.44   & 1.98  & 1.18   & 7.58  & 11.94    & 0.89  & 5.54    \\
1    & 45.00     & 53.80  & 30.84    & 68.79     & 68.18  & 64.18  & 65.87 & 66.93  & 49.50 & 45.99    & 76.19 & 59.78   \\
2    & 24.17     & 19.20  & 39.72    & 12.10     & 13.37  & 14.04  & 9.52  & 9.45   & 20.56 & 22.82    & 11.01 & 18.45   \\
3    & 20.83     & 21.97  & 17.29    & 17.83     & 16.31  & 16.33  & 22.62 & 22.44  & 22.36 & 19.25    & 11.90 & 16.24   \\ \bottomrule
\end{tabular}
}
\end{table}

\paragraph{5 clusters}

\begin{figure}[H]
     \centering
     \begin{subfigure}{0.49\textwidth}
         \centering
         \includegraphics[width=\textwidth]{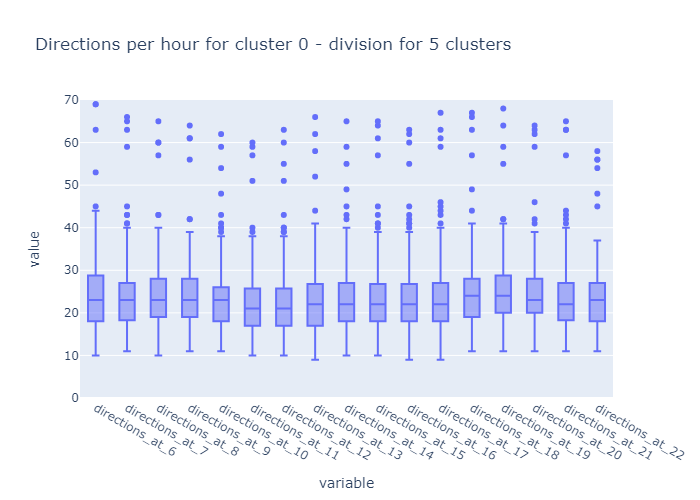}
         \caption{Directions - Cluster 0}
         \label{cluster5:0-directions}
         \label{cluster6:0-directions}
         \label{cluster7:0-directions}
     \end{subfigure}
     \begin{subfigure}{0.49\textwidth}
         \centering
         \includegraphics[width=\textwidth]{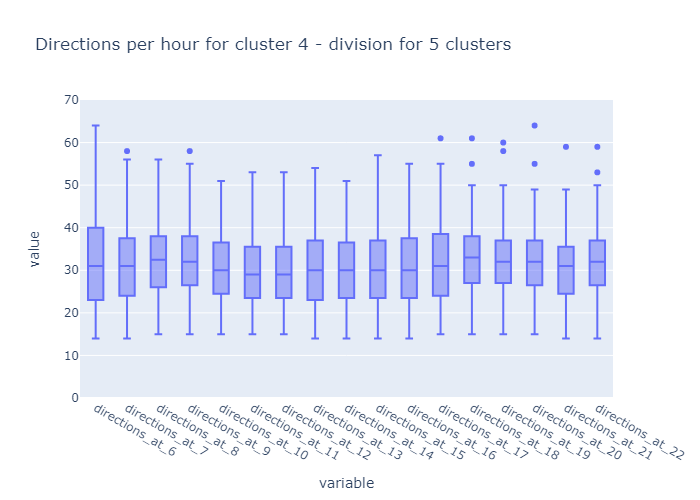}
         \caption{Directions - Cluster 4}
         \label{cluster5:4-directions}
         \label{cluster6:4-directions}
         \label{cluster7:4-directions}
         \label{cluster8:4-directions}
         \label{cluster9:4-directions}
     \end{subfigure}
          \begin{subfigure}{0.49\textwidth}
         \centering
         \includegraphics[width=\textwidth]{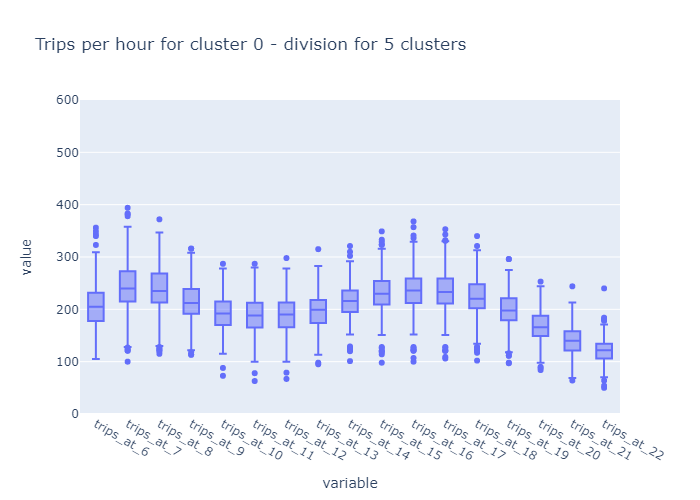}
         \caption{Trips - Cluster 0}
         \label{cluster5:0-trips}
         \label{cluster6:0-trips}
         \label{cluster7:0-trips}
     \end{subfigure}
     \begin{subfigure}{0.49\textwidth}
         \centering
         \includegraphics[width=\textwidth]{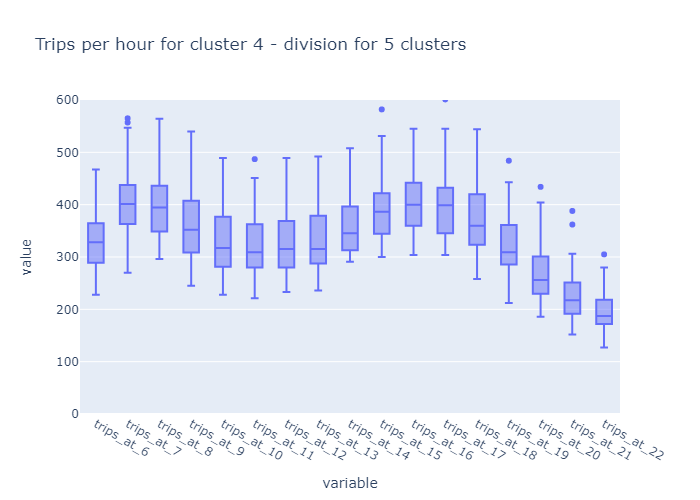}
         \caption{Trips - Cluster 4}
         \label{cluster5:4-trips}
         \label{cluster6:4-trips}
         \label{cluster7:4-trips}
         \label{cluster8:4-trips}
         \label{cluster9:4-trips}
     \end{subfigure}
    \caption{Directions and trips during a day for new clusters - division for 5 clusters}
    \label{fig:5-clusters:directions-trips}
\end{figure}

Starting from the division for 5 clusters, the plots will only include two clusters which were formed at this level, as a result of a single division. Other clusters will have the same values distributions as previous levels. Numbering of clusters is maintained for all levels, so it is easy to find matching plots. To give an example, at level 5 cluster 0 is divided and forms clusters 0 and 4. Those two will be presented in this part, since clusters 1, 2, 3 have not changed compared to level 4. To make it easier to find the right plots, each section will start with references to correct plots. For level 5 those are: 
\begin{itemize}
    \item cluster 0 - \ref{cluster5:0-directions} and \ref{cluster5:0-trips}
    \item cluster 1 - \ref{cluster5:1-directions} and \ref{cluster5:1-trips}
    \item cluster 2 - \ref{cluster5:2-directions} and \ref{cluster5:2-trips}
    \item cluster 3 - \ref{cluster5:3-directions} and \ref{cluster5:3-trips}
    \item cluster 4 - \ref{cluster5:4-directions} and \ref{cluster5:4-trips}
\end{itemize}

\begin{figure}[H]
    \centering
    \includegraphics[width=0.75\textwidth]{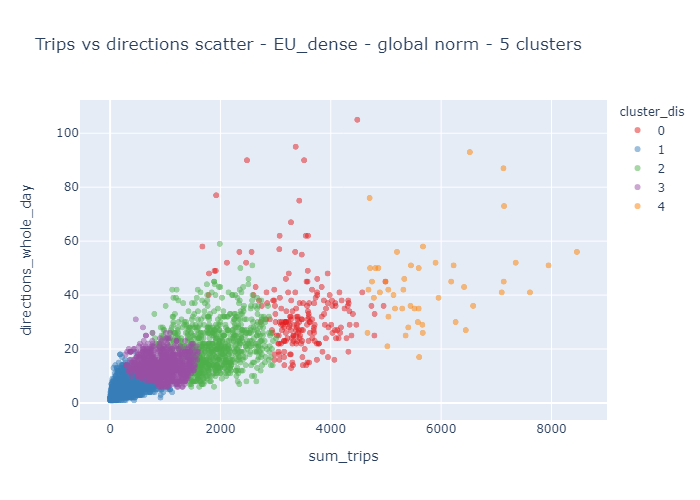}
    \caption{Aggregated features scatter plot - division for 5 clusters}
    \label{fig:5-clusters:aggregated}
\end{figure}

Division for 5 clusters breaks cluster \textbf{0} from the previous level, which contained regions identifies as transportation hubs with a high volume of public transport. Looking at a Figure \ref{fig:5-clusters:aggregated}, which presents a scatter plot of aggregated features, reveals that there was a first vertical division between clusters. By vertical one should understand the difference in trips features not directions. It is another distinction that the previous division for 4 clusters concludes a level in a typology that contained only divisions based on total public transport availability.

Figure \ref{fig:5-clusters:directions-trips} presents box plots of features describing directions available from clusters in each hour. There is no clear distinction between cluster \textbf{0} and \textbf{4} in terms of average value and variance. Cluster \textbf{0} shows more outliers, which suggests that this cluster may be divided when moving deeper into a dendrogram. On the other hand, a difference in the sum of trips in each hour is significant, which can be seen in Figure \ref{fig:5-clusters:directions-trips}. It is in compliance with a previous conclusion based on aggregated features. 

Table \ref{tab:5-clusters:portion} presents share of regions in each cluster for all cities included in research. For the first time, there are cities without regions in one of the clusters (cluster \textbf{4} - the highest volume of public transport). This is another argument for the use of the four clusters when defining the typology. Furthermore, regions from cluster \textbf{4} are in minority in each city, which suggests, that this cluster represents clear hubs in terms of the number of trips available.

\begin{figure}[H]
    \centering
    \begin{subfigure}{0.40\textwidth}
         \centering
         \includegraphics[width=\textwidth]{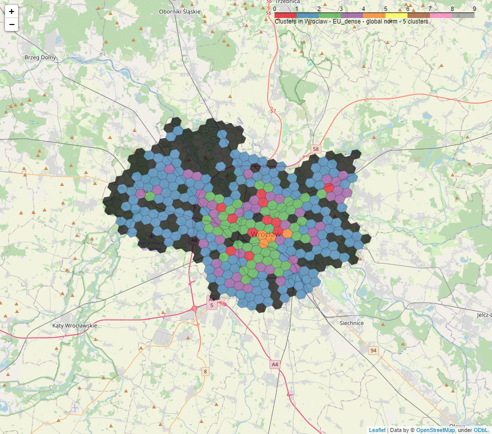}
         \caption{Wroclaw}
    \end{subfigure}
    \hspace{0.15\textwidth}%
    \begin{subfigure}{0.40\textwidth}
         \centering
         \includegraphics[width=\textwidth]{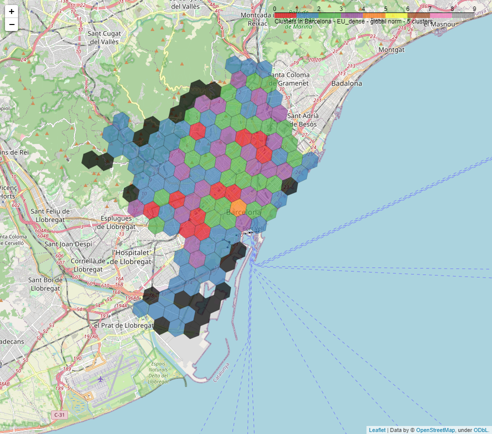}
         \caption{Barcelona}
    \end{subfigure}
    \begin{subfigure}{0.40\textwidth}
         \centering
         \includegraphics[width=\textwidth]{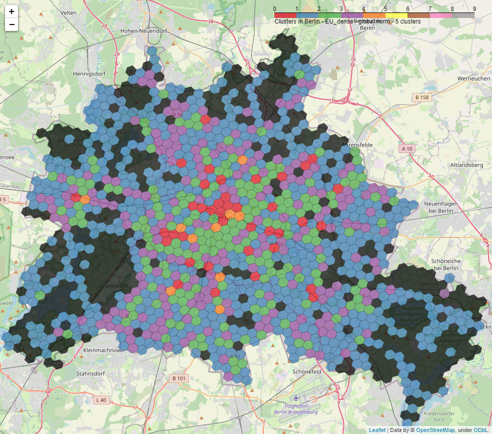}
         \caption{Berlin}
     \end{subfigure}
     \hspace{0.15\textwidth}%
     \begin{subfigure}{0.40\textwidth}
         \centering
         \includegraphics[width=\textwidth]{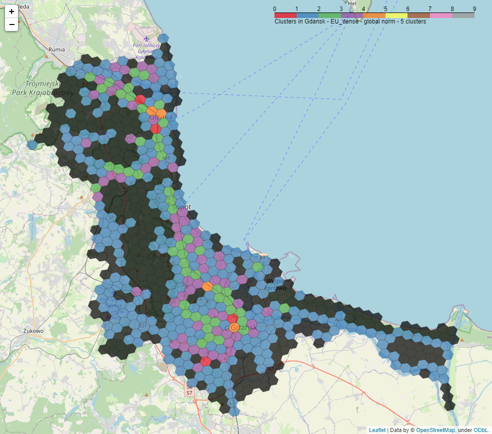}
         \caption{Tricity (Gdansk, Gdynia, Sopot)}
     \end{subfigure}
    \caption{Clusters visualization on maps - division for 5 clusters}
    \label{fig:5-clusters:maps}
\end{figure}

Finally, in Figure \ref{fig:5-clusters:maps} visualizations of clusters on a map for a subset of cities are presented. Newly identifies cluster \textbf{4} - marked with orange color - is located in places which have a lot of public transport options available, such as train stations. Moreover, the distribution of those regions in each city is different which may be an indicator of public transport quality. Berlin for example has a similar share of those to Wrocław, but they are better distributed within the city. 

\begin{table}[h]
\centering
\caption{Percentage of regions in each cluster for all cities - division for 5 clusters}
\label{tab:5-clusters:portion}
\resizebox{\columnwidth}{!}{%
\begin{tabular}{@{}rrrrrrrrrrrrr@{}}
\toprule
 & \rotatebox{90}{Barcelona} & \rotatebox{90}{Berlin} & \rotatebox{90}{Brussels} & \rotatebox{90}{Bydgoszcz} & \rotatebox{90}{Gdansk} 
& \rotatebox{90}{Krakow} & \rotatebox{90}{Liepzig} & \rotatebox{90}{Poznan} & \rotatebox{90}{Prague} & \rotatebox{90}{Warsaw} 
& \rotatebox{90}{Vilnius} & \rotatebox{90}{Wroclaw} \\ \midrule
0    & 9.17      & 4.00   & 8.88     & 1.27      & 1.07   & 4.58   & 1.98  & 1.18   & 6.39  & 9.63     & 0.89  & 4.06    \\
1    & 45.00     & 53.80  & 30.84    & 68.79     & 68.18  & 64.18  & 65.87 & 66.93  & 49.50 & 45.99    & 76.19 & 59.78   \\
2    & 24.17     & 19.20  & 39.72    & 12.10     & 13.37  & 14.04  & 9.52  & 9.45   & 20.56 & 22.82    & 11.01 & 18.45   \\
3    & 20.83     & 21.97  & 17.29    & 17.83     & 16.31  & 16.33  & 22.62 & 22.44  & 22.36 & 19.25    & 11.90 & 16.24   \\
4    & 0.83      & 1.03   & 3.27     & 0.00      & 1.07   & 0.86   & 0.00  & 0.00   & 1.20  & 2.32     & 0.00  & 1.48    \\ \bottomrule
\end{tabular}
}
\end{table}

\paragraph{6 clusters}

\begin{figure}[H]
     \centering
     \begin{subfigure}{0.49\textwidth}
         \centering
         \includegraphics[width=\textwidth]{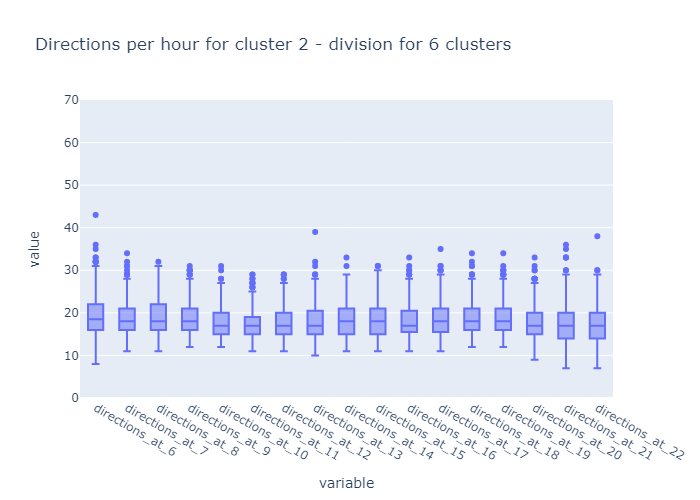}
         \caption{Cluster 2}
         \label{cluster6:2-directions}
         \label{cluster7:2-directions}
         \label{cluster8:2-directions}
     \end{subfigure}
     \begin{subfigure}{0.49\textwidth}
         \centering
         \includegraphics[width=\textwidth]{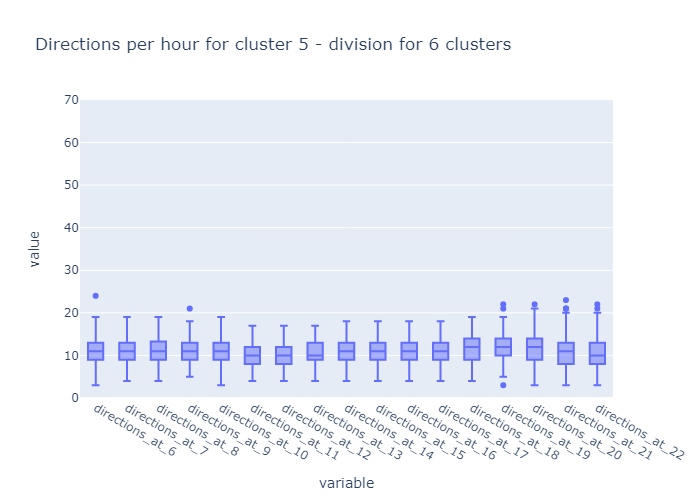}
         \caption{Cluster 5}
         \label{cluster6:5-directions}
         \label{cluster7:5-directions}
         \label{cluster8:5-directions}
         \label{cluster9:5-directions}
     \end{subfigure}
    \caption{Directions during a day for new clusters - division for 6 clusters}
    \label{fig:6-clusters:directions}
\end{figure}

\begin{figure}[H]
     \centering
     \begin{subfigure}{0.49\textwidth}
         \centering
         \includegraphics[width=\textwidth]{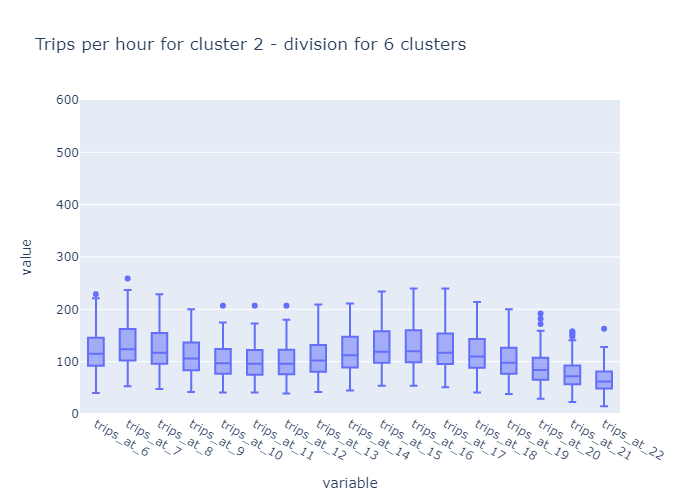}
         \caption{Cluster 2}
         \label{cluster6:2-trips}
         \label{cluster7:2-trips}
         \label{cluster8:2-trips}
     \end{subfigure}
     \begin{subfigure}{0.49\textwidth}
         \centering
         \includegraphics[width=\textwidth]{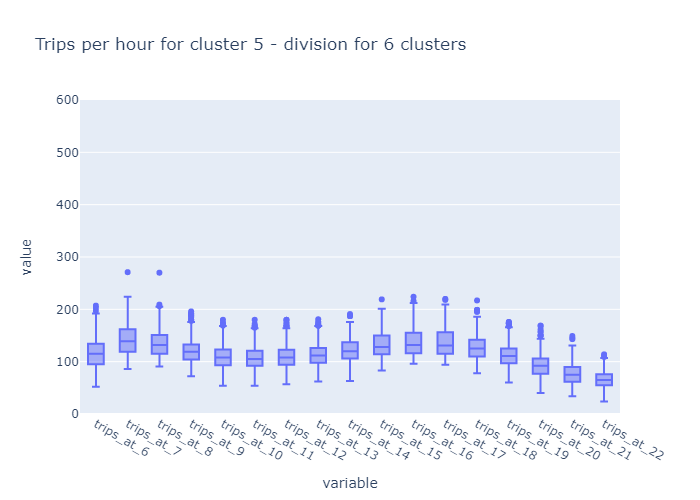}
         \caption{Cluster 5}
         \label{cluster6:5-trips}
         \label{cluster7:5-trips}
         \label{cluster8:5-trips}
         \label{cluster9:5-trips}
     \end{subfigure}
    \caption{Trips during a day for new clusters - division for 6 clusters}
    \label{fig:6-clusters:trips}
\end{figure}

Plots for this level:

\begin{itemize}
    \item cluster 0 - \ref{cluster6:0-directions} and \ref{cluster6:0-trips}
    \item cluster 1 - \ref{cluster6:1-directions} and \ref{cluster6:1-trips}
    \item cluster 2 - \ref{cluster6:2-directions} and \ref{cluster6:2-trips}
    \item cluster 3 - \ref{cluster6:3-directions} and \ref{cluster6:3-trips}
    \item cluster 4 - \ref{cluster6:4-directions} and \ref{cluster6:4-trips}
    \item cluster 5 - \ref{cluster6:5-directions} and \ref{cluster6:5-trips}
\end{itemize}

\begin{figure}[H]
    \centering
    \includegraphics[width=0.75\textwidth]{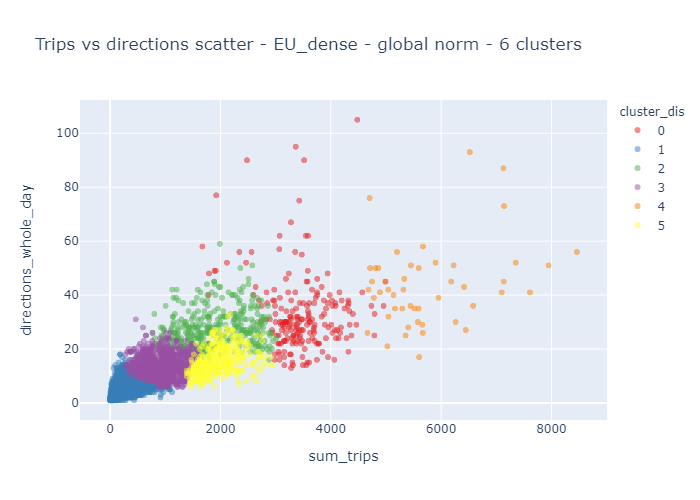}
    \caption{Aggregated features scatter plot - division for 6 clusters}
    \label{fig:6-clusters:aggregated}
\end{figure}

When division for 5 clusters was presented a vertical division was first observed. Now, at a level with 6 clusters first horizontal division occurs, which can be seen in Figure \ref{fig:6-clusters:aggregated}. It shows a division in cluster \textbf{2}, which represented regions from the city center without a high volume of public transport. Now, with an occurrence of cluster \textbf{5}, this cluster is split in terms of available directions. 

Figures \ref{fig:6-clusters:directions} brings more insight on how did this division affected clustering. For comparison, a combined cluster is presented in Figure \ref{cluster5:2-directions}. New cluster \textbf{5} has smaller values with less variance on those dimensions. Sum of trips is similar in both cluster \textbf{2} and \textbf{5} (see \ref{fig:6-clusters:trips}) which fits an observation that this division is based mostly on available directions. To conclude, the differences between those clusters are marginal, and therefore further divisions in this area of "medium" public transport should not be taken into considerations when defining a final typology.

\begin{figure}[H]
    \centering
    \begin{subfigure}{0.40\textwidth}
         \centering
         \includegraphics[width=\textwidth]{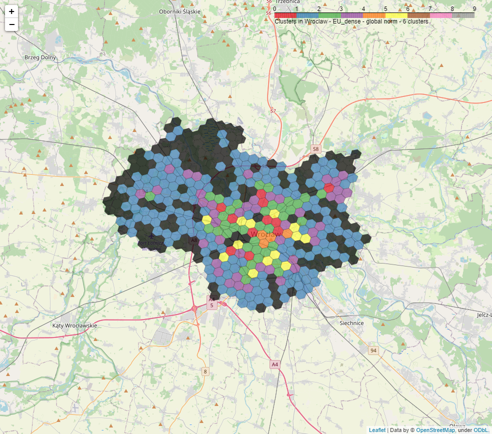}
         \caption{Wroclaw}
    \end{subfigure}
    \hspace{0.15\textwidth}%
    \begin{subfigure}{0.40\textwidth}
         \centering
         \includegraphics[width=\textwidth]{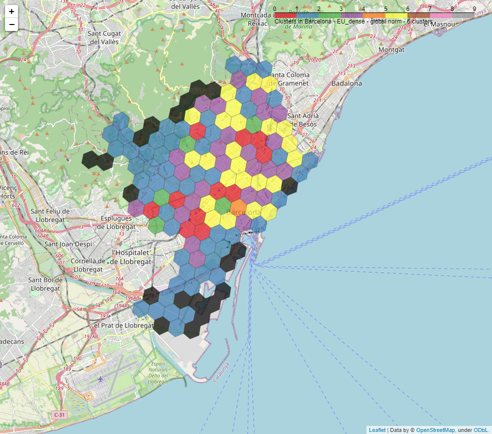}
         \caption{Barcelona}
    \end{subfigure}
    \begin{subfigure}{0.40\textwidth}
         \centering
         \includegraphics[width=\textwidth]{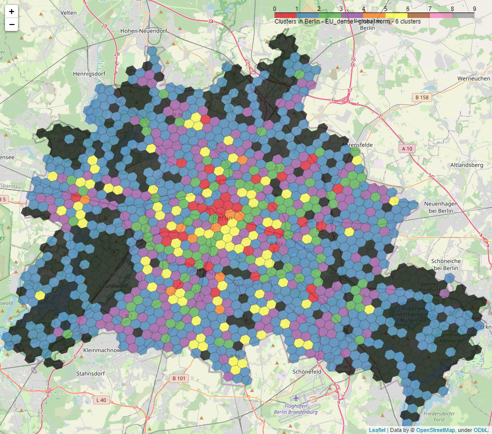}
         \caption{Berlin}
     \end{subfigure}
     \hspace{0.15\textwidth}%
     \begin{subfigure}{0.40\textwidth}
         \centering
         \includegraphics[width=\textwidth]{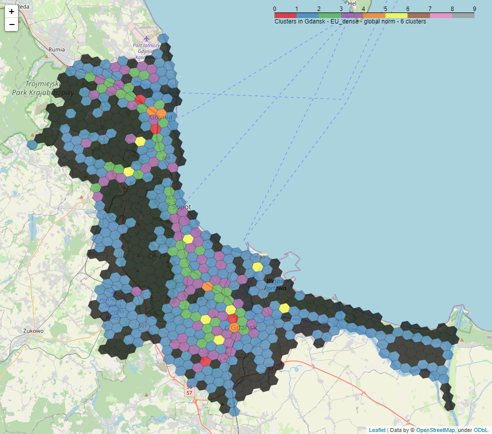}
         \caption{Tricity (Gdansk, Gdynia, Sopot)}
     \end{subfigure}
    \caption{Clusters visualization on maps - division for 6 clusters}
    \label{fig:6-clusters:maps}
\end{figure}

Another important insight on this matter can be noticed when analyzing Table \ref{tab:6-clusters:portion} with shared of each cluster in cities. There are many cities with an overwhelming number of regions from one of those clusters compared to another. It may suggest that this division starts to differentiate cities and not just regions, which is not the desired output. Another argument to back this theory up is visible on maps presented in Figure \ref{fig:6-clusters:maps}. When comparing Barcelona with Tricity it is clear that in one city cluster \textbf{2} (green) dominates while cluster \textbf{5} (yellow) is in majority in the other one. 

\begin{table}[h]
\centering
\caption{Percentage of regions in each cluster for all cities - division for 6 clusters}
\label{tab:6-clusters:portion}
\resizebox{\columnwidth}{!}{%
\begin{tabular}{@{}rrrrrrrrrrrrr@{}}
\toprule
 & \rotatebox{90}{Barcelona} & \rotatebox{90}{Berlin} & \rotatebox{90}{Brussels} & \rotatebox{90}{Bydgoszcz} & \rotatebox{90}{Gdansk} 
& \rotatebox{90}{Krakow} & \rotatebox{90}{Liepzig} & \rotatebox{90}{Poznan} & \rotatebox{90}{Prague} & \rotatebox{90}{Warsaw} 
& \rotatebox{90}{Vilnius} & \rotatebox{90}{Wroclaw} \\ \midrule
0    & 9.17      & 4.00   & 8.88     & 1.27      & 1.07   & 4.58   & 1.98  & 1.18   & 6.39  & 9.63     & 0.89  & 4.06    \\
1    & 45.00     & 53.80  & 30.84    & 68.79     & 68.18  & 64.18  & 65.87 & 66.93  & 49.50 & 45.99    & 76.19 & 59.78   \\
2    & 4.17      & 10.57  & 3.27     & 10.83     & 11.23  & 10.32  & 7.94  & 3.94   & 9.38  & 12.48    & 7.74  & 13.65   \\
3    & 20.83     & 21.97  & 17.29    & 17.83     & 16.31  & 16.33  & 22.62 & 22.44  & 22.36 & 19.25    & 11.90 & 16.24   \\
4    & 0.83      & 1.03   & 3.27     & 0.00      & 1.07   & 0.86   & 0.00  & 0.00   & 1.20  & 2.32     & 0.00  & 1.48    \\
5    & 20.00     & 8.62   & 36.45    & 1.27      & 2.14   & 3.72   & 1.59  & 5.51   & 11.18 & 10.34    & 3.27  & 4.80    \\ \bottomrule
\end{tabular}
}
\end{table}

\paragraph{7 clusters}

\begin{figure}[H]
     \centering
     \begin{subfigure}{0.49\textwidth}
         \centering
         \includegraphics[width=\textwidth]{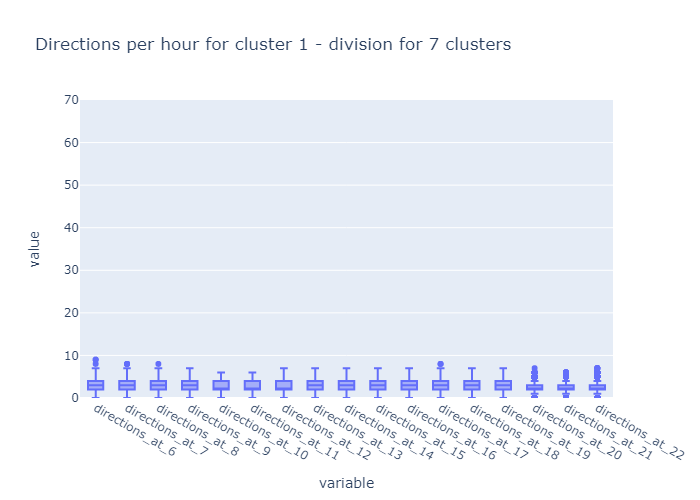}
         \caption{Cluster 1}
         \label{cluster7:1-directions}
         \label{cluster8:1-directions}
         \label{cluster9:1-directions}
     \end{subfigure}
     \begin{subfigure}{0.49\textwidth}
         \centering
         \includegraphics[width=\textwidth]{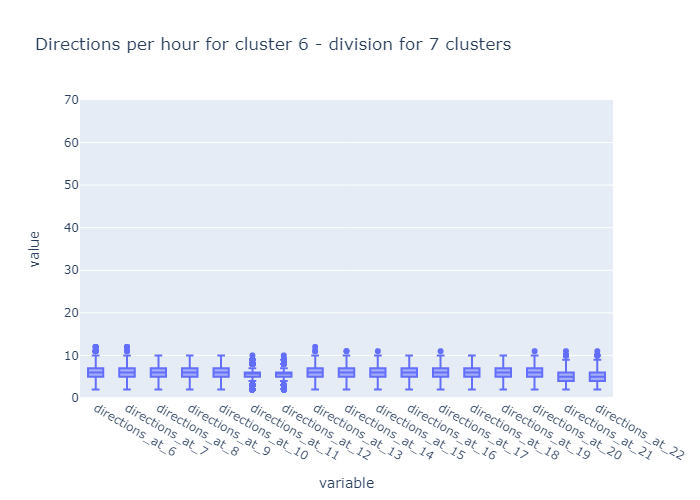}
         \caption{Cluster 6}
         \label{cluster7:6-directions}
         \label{cluster8:6-directions}
         \label{cluster9:6-directions}
     \end{subfigure}
     \begin{subfigure}{0.49\textwidth}
         \centering
         \includegraphics[width=\textwidth]{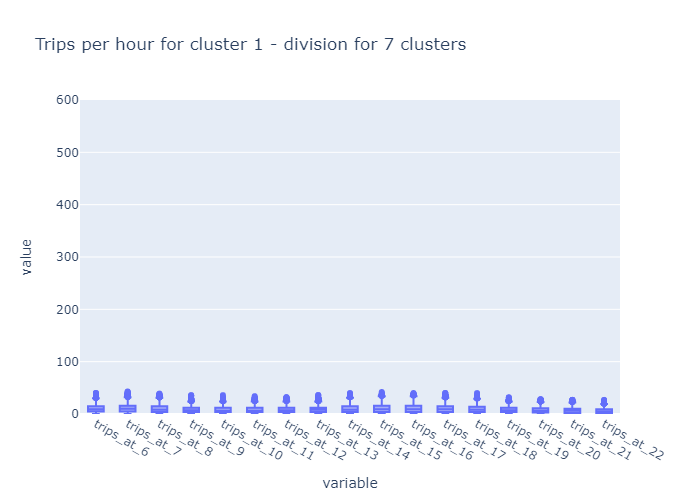}
         \caption{Cluster 1}
         \label{cluster7:1-trips}
         \label{cluster8:1-trips}
         \label{cluster9:1-trips}
     \end{subfigure}
     \begin{subfigure}{0.49\textwidth}
         \centering
         \includegraphics[width=\textwidth]{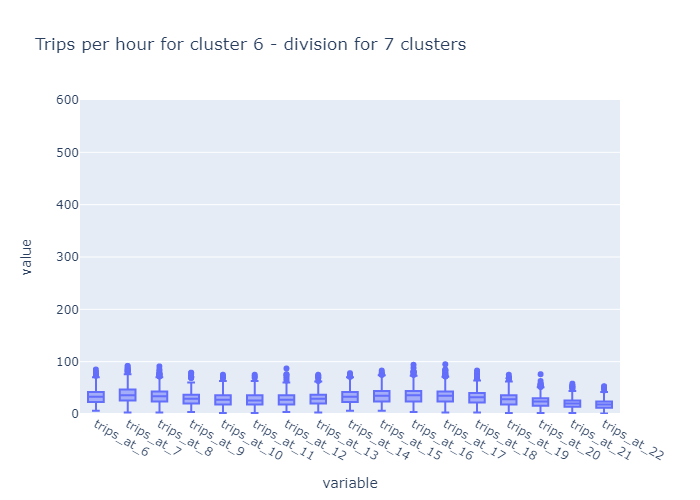}
         \caption{Cluster 6}
         \label{cluster7:6-trips}
         \label{cluster8:6-trips}
         \label{cluster9:6-trips}
     \end{subfigure}
    \caption{Trips and directions during a day for new clusters - division for 7 clusters}
    \label{fig:7-clusters:trips-directions}
\end{figure}

Plots for this level:

\begin{itemize}
    \item cluster 0 - \ref{cluster7:0-directions} and \ref{cluster7:0-trips}
    \item cluster 1 - \ref{cluster7:1-directions} and \ref{cluster7:1-trips}
    \item cluster 2 - \ref{cluster7:2-directions} and \ref{cluster7:2-trips}
    \item cluster 3 - \ref{cluster7:3-directions} and \ref{cluster7:3-trips}
    \item cluster 4 - \ref{cluster7:4-directions} and \ref{cluster7:4-trips}
    \item cluster 5 - \ref{cluster7:5-directions} and \ref{cluster7:5-trips}
    \item cluster 6 - \ref{cluster7:6-directions} and \ref{cluster7:6-trips}
\end{itemize}

\begin{figure}[H]
    \centering
    \includegraphics[width=0.75\textwidth]{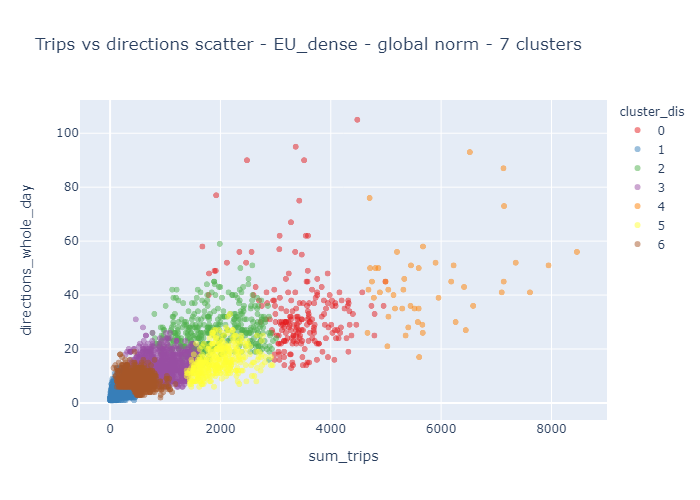}
    \caption{Aggregated features scatter plot - division for 7 clusters}
    \label{fig:7-clusters:aggregated}
\end{figure}

The introduction of the 7th cluster resulted in the first split in cluster describing suburban areas. This is in line with a previous observation that this cluster is significantly different from the others. Analysis of Figure \ref{fig:7-clusters:aggregated}, which contains a scatter plot of aggregated features, shows that this division happened along both directions and trips axis. This suggests that there were suburban areas with better connection with city center separated in cluster \textbf{6}.

\begin{figure}[H]
    \centering
    \begin{subfigure}{0.40\textwidth}
         \centering
         \includegraphics[width=\textwidth]{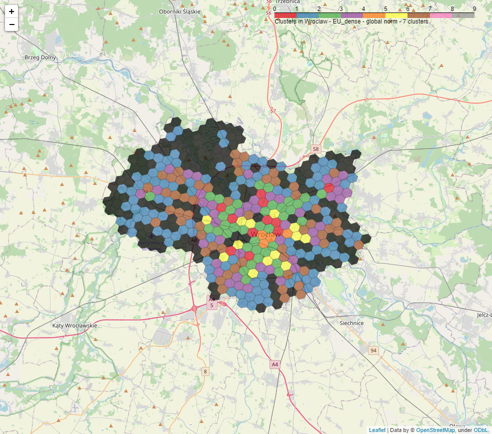}
         \caption{Wroclaw}
    \end{subfigure}
    \hspace{0.15\textwidth}%
    \begin{subfigure}{0.40\textwidth}
         \centering
         \includegraphics[width=\textwidth]{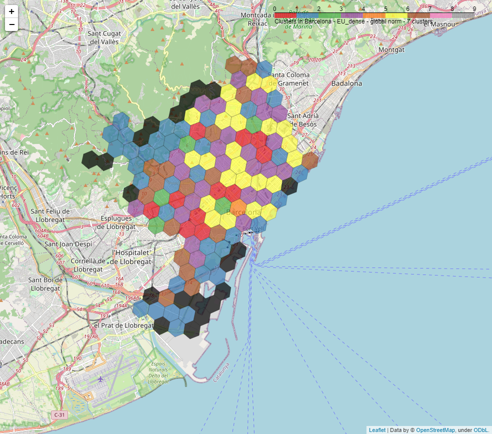}
         \caption{Barcelona}
    \end{subfigure}
    \begin{subfigure}{0.40\textwidth}
         \centering
         \includegraphics[width=\textwidth]{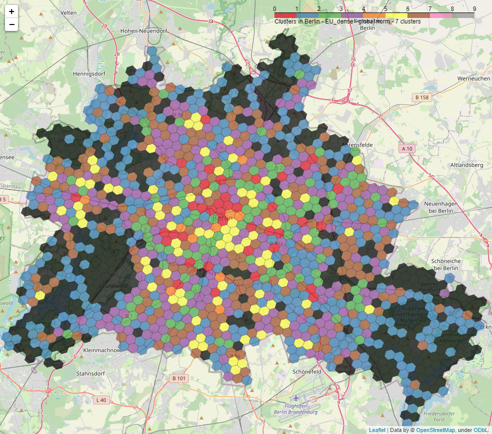}
         \caption{Berlin}
     \end{subfigure}
     \hspace{0.15\textwidth}%
     \begin{subfigure}{0.40\textwidth}
         \centering
         \includegraphics[width=\textwidth]{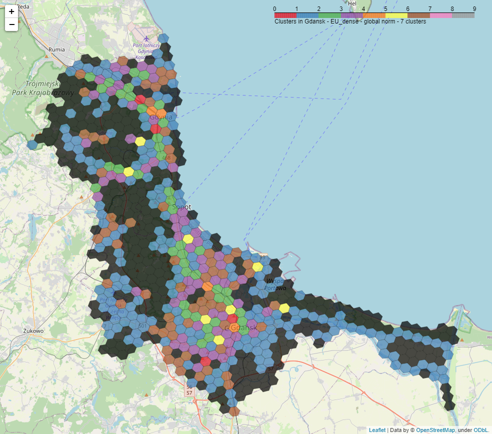}
         \caption{Tricity (Gdansk, Gdynia, Sopot)}
     \end{subfigure}
    \caption{Clusters visualization on maps - division for 7 clusters}
    \label{fig:7-clusters:maps}
\end{figure}

Figure \ref{fig:7-clusters:trips-directions} contains box plots for per-hour features. They show that the difference between those clusters is significant, almost two times greater are values in cluster \textbf{6} than in cluster \textbf{1}. This difference is greater than ones on previous levels, however, due to smaller values of features and relatively small variance they were not separated until this deep in the dendrogram. It may be an issue if relative differences were the most important aspect of typology since regions in cluster \textbf{6} seem to act as transition points for communication with the rest of a city. This theory is backed up by data in Table \ref{tab:7-clusters:portion}, which shows that in the majority of cities, there are more regions from cluster \textbf{1} than from cluster \textbf{6} which provides a better public transport availability. Analysis of maps on Figure \ref{fig:7-clusters:maps} also supports this thesis, since brown regions (cluster \textbf{6}) are in closer proximity to the city center in all four cities. 

\begin{table}[h]
\centering
\caption{Percentage of regions in each cluster for all cities - division for 7 clusters}
\label{tab:7-clusters:portion}
\resizebox{\columnwidth}{!}{%
\begin{tabular}{@{}rrrrrrrrrrrrr@{}}
\toprule
 & \rotatebox{90}{Barcelona} & \rotatebox{90}{Berlin} & \rotatebox{90}{Brussels} & \rotatebox{90}{Bydgoszcz} & \rotatebox{90}{Gdansk} 
& \rotatebox{90}{Krakow} & \rotatebox{90}{Liepzig} & \rotatebox{90}{Poznan} & \rotatebox{90}{Prague} & \rotatebox{90}{Warsaw} 
& \rotatebox{90}{Vilnius} & \rotatebox{90}{Wroclaw} \\ \midrule
0    & 9.17      & 4.00   & 8.88     & 1.27      & 1.07   & 4.58   & 1.98  & 1.18   & 6.39  & 9.63     & 0.89  & 4.06    \\
1    & 30.83     & 32.14  & 14.49    & 53.50     & 51.07  & 44.99  & 44.84 & 48.82  & 28.74 & 29.06    & 64.58 & 39.48   \\
2    & 4.17      & 10.57  & 3.27     & 10.83     & 11.23  & 10.32  & 7.94  & 3.94   & 9.38  & 12.48    & 7.74  & 13.65   \\
3    & 20.83     & 21.97  & 17.29    & 17.83     & 16.31  & 16.33  & 22.62 & 22.44  & 22.36 & 19.25    & 11.90 & 16.24   \\
4    & 0.83      & 1.03   & 3.27     & 0.00      & 1.07   & 0.86   & 0.00  & 0.00   & 1.20  & 2.32     & 0.00  & 1.48    \\
5    & 20.00     & 8.62   & 36.45    & 1.27      & 2.14   & 3.72   & 1.59  & 5.51   & 11.18 & 10.34    & 3.27  & 4.80    \\
6    & 14.17     & 21.66  & 16.36    & 15.29     & 17.11  & 19.20  & 21.03 & 18.11  & 20.76 & 16.93    & 11.61 & 20.30   \\ \bottomrule
\end{tabular}
}
\end{table}

\paragraph{8 clusters}

\begin{figure}[H]
     \centering
     \begin{subfigure}{0.49\textwidth}
         \centering
         \includegraphics[width=\textwidth]{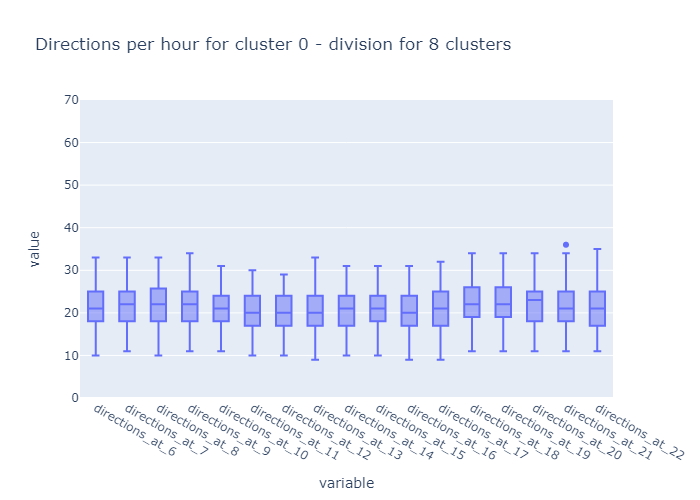}
         \caption{Cluster 0}
         \label{cluster8:0-directions}
         \label{cluster9:0-directions}
     \end{subfigure}
     \begin{subfigure}{0.49\textwidth}
         \centering
         \includegraphics[width=\textwidth]{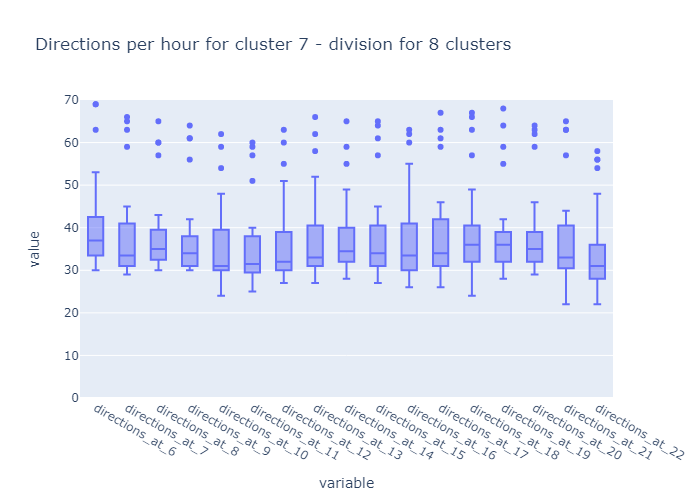}
         \caption{Cluster 7}
         \label{cluster8:7-directions}
         \label{cluster9:7-directions}
     \end{subfigure}
    \caption{Directions during a day for new clusters - division for 8 clusters}
    \label{fig:8-clusters:directions}
\end{figure}

\begin{figure}[H]
     \centering
     \begin{subfigure}{0.49\textwidth}
         \centering
         \includegraphics[width=\textwidth]{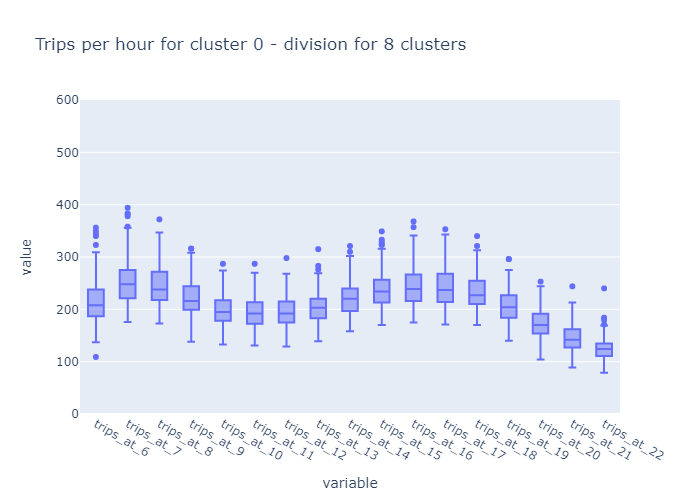}
         \caption{Cluster 0}
         \label{cluster8:0-trips}
         \label{cluster9:0-trips}
     \end{subfigure}
     \begin{subfigure}{0.49\textwidth}
         \centering
         \includegraphics[width=\textwidth]{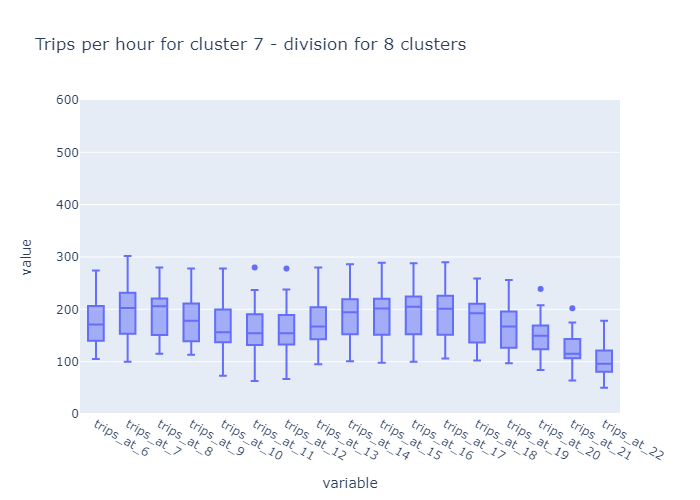}
         \caption{Cluster 7}
         \label{cluster8:7-trips}
         \label{cluster9:7-trips}
     \end{subfigure}
    \caption{Trips during a day for new clusters - division for 8 clusters}
    \label{fig:8-clusters:trips}
\end{figure}

Plots for this level:

\begin{itemize}
    \item cluster 0 - \ref{cluster8:0-directions} and \ref{cluster8:0-trips}
    \item cluster 1 - \ref{cluster8:1-directions} and \ref{cluster8:1-trips}
    \item cluster 2 - \ref{cluster8:2-directions} and \ref{cluster8:2-trips}
    \item cluster 3 - \ref{cluster8:3-directions} and \ref{cluster8:3-trips}
    \item cluster 4 - \ref{cluster8:4-directions} and \ref{cluster8:4-trips}
    \item cluster 5 - \ref{cluster8:5-directions} and \ref{cluster8:5-trips}
    \item cluster 6 - \ref{cluster8:6-directions} and \ref{cluster8:6-trips}
    \item cluster 7 - \ref{cluster8:7-directions} and \ref{cluster8:7-trips}
\end{itemize}

\begin{figure}[H]
    \centering
    \includegraphics[width=0.75\textwidth]{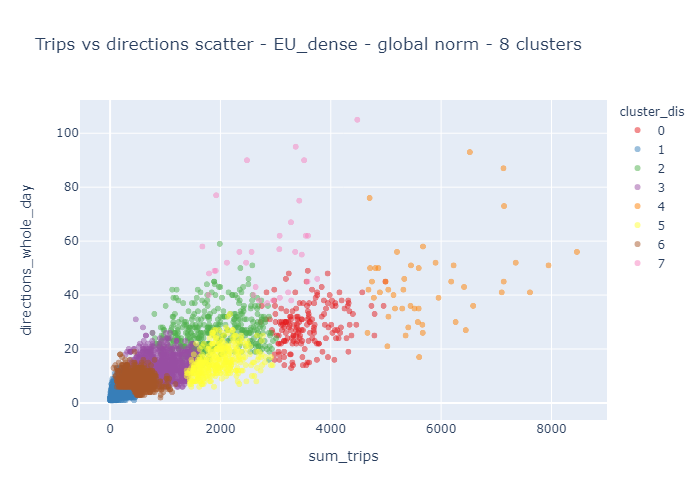}
    \caption{Aggregated features scatter plot - division for 8 clusters}
    \label{fig:8-clusters:aggregated}
\end{figure}

With the increase in the number of clusters to 8, another type of public transport hubs is separated from high volume cluster \textbf{0}. Previously cluster \textbf{4} was characterized with a greater sum of trips and this time, the new cluster \textbf{7} shows an increased number of directions. It is visualized in Figure \ref{fig:8-clusters:aggregated}. 

Looking at Figures \ref{fig:8-clusters:directions} and \ref{fig:8-clusters:trips} there are additional insight on this division available. The number of available directions is substantially bigger for the new cluster \textbf{7}, while the sum of trips is similar or quite smaller depending on the time of a day. This suggests that this cluster may be impacted by rail connections, which have a lower frequency of trips but reach more directions. This theory will be analyzed further in Section \ref{sec:typology-description}, where selected typology will be investigated in detail. 

Analysis of Table \ref{tab:8-clusters:portion} with shares of clusters for cities shows that cluster \textbf{7} is another example of cluster characteristic for some cities only. It is justifiable since such a combination of very high diversity and lower intensity is quite a unique combination in public transport. 

\begin{figure}[H]
    \centering
    \begin{subfigure}{0.40\textwidth}
         \centering
         \includegraphics[width=\textwidth]{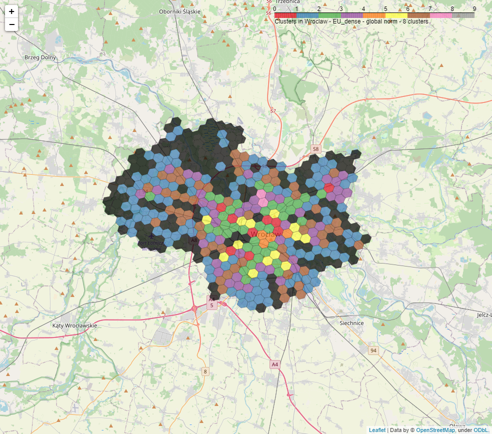}
         \caption{Wroclaw}
    \end{subfigure}
    \hspace{0.15\textwidth}%
    \begin{subfigure}{0.40\textwidth}
         \centering
         \includegraphics[width=\textwidth]{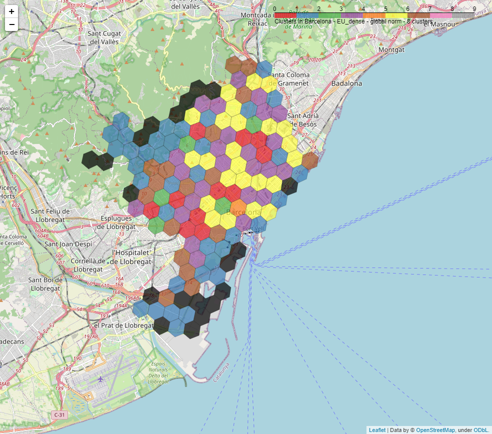}
         \caption{Barcelona}
    \end{subfigure}
    \begin{subfigure}{0.40\textwidth}
         \centering
         \includegraphics[width=\textwidth]{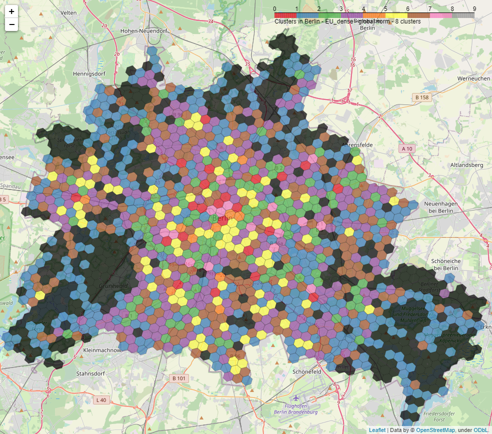}
         \caption{Berlin}
     \end{subfigure}
     \hspace{0.15\textwidth}%
     \begin{subfigure}{0.40\textwidth}
         \centering
         \includegraphics[width=\textwidth]{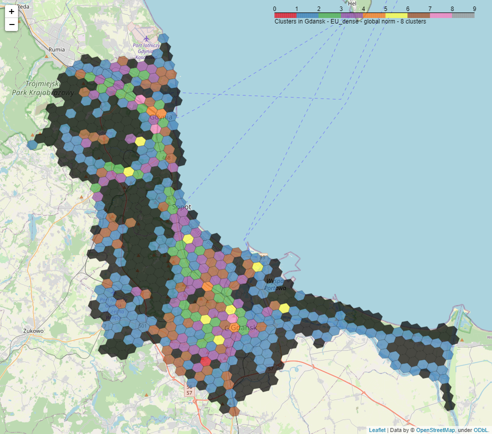}
         \caption{Tricity (Gdansk, Gdynia, Sopot)}
     \end{subfigure}
    \caption{Clusters visualization on maps - division for 8 clusters}
    \label{fig:8-clusters:maps}
\end{figure}

Visualizations on maps, presented in Figure \ref{fig:8-clusters:maps}, allow identifying those regions in all cities but Barcelona, which does not have regions of this characteristic. In Berlin and Tricity this cluster (marked in pink) matches with for example main rail station. In Wrocław it is in the northern part of a city, which provides communication for an entire northern part of a city, but is served mostly by buses, which have a lower frequency of trips than trams.  

\begin{table}[h]
\centering
\caption{Percentage of regions in each cluster for all cities - division for 8 clusters}
\label{tab:8-clusters:portion}
\resizebox{\columnwidth}{!}{%
\begin{tabular}{@{}rrrrrrrrrrrrr@{}}
\toprule
 & \rotatebox{90}{Barcelona} & \rotatebox{90}{Berlin} & \rotatebox{90}{Brussels} & \rotatebox{90}{Bydgoszcz} & \rotatebox{90}{Gdansk} 
& \rotatebox{90}{Krakow} & \rotatebox{90}{Liepzig} & \rotatebox{90}{Poznan} & \rotatebox{90}{Prague} & \rotatebox{90}{Warsaw} 
& \rotatebox{90}{Vilnius} & \rotatebox{90}{Wroclaw} \\ \midrule
0    & 9.17      & 2.36   & 8.88     & 1.27      & 0.27   & 4.01   & 0.79  & 1.18   & 5.99  & 8.91     & 0.89  & 3.32    \\
1    & 30.83     & 32.14  & 14.49    & 53.50     & 51.07  & 44.99  & 44.84 & 48.82  & 28.74 & 29.06    & 64.58 & 39.48   \\
2    & 4.17      & 10.57  & 3.27     & 10.83     & 11.23  & 10.32  & 7.94  & 3.94   & 9.38  & 12.48    & 7.74  & 13.65   \\
3    & 20.83     & 21.97  & 17.29    & 17.83     & 16.31  & 16.33  & 22.62 & 22.44  & 22.36 & 19.25    & 11.90 & 16.24   \\
4    & 0.83      & 1.03   & 3.27     & 0.00      & 1.07   & 0.86   & 0.00  & 0.00   & 1.20  & 2.32     & 0.00  & 1.48    \\
5    & 20.00     & 8.62   & 36.45    & 1.27      & 2.14   & 3.72   & 1.59  & 5.51   & 11.18 & 10.34    & 3.27  & 4.80    \\
6    & 14.17     & 21.66  & 16.36    & 15.29     & 17.11  & 19.20  & 21.03 & 18.11  & 20.76 & 16.93    & 11.61 & 20.30   \\
7    & 0.00      & 1.64   & 0.00     & 0.00      & 0.80   & 0.57   & 1.19  & 0.00   & 0.40  & 0.71     & 0.00  & 0.74    \\ \bottomrule
\end{tabular}
}
\end{table}

\paragraph{9 clusters}

\begin{figure}[H]
     \centering
     \begin{subfigure}{0.49\textwidth}
         \centering
         \includegraphics[width=\textwidth]{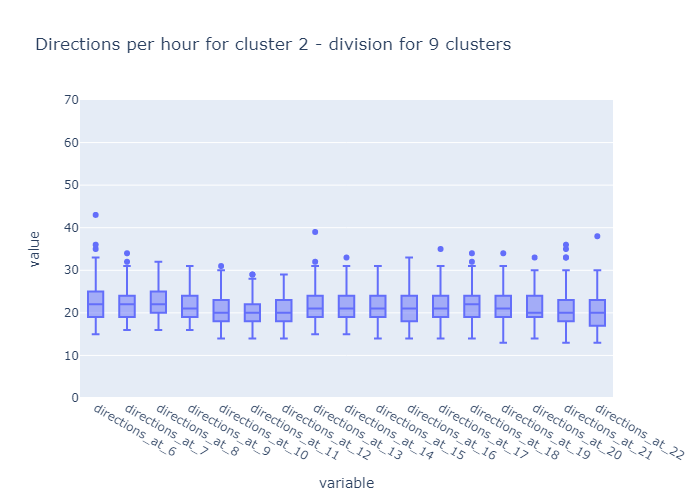}
         \caption{Cluster 2}
         \label{cluster9:2-directions}
     \end{subfigure}
     \begin{subfigure}{0.49\textwidth}
         \centering
         \includegraphics[width=\textwidth]{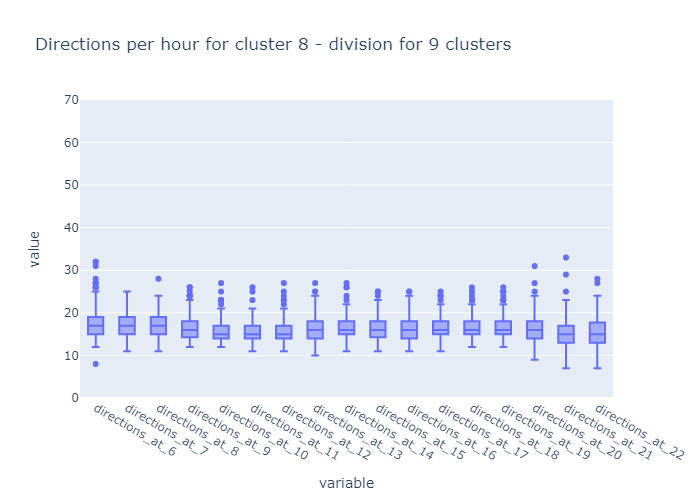}
         \caption{Cluster 8}
         \label{cluster9:8-directions}
     \end{subfigure}
    \caption{Directions during a day for new clusters - division for 9 clusters}
    \label{fig:9-clusters:directions}
\end{figure}

\begin{figure}[H]
     \centering
     \begin{subfigure}{0.49\textwidth}
         \centering
         \includegraphics[width=\textwidth]{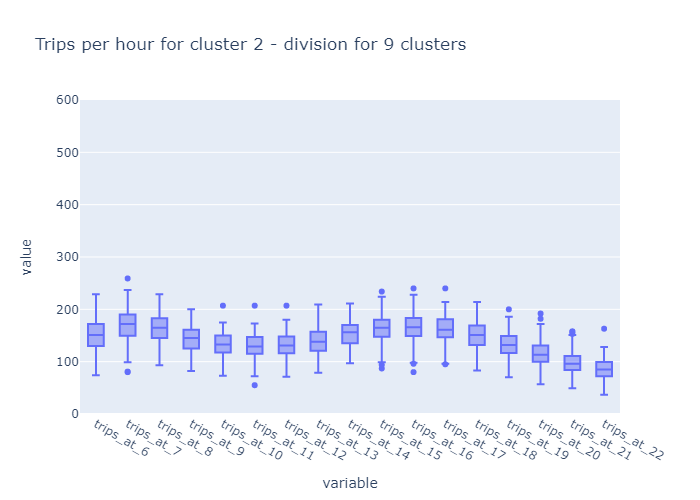}
         \caption{Cluster 2}
         \label{cluster9:2-trips}
     \end{subfigure}
     \begin{subfigure}{0.49\textwidth}
         \centering
         \includegraphics[width=\textwidth]{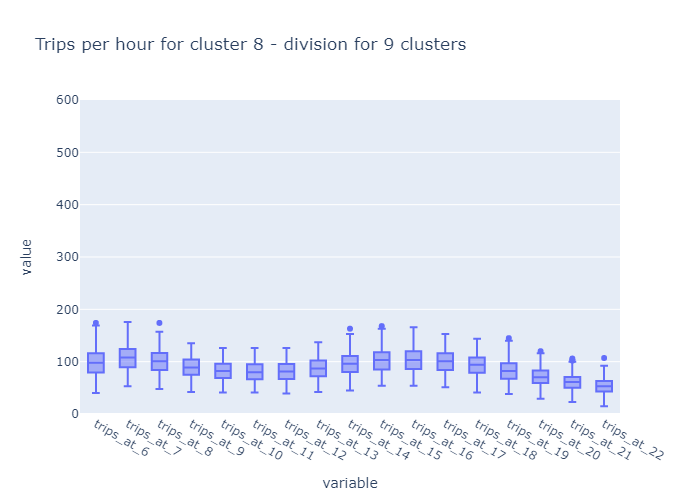}
         \caption{Cluster 8}
         \label{cluster9:8-trips}
     \end{subfigure}
    \caption{Trips during a day for new clusters - division for 9 clusters}
    \label{fig:9-clusters:trips}
\end{figure}

Plots for this level:

\begin{itemize}
    \item cluster 0 - \ref{cluster9:0-directions} and \ref{cluster9:0-trips}
    \item cluster 1 - \ref{cluster9:1-directions} and \ref{cluster9:1-trips}
    \item cluster 2 - \ref{cluster9:2-directions} and \ref{cluster9:2-trips}
    \item cluster 3 - \ref{cluster9:3-directions} and \ref{cluster9:3-trips}
    \item cluster 4 - \ref{cluster9:4-directions} and \ref{cluster9:4-trips}
    \item cluster 5 - \ref{cluster9:5-directions} and \ref{cluster9:5-trips}
    \item cluster 6 - \ref{cluster9:6-directions} and \ref{cluster9:6-trips}
    \item cluster 7 - \ref{cluster9:7-directions} and \ref{cluster9:7-trips}
    \item cluster 8 - \ref{cluster9:8-directions} and \ref{cluster9:8-trips}
\end{itemize}

\begin{figure}[H]
    \centering
    \includegraphics[width=0.75\textwidth]{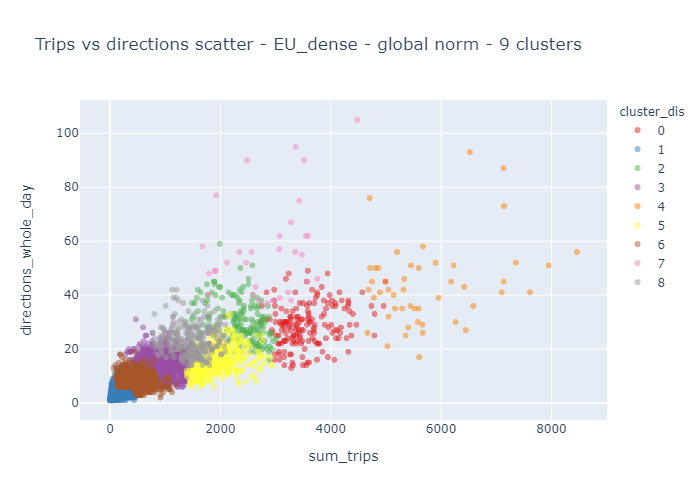}
    \caption{Aggregated features scatter plot - division for 9 clusters}
    \label{fig:9-clusters:aggregated}
\end{figure}

Introduction of 9th cluster creates another division in "medium" cluster \textbf{2}. As mentioned before, when this cluster was first divided, it should not be divided anymore as it does not bring valuable insight in terms of typology definition. This time the difference is more clear in both directions and trips (see Figures \ref{fig:9-clusters:aggregated}, \ref{fig:9-clusters:directions}, \ref{fig:9-clusters:trips}) but is not important in terms of identifying unique types of regions in terms of public transport availability. Therefore, a final typology will contain 8 clusters.

Following on analysis of this division for 9 clusters, new cluster \textbf{8} seems to mean areas with quite high-quality public transport. Looking at shares in Table \ref{tab:9-clusters:portion}, it is present in all cities in a significant amount, but at the same time, it is almost equally more present than cluster \textbf{2} form which it was separated. This is another sign, that this division is not important in terms of inter-city typology. 

\begin{figure}[H]
    \centering
    \begin{subfigure}{0.40\textwidth}
         \centering
         \includegraphics[width=\textwidth]{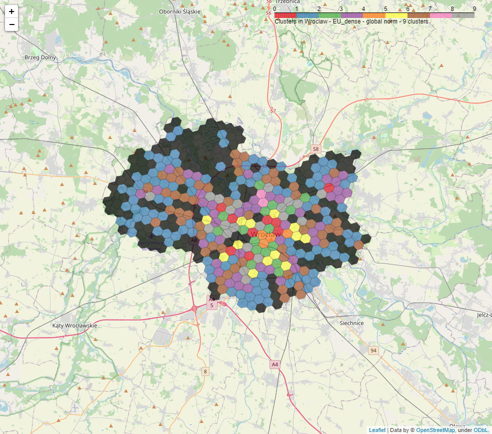}
         \caption{Wroclaw}
    \end{subfigure}
    \hspace{0.15\textwidth}%
    \begin{subfigure}{0.40\textwidth}
         \centering
         \includegraphics[width=\textwidth]{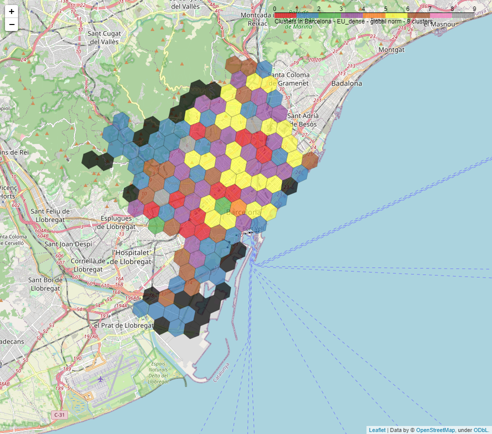}
         \caption{Barcelona}
    \end{subfigure}
    \begin{subfigure}{0.40\textwidth}
         \centering
         \includegraphics[width=\textwidth]{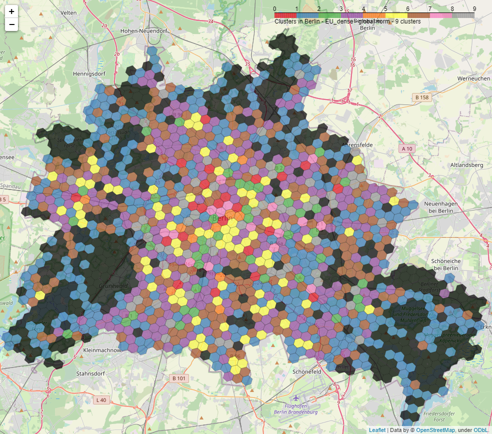}
         \caption{Berlin}
     \end{subfigure}
     \hspace{0.15\textwidth}%
     \begin{subfigure}{0.40\textwidth}
         \centering
         \includegraphics[width=\textwidth]{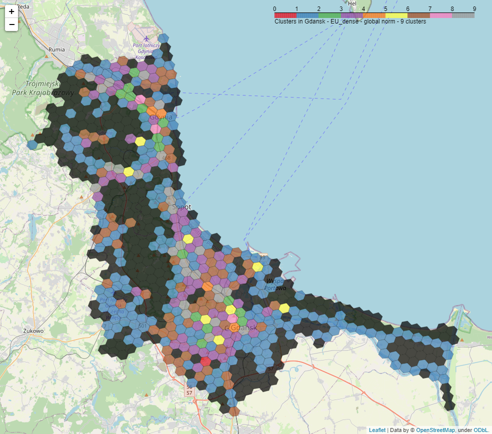}
         \caption{Tricity (Gdansk, Gdynia, Sopot)}
     \end{subfigure}
    \caption{Clusters visualization on maps - division for 9 clusters}
    \label{fig:9-clusters:maps}
\end{figure}

Analysis of maps (Figure \ref{fig:9-clusters:maps}) suggest that it replaced most of green (cluster \textbf{2}) regions and almost all are close to the remaining ones. That is another argument against including it in typology definition and analyzing dendrogram any deeper. Still, it provides a justifiable split in terms of defined cities, which is an argument in favor of the proposed solution for regions embedding and clustering.

\begin{table}[h]
\centering
\caption{Percentage of regions in each cluster for all cities - division for 9 clusters}
\label{tab:9-clusters:portion}
\resizebox{\columnwidth}{!}{%
\begin{tabular}{@{}rrrrrrrrrrrrr@{}}
\toprule
 & \rotatebox{90}{Barcelona} & \rotatebox{90}{Berlin} & \rotatebox{90}{Brussels} & \rotatebox{90}{Bydgoszcz} & \rotatebox{90}{Gdansk} 
& \rotatebox{90}{Krakow} & \rotatebox{90}{Liepzig} & \rotatebox{90}{Poznan} & \rotatebox{90}{Prague} & \rotatebox{90}{Warsaw} 
& \rotatebox{90}{Vilnius} & \rotatebox{90}{Wroclaw} \\ \midrule
0    & 9.17      & 2.36   & 8.88     & 1.27      & 0.27   & 4.01   & 0.79  & 1.18   & 5.99  & 8.91     & 0.89  & 3.32    \\
1    & 30.83     & 32.14  & 14.49    & 53.50     & 51.07  & 44.99  & 44.84 & 48.82  & 28.74 & 29.06    & 64.58 & 39.48   \\
2    & 1.67      & 4.31   & 1.87     & 1.91      & 4.01   & 4.58   & 1.59  & 1.97   & 3.59  & 5.35     & 2.98  & 5.90    \\
3    & 20.83     & 21.97  & 17.29    & 17.83     & 16.31  & 16.33  & 22.62 & 22.44  & 22.36 & 19.25    & 11.90 & 16.24   \\
4    & 0.83      & 1.03   & 3.27     & 0.00      & 1.07   & 0.86   & 0.00  & 0.00   & 1.20  & 2.32     & 0.00  & 1.48    \\
5    & 20.00     & 8.62   & 36.45    & 1.27      & 2.14   & 3.72   & 1.59  & 5.51   & 11.18 & 10.34    & 3.27  & 4.80    \\
6    & 14.17     & 21.66  & 16.36    & 15.29     & 17.11  & 19.20  & 21.03 & 18.11  & 20.76 & 16.93    & 11.61 & 20.30   \\
7    & 0.00      & 1.64   & 0.00     & 0.00      & 0.80   & 0.57   & 1.19  & 0.00   & 0.40  & 0.71     & 0.00  & 0.74    \\
8    & 2.50      & 6.26   & 1.40     & 8.92      & 7.22   & 5.73   & 6.35  & 1.97   & 5.79  & 7.13     & 4.76  & 7.75    \\ \bottomrule
\end{tabular}
}
\end{table}

\subsection{Typology description}\label{sec:typology-description}

This section will conclude research in the direction of typology identification, which was presented in detail in the previous section. A resulting typology will be defined on multiple levels of precision. 

\paragraph{Level 1:}

Based on an in-depth analysis of clustering results, a number of \textbf{2} clusters appears to define a very clear distinction between two types of regions. An analysis already identified them as suburban areas separated from a city center. 

A suburban area can be defined as areas with a limited variety of public transport. They often only have connections to the city center or other most important hubs, hence the average number of directions is very low. The number of trips at each hour is more diversified and outliers reach out to almost 100 trips per hour. A small trend towards an M-shaped trips plot is observed (M-shape is a result of an increased number of trips in rush hours). 

The other cluster, which can be called a city center area, has significantly higher diversity. As more types of public transport offer are added to this typology, most of them will be separated from this type. Therefore, more precise, numerical analysis cannot be conducted just yet. 

\begin{figure}[t]
    \centering
    \begin{subfigure}{0.40\textwidth}
         \centering
         \includegraphics[width=\textwidth]{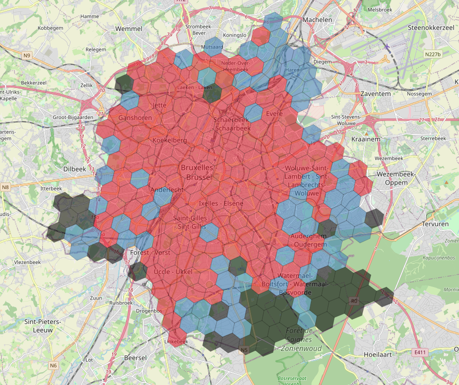}
         \caption{Brussels}
    \end{subfigure}
    \hspace{0.15\textwidth}%
    \begin{subfigure}{0.40\textwidth}
         \centering
         \includegraphics[width=\textwidth]{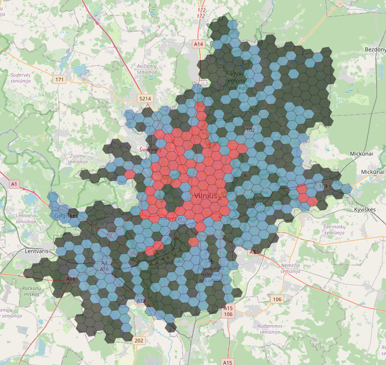}
         \caption{Vilnius}
    \end{subfigure}
    \caption{Comparison of cities with max and min share of suburban areas}
    \label{fig:2-typology:wilno-bruksela}
\end{figure}

This level of the hierarchy can be useful in a variety of comparisons. Firstly, the overall quality of public transport can be assessed based on just a ratio of those two clusters. This of course depends on administrative boundaries, and cities with boundaries closer to the city center will benefit from this. Even though, it is a good indicator of how well a given area is served with public transport. An example of this can be seen in Figure \ref{fig:2-typology:wilno-bruksela}, which compares the city with the lowest (Brussels) and highest (Vilnius) share of suburban areas. Partially this difference is impacted by boundaries placement, however, in Brussels, there are no \textit{holes} in the red area. 

\begin{figure}[h]
    \centering
    \begin{subfigure}{0.40\textwidth}
         \centering
         \includegraphics[width=\textwidth]{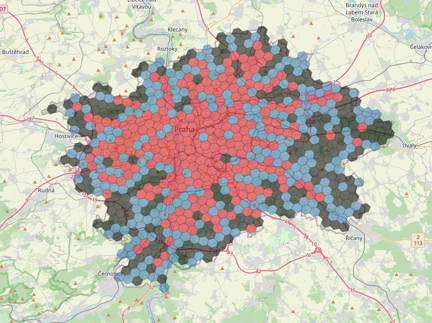}
         \caption{Prague}
         \label{fig:2-typology:praga}
    \end{subfigure}
    \hspace{0.15\textwidth}%
    \begin{subfigure}{0.40\textwidth}
         \centering
         \includegraphics[width=\textwidth]{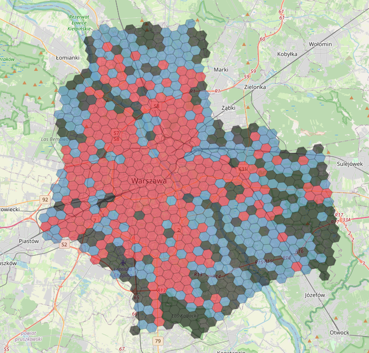}
         \caption{Warsaw}
         \label{fig:2-typology:warszawa}
    \end{subfigure}
    \caption{Usage of 1st level typology in analysis of cities}
\end{figure}

Another measure of public transport offer quality across an entire city can be assessed by judging how far do the red regions span in the city. In Figure \ref{fig:2-typology:praga} this is illustrated using the city of Prague, which has a very good connection to the city borders in multiple directions. Cities with a ring of suburban areas can be treated as ones with poor public transport offer for citizens living further from a city center (an example can be Wroclaw, which was presented earlier in Figure \ref{fig:2-clusters:maps}).

This level of typology can also serve as a good way of comparing different areas of a city. An example is presented in Figure \ref{fig:2-typology:warszawa}, which shows a clear inequality in public transport quality on left- and right-bank part of a city. 

Due to the fact, that there are only 2 clusters, no obvious similarities can be observed between particular regions in each cluster. This will be possible to observe when smaller clusters will be separated on the next levels of this hierarchical typology. 

\paragraph{Level 2:}

The second level of an obtained typology can be described by clustering for \textbf{4} clusters. Introduction of 5th cluster introduced splits based on only one type of features, which is a beginning of the new type of divisions and therefore should be included in the next level of typology. Meanwhile, four clusters provide a good differentiation between regions in terms of general public transport availability.

Firstly, the suburban areas remain unchanged from the previous level of typology, therefore, this type will not be analyzed in detail for the second time. On the opposite side of public transport quality, a group of regions with very high public transport availability. This type represents transportation hubs, which on this level of typology are not differentiated between local hubs in districts and city-scale transportation hubs. Those regions have an average number of directions available of over 25 per hour. This means, that multiple routes cross in these regions making them great transfer points when traveling for far distances in a city. When it comes to the number of trips through those regions every hour, a clear M-pattern occurs, with an average number of trips in rush hours above 200 per hour. 

The other two types of public transport offer reside in between two described above. They differ in both the number of trips per hour and directions. Cluster \textbf{2}, which is associated with better availability of public transport, has also more variance in terms of those values in each hour. Moreover, regions from this cluster show a more noticeable M-pattern, which is a result of an increased amount of routes (determined by more directions available), which increase the frequency of their trips in rush hours. 

\begin{figure}[t]
    \centering
    \begin{subfigure}{0.45\textwidth}
         \centering
         \includegraphics[width=\textwidth]{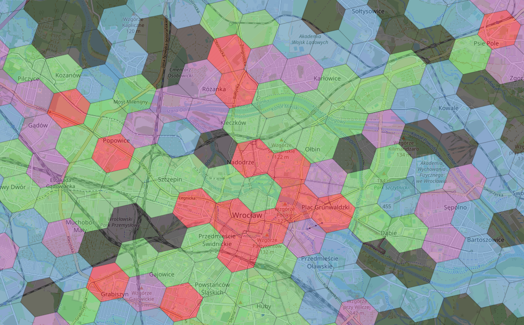}
         \caption{Wroclaw}
         \label{fig:4-typology:wroclaw}
    \end{subfigure}
    \begin{subfigure}{0.45\textwidth}
         \centering
         \includegraphics[width=\textwidth]{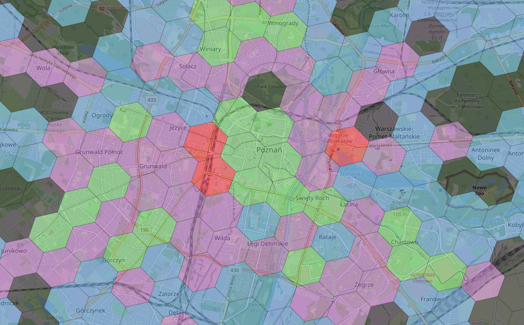}
         \caption{Poznan}
         \label{fig:4-typology:poznan}
    \end{subfigure}
    \caption{Hubs location and difference in public transport availability between Wroclaw and Poznan}
    \label{fig:4-typology:hubs}
\end{figure}

This level of typology, allows for similarities search across the cities, however only in terms of hubs identification. This can be seen in Figure \ref{fig:4-typology:hubs}, which compares centers of Wroclaw and Poznan. Red clusters witch indicate transportation hubs show which areas in both cities share this role. In Poznan, those are areas near the main railway station and big intersections in the eastern part of the city. Wroclaw has significantly more of those regions distributed around the city. Moreover, regions of FAT, main railway station, Regan Roundabout, or Nadodrze rail station are important hubs in Wroclaw's public transport system. This is a strong argument, that the proposed method can extract knowledge about the public transport system from raw timetables data. 

Another way of utilizing this level of typology in public transport quality analysis is to compare the ratio of regions in clusters \textbf{2} and \textbf{3}, which are associated with \textbf{medium} types in this typology. An example can be seen when comparing city center of Wroclaw (Figure \ref{fig:4-typology:wroclaw}) and Poznan (Figure \ref{fig:4-typology:poznan}). In Wroclaw, green (better transport) regions are in majority, while Poznan is dominated by purple regions. This can be an easy way to compare public transport availability between centers of cities.

\paragraph{Level 3:}

The last level of typology can be based on the division for \textbf{8} clusters. As shown in the previous section, the addition of the 9th cluster introduced more precise splits for regions with \textit{medium} accessibility of public transport. Therefore, this 3rd level of typology should be defined before this split, hence a selection of 8 clusters. Obtained types are shortly characterized below. To maintain compatibility with level 2 of this typology, they are organized according to types from level 2.

\begin{itemize}
    \item suburban areas
    
    \begin{itemize}
        \item cluster \textbf{1} - regions with poor public transport availability
        \item cluster \textbf{6} - parts of suburbs with better connection to the rest of the city - some sort of entry points for suburbs
    \end{itemize}
    
    \item hubs
    
    \begin{itemize}
        \item cluster \textbf{0} - local transportation hubs/regions with very high public transport availability.
        \item cluster \textbf{4} - main transportation hubs with very high throughput
        \item cluster \textbf{7} - transportation hubs with lower number of trips - related to railway stations or big transfer points
    \end{itemize}
    
    \item mid-city (combines two types form level 2)
        
    \begin{itemize}
        \item cluster \textbf{2} - regions with high diversity of directions available, probably regions with intersections of public transport routes
        \item cluster \textbf{3} - regions from mid-city with slightly better public transport offer than in the suburbs. Those can serve as regions inside the city in between major stops or as suburbs with really good offer
        \item cluster \textbf{5} - mid-city regions with lower diversity of public transport offer
    \end{itemize}
\end{itemize}

This level provides a good opportunity to analyze similarities in obtained clusters and how does it translate to the real world. Shares of particular clusters are less informative compared to level 2 of typology, because of increased diversity in types and the fact that it is more difficult to say that one cluster relates to better public transport offer than others.

\begin{figure}[h]
    \centering
    \begin{subfigure}{0.3\textwidth}
         \centering
         \includegraphics[width=\textwidth]{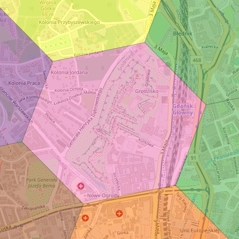}
         \caption{Gdansk}
         \label{fig:8-typology:rail-gdansk}
    \end{subfigure}
    \begin{subfigure}{0.3\textwidth}
         \centering
         \includegraphics[width=\textwidth]{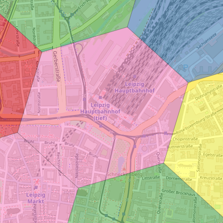}
         \caption{Liepzig}
         \label{fig:8-typology:rail-lipsk}
    \end{subfigure}
    \begin{subfigure}{0.3\textwidth}
         \centering
         \includegraphics[width=\textwidth]{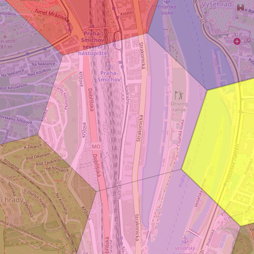}
         \caption{Prague}
         \label{fig:8-typology:rail-praha}
    \end{subfigure}
    \begin{subfigure}{0.3\textwidth}
         \centering
         \includegraphics[width=\textwidth]{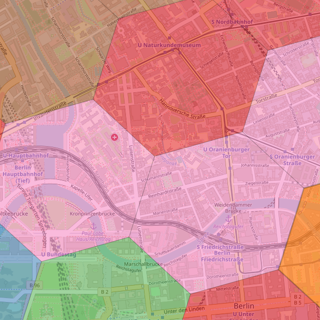}
         \caption{Berlin Hauptbahnhof}
         \label{fig:8-typology:rail-berlin-main}
    \end{subfigure}
    \begin{subfigure}{0.3\textwidth}
         \centering
         \includegraphics[width=\textwidth]{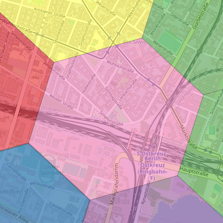}
         \caption{Berlin Ostbahnhof}
         \label{fig:8-typology:rail-berlin-ost}
    \end{subfigure}
    \begin{subfigure}{0.3\textwidth}
         \centering
         \includegraphics[width=\textwidth]{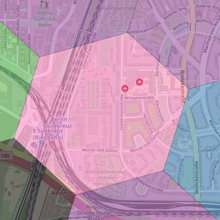}
         \caption{Berlin Sudbahnhof}
         \label{fig:8-typology:rail-berlin-sud}
    \end{subfigure}
    \caption{Railway stations identified by cluster 7}
    \label{fig:8-typology:rail}
\end{figure}

\subparagraph{Railway stations}

The first observation is connected with cluster \textbf{7}, which was described in a typology as hubs with a lower number of trips. After an analysis of all cities, it was identified to contain most of the railway stations. This was the case only for cities that included railways in GTFS timetables, which was predictable. Those railway stations are presented in Figure \ref{fig:8-typology:rail}. In the first three cities, there were only single train stations which were classified to this type, while in Berlin almost all fell into this category. This shows, that railway connections in Berlin are developed. Also, as it is shown in Figures \ref{fig:8-typology:rail-berlin-ost} and \ref{fig:8-typology:rail-berlin-sud}, some of those rail stations serve as intersection which increases number of directions available even more. 

\begin{figure}[h]
    \centering
    \begin{subfigure}{0.3\textwidth}
         \centering
         \includegraphics[width=\textwidth]{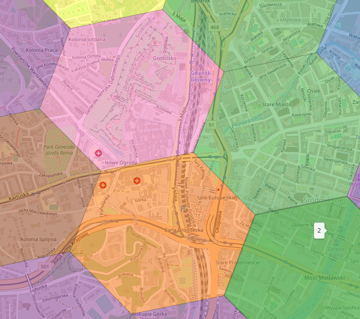}
         \caption{Gdansk}
    \end{subfigure}
    \hspace{0.15\textwidth}%    
    \begin{subfigure}{0.3\textwidth}
         \centering
         \includegraphics[width=\textwidth]{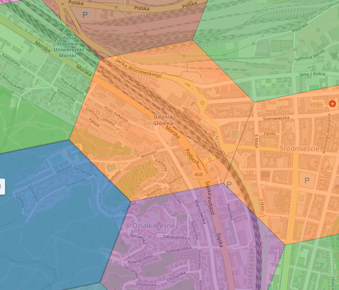}
         \caption{Gdynia}
    \end{subfigure}
    \caption{Influence of micro regions selection}
    \label{fig:8-typology:rail-regions-limitation}
\end{figure}

This analysis also showed that this method remains dependent on the method of region selection. As presented in Figure \ref{fig:8-typology:rail-regions-limitation}, railway station in Gdynia is not in the same cluster. The probable reason would be a placement of hexagons which included public transport stops near the station in Gdynia to the same region as the station. This is not the case in Gdansk. However, this is still a difference in public transport availability, so it should not be considered as a flaw of the proposed method. 

\subparagraph{Main public transportation hubs}

\begin{figure}[H]
    \centering
    \begin{subfigure}{0.52\textwidth}
         \centering
         \includegraphics[width=\textwidth]{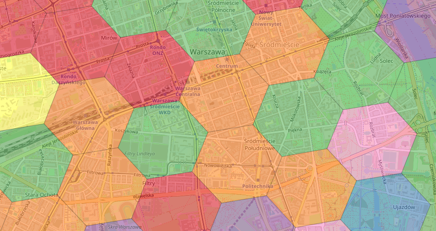}
         \caption{Warsaw}
    \end{subfigure}
    \begin{subfigure}{0.28\textwidth}
         \centering
         \includegraphics[width=\textwidth]{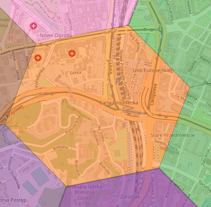}
         \caption{Gdansk}
    \end{subfigure}
    \begin{subfigure}{0.51\textwidth}
         \centering
         \includegraphics[width=\textwidth]{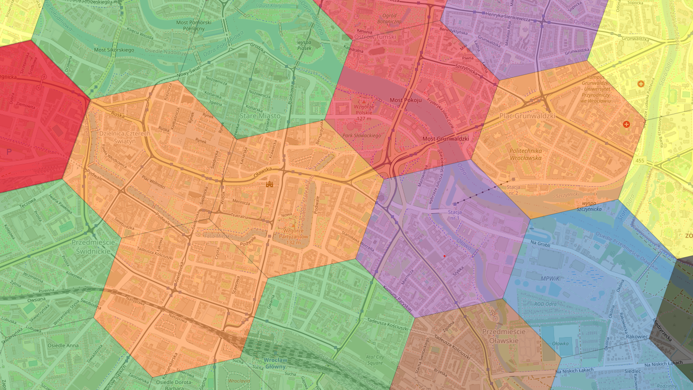}
         \caption{Wroclaw}
    \end{subfigure}
    \begin{subfigure}{0.29\textwidth}
         \centering
         \includegraphics[width=\textwidth]{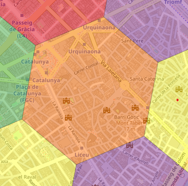}
         \caption{Barcelona}
    \end{subfigure}
    \begin{subfigure}{0.8\textwidth}
         \centering
         \includegraphics[width=\textwidth]{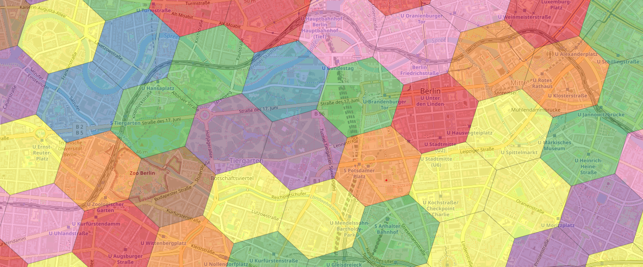}
         \caption{Berlin}
    \end{subfigure}
    \caption{Main hubs identified by cluster 4}
    \label{fig:8-typology:hubs}
\end{figure}

Another type from described typology which can be compared across the cities is the one identified in cluster \textbf{4}. It contains big transportation hubs with a high frequency of trips. Figure \ref{fig:8-typology:hubs} presents examples of major hubs identified in this cluster. Following the order from the figure they are:
\begin{itemize}
    \item Warsaw: Warszawa Centralna - main railway station and public transport hub, Warszawa Glowna which serves for the suburban railway, major intersections in the city center.
    \item Gdansk - one of the main intersections of public transport south from the main railway station.
    \item Wroclaw - all major tram and bus transfer stations are included in those regions (Capitol, Gallery, Regan's Roundabout, main railway station).
    \item Barcelona - the only region of this type includes La Rambla which is one of the major streets in the city.
    \item Berlin - two major metro stations (Alexanderplatz and Potsdamer Platz) and Zoo station which is the main hub in Western Berlin.
\end{itemize}

\subsection{Research questions - answers}

Multiple research questions were formulated at the beginning of this experiment. All of them were answered in previous sections, however, for clarity, they will be summarized in this section. 

\paragraph{RQ1}: Can regions in a city be differentiated based on public transport timetables?

Yes, regions in the city are well-differentiated based only on static timetable data from o single region even without precise information about travel times and connections with other regions. Those will be analyzed in future works as described in \ref{sec:future}.

Multiple divisions in hierarchical clustering were analyzed and they were providing a clear differentiation between regions, which was confirmed when analyzing how are they distributed in the cities, for which public transport characteristic is known to the author.  

\paragraph{RQ2}: Is it possible to define a typology based on those differences?

Yes, as shown in \ref{sec:typology-description}, multiple typologies can be defined based on obtained clusters. Those types of public transport offer can be also matched with specific regions of a city (for example suburban areas) or with regions of similar function (hubs of public transport).

\paragraph{RQ3}: Will usage of hierarchical clustering result in a multi-level typology?

Yes, as presented in \ref{sec:typology-description}, the three levels of the defined typology are closely interrelated and represent further refinements of the types of public transport offers. This can be seen in hubs, which were defined as separate types on level 2 and divides into three kinds on level 3.

\paragraph{RQ4}: Can this typology differentiate between whole cities and big parts of cities? 

Yes, especially with levels 1 and 2 which provide types which can be compared in terms of which type resembles better availability of public transport. As presented in Figure \ref{fig:2-typology:wilno-bruksela}, the share of regions with a suburban type of transport can provide a measure to differentiate cities. Furthermore, as those types can be ranked in terms of offer quality, it can serve as a measure between beg areas in the city, which was presented on the example of Warsaw, and differences between both sides of the river. Level 2 of typology provides a good differentiation inside regions of the city center by providing two types of public transport with different availability. This was used to compare Wroclaw and Poznan in Figure \ref{fig:4-typology:hubs}.

\paragraph{RQ5}: Will this typology be able to extract some types of regions which are repeated across the cities?

Yes, regions of some unique characteristics were extracted when defining level 3 of the typology. This is best illustrated by the types that have the smallest representation, like transportation hubs of different types. For example, the proposed typology was able to describe an offer type typical for suburban railway and included train stations from multiple cities (see \ref{fig:8-typology:rail}). It also was able to extract differences in the placement of other public transport types near train stations (see a comparison of Gdansk and Gdynia in Figure \ref{fig:8-typology:rail-regions-limitation}).

Another example is high-intensity transportation hubs identified in some cities. All of those match with regions where main transfer points in public transport networks occur. They are also not present in every city, which is another measure to compare cities.

\section{Experiment 2: 48 cities}

This section describes the second experiment which involved extending a set of cities tested using the proposed solution. This test will verify how well does the proposed solution scale to a larger number of cities and how does it affect the typology defined in the previous section. Another advantage is that it will show how representative the previously selected set of cities was.

\subsection{Research questions}

After the previous experiment results were analyzed, additional research questions were formulated to provide a more in-depth analysis of the proposed solution for typology identification. 

\paragraph{RQ6}: Will the character of differences change with the addition of more cities?

\paragraph{RQ7}: Will the typology be applicable for more cities?

\paragraph{RQ8}: What types of public transport offer was not present in previously selected 12 cities?

\subsection{Data}

To extend the research on the proposed method, the number of cities was significantly increased. The previously used list of twelve cities was expanded to 48. A list of new cities is presented in Table \ref{tab:48-cities}. To obtain GTFS files for those cities, \textit{Open Mobility Data}\cite{openmobilitydata} service was used.\footnote{with an exception of Zurich, which was downloaded from \url{https://data.stadt-zuerich.ch/dataset/vbz_fahrplandaten_gtfs} [Accessed: 02.06.2021]} Since, there are many more cities available in this archive, some exclusion criteria were introduced:
\begin{itemize}
    \item city population of at least 200,000 people,
    \item at least 20 routes of public transport,
    \item feed must be complete, meaning no public transport types should be missing,
    \item GTFS file should not be for an entire country (those include intercity trains, which may affect feature engineering process.
\end{itemize}
While processing feeds some issues occurred, which resulted in an additional exclusion criterion - a \textit{trip\_headsign} column in the trips table must be correct, since it is used to calculate a number of directions available from a region. Based on that, some cities were also excluded (for example Paris, Lisbon). 

\begin{table}[h]
\centering
\caption{Extended list of cities}
\label{tab:48-cities}
\begin{tabular}{@{}llllll@{}}
\toprule
Zagrzeb & Helsinki  & Espoo     & Tampere   & Oulu   & Lille     \\
Nice    & Toulouse  & Nantes    & Cologne   & Aachen & Hamburg   \\
Erfurt  & Mannheim  & Antwerp   & Gent      & Wieden & Karlsruhe \\
Ateny   & Budapest  & Wenecja   & Florencja & Genoa  & Turin     \\
Palermo & Milan     & Oslo      & Szczecin  & Lublin & Radom     \\
Belgrad & Valencia  & Zurych    & Dublin    & Bilbao & Madryt    \\
Wilno   & Barcelona & Praga     & Warszawa  & Lipsk  & Wroclaw   \\
Poznan  & Krakow    & Bydgoszcz & Gdansk    & Berlin & Bruksela  \\ \bottomrule
\end{tabular}
\end{table}

\subsection{Clustering results analysis}

The process of identifying typology was simplified compared to one described in Section \ref{sec:typology-identification-methodology}. Since a typology was defined in the previous experiment, this one will focus on extracting differences and verifying whether the proposed typology can be applied to an increased number of cities. This section will discuss the results of this experiment, and compare results to those obtained in a smaller number of cities. 

\begin{figure}[ht]
    \centering
    \includegraphics[width=0.75\textwidth]{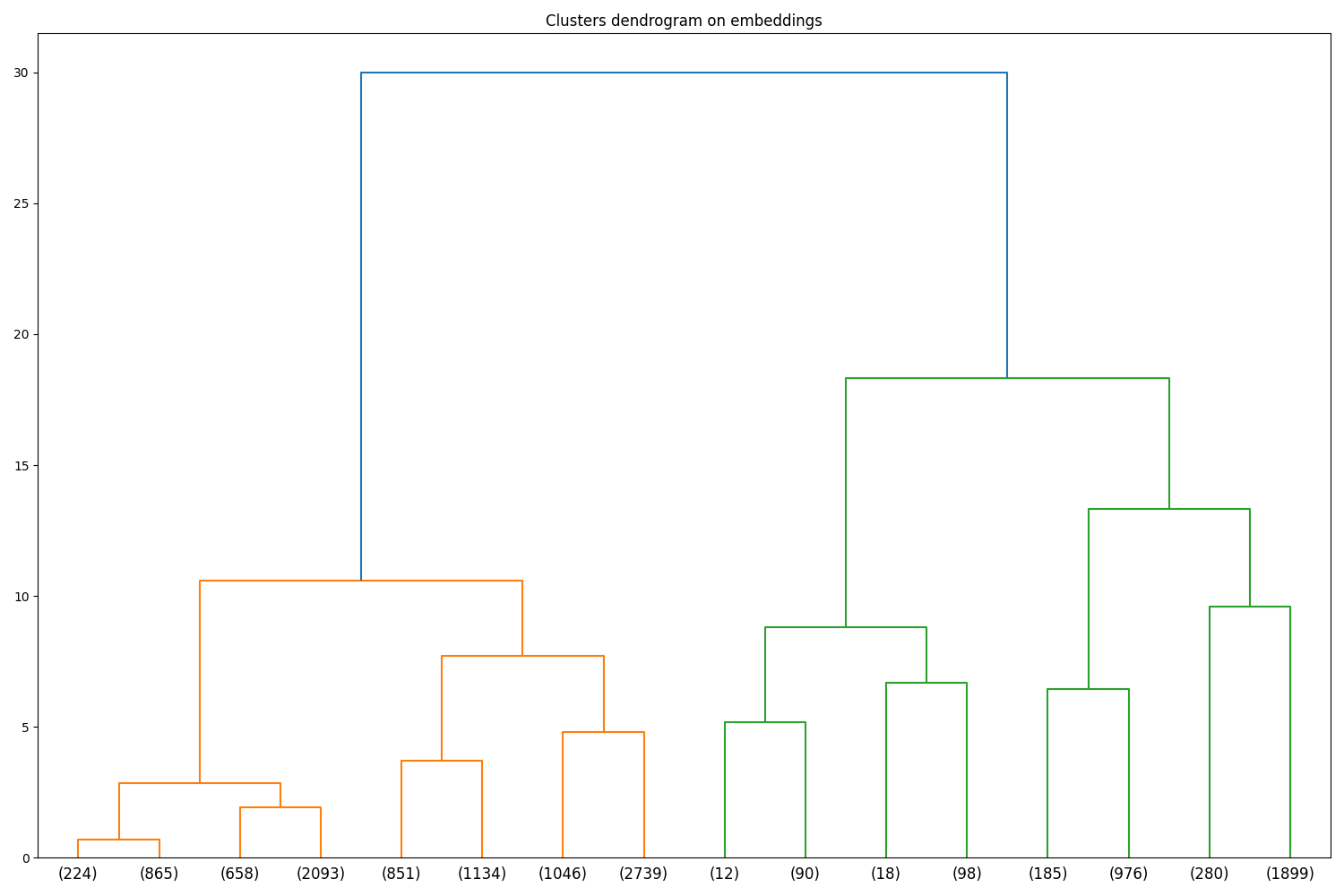}
    \caption{Dendrogram from agglomerative clustering process - 48 cities}
    \label{fig:48:dendrogram}
\end{figure}

In the beginning, a dendrogram was created to analyze how the process of dividing clusters proceeded (see Figure \ref{fig:48:dendrogram}). Two main observations can be made compared to an experiment with twelve cities. Firstly, the distances between clusters are smaller, which suggests that with the addition of cities, the embedding space density of regions increased. Secondly, the left branch is higher than previously, which suggests that the cluster which contains suburban areas is less separated.

\begin{figure}[h]
    \centering
    \begin{subfigure}{0.49\textwidth}
        \centering
        \includegraphics[width=\textwidth]{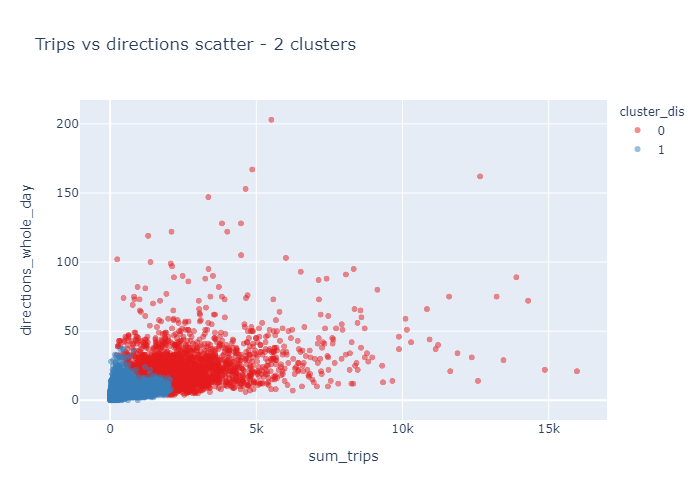}
        \caption{2 clusters}
        \label{fig:48:scatter-2}
    \end{subfigure}
    \begin{subfigure}{0.49\textwidth}
        \centering
        \includegraphics[width=\textwidth]{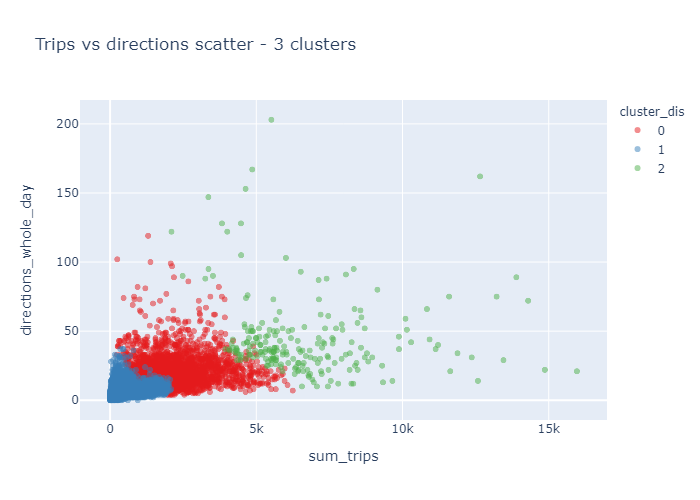}
        \caption{3 clusters}
        \label{fig:48:scatter-3}
    \end{subfigure}
    \begin{subfigure}{0.49\textwidth}
        \centering
        \includegraphics[width=\textwidth]{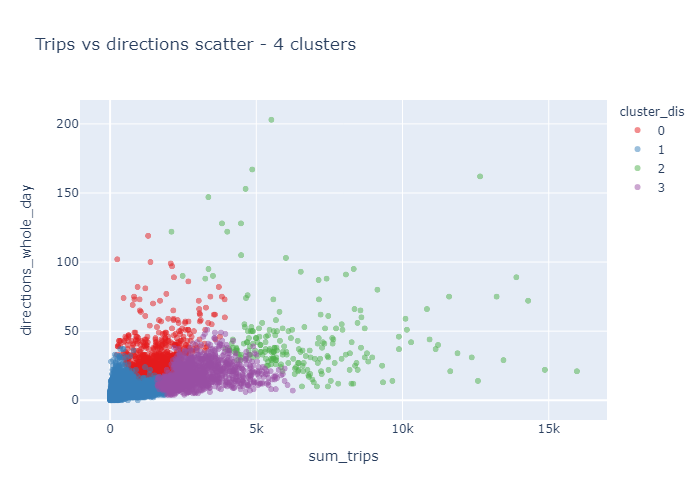}
        \caption{4 clusters}
        \label{fig:48:scatter-4}
    \end{subfigure}
    \begin{subfigure}{0.49\textwidth}
        \centering
        \includegraphics[width=\textwidth]{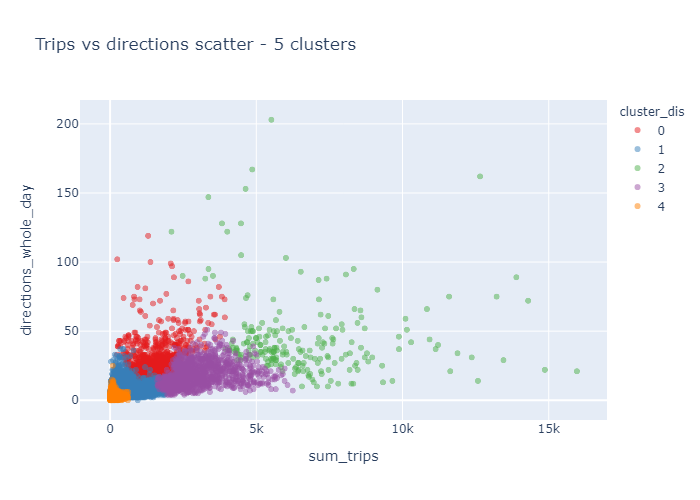}
        \caption{5 clusters}
        \label{fig:48:scatter-5}
    \end{subfigure}
    \caption{Scatter plots for aggregated features - division for 2-5 clusters}
    \label{fig:48:scatter-2-5}
\end{figure}

\begin{figure}[t]
    \centering
    \begin{subfigure}{0.49\textwidth}
        \centering
        \includegraphics[width=\textwidth]{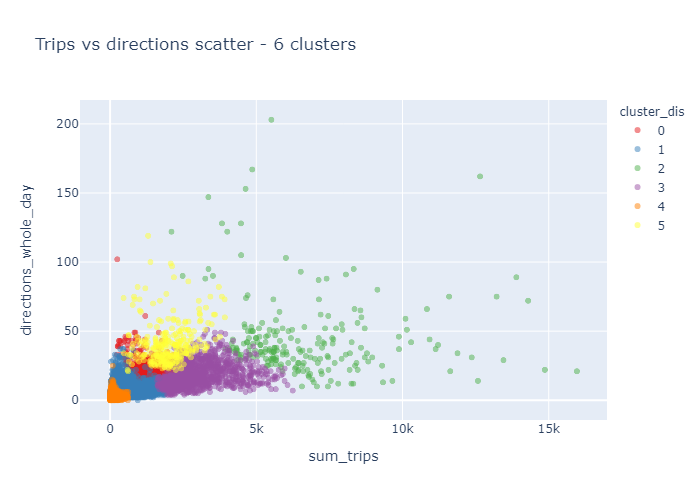}
        \caption{6 clusters}
        \label{fig:48:scatter-6}
    \end{subfigure}
    \begin{subfigure}{0.49\textwidth}
        \centering
        \includegraphics[width=\textwidth]{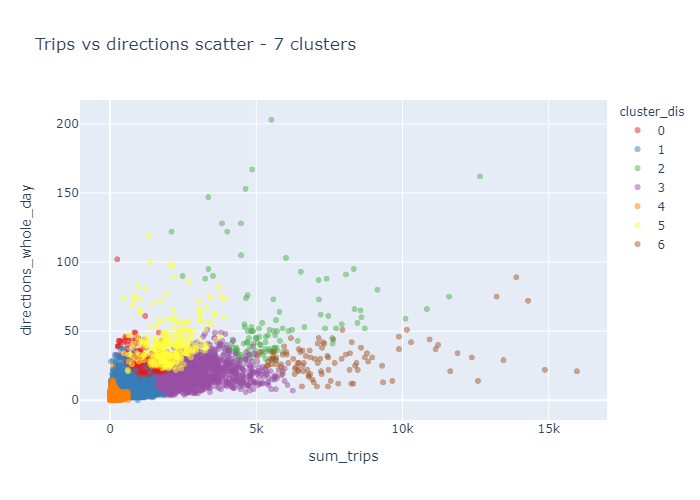}
        \caption{7 clusters}
        \label{fig:48:scatter-7}
    \end{subfigure}
    \begin{subfigure}{0.49\textwidth}
        \centering
        \includegraphics[width=\textwidth]{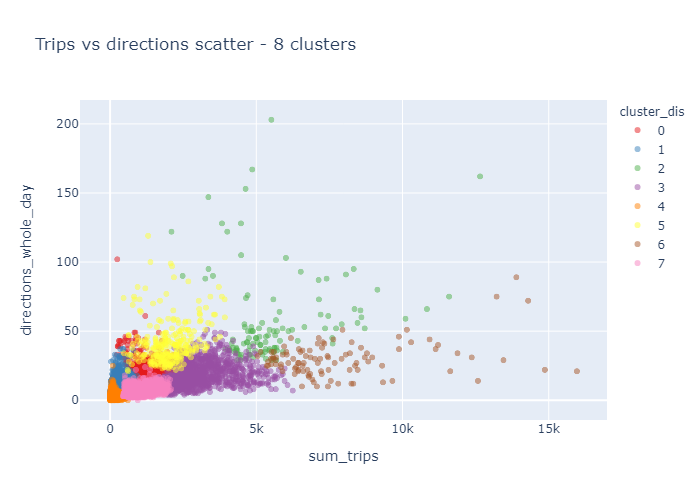}
        \caption{8 clusters}
        \label{fig:48:scatter-8}
    \end{subfigure}
    \begin{subfigure}{0.49\textwidth}
        \centering
        \includegraphics[width=\textwidth]{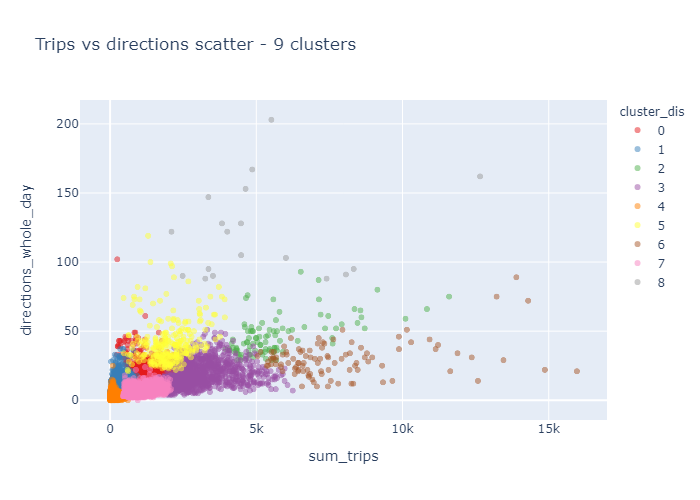}
        \caption{9 clusters}
        \label{fig:48:scatter-9}
    \end{subfigure}
    \caption{Scatter plots for aggregated features - division for 6-9 clusters}
    \label{fig:48:scatter-6-9}
\end{figure}

The second step was to analyze results of clustering on different levels, in search of a number of clusters that defines a logical typology. To achieve that, a top-down approach was used. To visualize clusters, scatter plots of aggregated features (trips and directions from an entire day) were prepared. They are presented in Figures \ref{fig:48:scatter-2-5} and \ref{fig:48:scatter-6-9}. 

\paragraph{2 clusters}

The first step was to divide all regions into two clusters, which is illustrated in Figure \ref{fig:48:scatter-2}. The first group which was separated from the rest is concentrated close to the point $(0, 0)$, which means that there is relatively poor public transport availability comparing to the second cluster. This cluster (\textbf{1}) is also in majority as shown in Table \ref{tab:48:2-clusters:portion}, which is different than in the previous experiment, where it was more balanced. This shows, that previously selected cities were characterized by good availability of public transport. Also, the addition of more cities moved the border between the suburban and central types of public transport availability. 

\begin{table}[h]
\centering
\caption{Percentage of regions in each cluster - division for 2 clusters, 48 cities}
\label{tab:48:2-clusters:portion}
\begin{tabular}{lrr}
\toprule
cluster &     0 &     1 \\
\midrule
\% of regions in cluster & 27.02 & 72.98 \\
\bottomrule
\end{tabular}
\end{table}

\paragraph{3 clusters}

The introduction of the 3rd cluster, similar to the previous experiment, resulted in the separation of regions with a high volume of public transport. A scatter plot presenting this division is in Figure \ref{fig:48:scatter-3}. This cluster contributes to less than 2\% of all regions (see Table \ref{tab:48:3-clusters:portion}), meaning that with an addition of cities with a region of very high public transport availability this cluster was decreased compared to the previous experiment with twelve cities.

\begin{table}[h]
\centering
\caption{Percentage of regions in each cluster - division for 3 clusters, 48 cities}
\label{tab:48:3-clusters:portion}
\begin{tabular}{lrrr}
\toprule
cluster &     0 &     1 &    2 \\
\midrule
\% of regions in cluster & 25.36 & 72.98 & 1.66 \\
\bottomrule
\end{tabular}
\end{table}

\paragraph{4 clusters}

The addition of the 4th cluster resulted in a division in the area of regions with medium public transport availability (see Figure \ref{fig:48:scatter-4}). Previously, this separation was made concerning both types of features (quantity and variety related). This time cluster \textbf{0} contains regions with a bigger variety, while cluster \textbf{3} groups regions with increased availability in terms of the number of trips. Table \ref{tab:48:4-clusters:portion} presents how many regions fall into each cluster. It shows that about a third of all medium regions fell into a cluster with a high number of trips. 

This number of clusters provides clear explanations for all groups and differentiates them significantly. Therefore, division for four clusters will be included in the typology definition. Four types of regions can be described as:

\begin{itemize}
    \item suburban areas,
    \item hubs,
    \item branching regions in the public transport network and smaller hubs,
    \item regions with high throughput and a limited number of directions (like metro stations or main streets).
\end{itemize}

\begin{table}[ht]
\centering
\caption{Percentage of regions in each cluster - division for 4 clusters, 48 cities}
\label{tab:48:4-clusters:portion}
\begin{tabular}{lrrrr}
\toprule
cluster &     0 &     1 &    2 &    3 \\
\midrule
\% of regions in cluster & 16.55 & 72.98 & 1.66 & 8.82 \\
\bottomrule
\end{tabular}
\end{table}

\paragraph{5 clusters}

The fifth cluster split a cluster with the lowest public transport availability concerning both numbers of trips and directions. The cluster which contained all of those was the biggest one, so division here is logical. A visualization of this division is presented in Figure \ref{fig:48:scatter-5}, and number of regions in each cluster is summarized in Table \ref{tab:48:5-clusters:portion}.

An analysis of per hour values in both types of features suggests that cluster \textbf{1} has higher values of both directions and trips. There is also a higher variance in those features. The number of trips in cluster \textbf{1} shows an M-shape, which is typical to public transport which has a higher intensity in morning and afternoon rush hour (forming a shape of letter M when plotting the number of trips). The plot in cluster \textbf{4} remains flat, which suggests that there is too little public transport to diversify between rush hours and regular time of day.

This separation is similar to the one which occurred when introducing the 7th cluster in the previous experiment. This shows, that the character of extracted types is similar, only relative distances changed resulting in changed order of types separation.

\begin{table}[h]
\centering
\caption{Percentage of regions in each cluster - division for 5 clusters, 48 cities}
\label{tab:48:5-clusters:portion}
\begin{tabular}{lrrrrr}
\toprule
cluster &     0 &     1 &    2 &    3 &     4 \\
\midrule
\% of regions in cluster & 16.55 & 43.82 & 1.66 & 8.82 & 29.16 \\
\bottomrule
\end{tabular}
\end{table}

\paragraph{6 clusters}

At the level of 6 clusters, there was a division in cluster \textbf{0}. A part with an increased number of both trips and directions was separated resulting in decreased variance on both feature types. Primarily, the regions with a higher number of directions, comparable with regions classified as transportation hubs, were separated. Those could work as transfer points in regions further from the city center or as main hubs in cities with the less developed public transport network. 

\begin{table}[h]
\centering
\caption{Percentage of regions in each cluster - division for 6 clusters, 48 cities}
\label{tab:48:6-clusters:portion}
\begin{tabular}{lrrrrrr}
\toprule
cluster &     0 &     1 &    2 &    3 &     4 &    5 \\
\midrule
\% of regions in cluster & 14.42 & 43.82 & 1.66 & 8.82 & 29.16 & 2.13 \\
\bottomrule
\end{tabular}
\end{table}

\paragraph{7 clusters}

Moving down in the dendrogram, the 7th cluster was separated from cluster \textbf{2} containing transportation hubs. A division was made based on a number of available directions, forming cluster \textbf{6} with a very high number of trips and a relatively small number of available directions. This can be seen in Figure \ref{fig:48:scatter-7}. This character seems to fit areas with high-frequency metro lines. Another reason may be a high density of stops resulting in many options to use public transport. 
This type of public transport offer was not separated when experimenting with twelve cities. This suggests that expanding the number of considered cities revealed types that were not present in the previously selected subset.

\begin{table}[h]
\centering
\caption{Percentage of regions in each cluster - division for 7 clusters, 48 cities}
\label{tab:48:7-clusters:portion}
\begin{tabular}{lrrrrrrr}
\toprule
cluster &     0 &     1 &    2 &    3 &     4 &    5 &    6 \\
\midrule
\% of regions in cluster & 14.42 & 43.82 & 0.88 & 8.82 & 29.16 & 2.13 & 0.77 \\
\bottomrule
\end{tabular}
\end{table}

\paragraph{8 clusters}

Introduction of 8th cluster resulted in another division in suburban type of public transport. Now, there are three types there (this was visualized in Figure \ref{fig:48:scatter-8}):
\begin{itemize}
    \item cluster \textbf{4} - low availability of public transport,
    \item cluster \textbf{1} - high variety of public transport,
    \item cluster \textbf{7} - good public transport availability, but with limited options.
\end{itemize}

\begin{table}[h]
\centering
\caption{Percentage of regions in each cluster - division for 8 clusters, 48 cities}
\label{tab:48:8-clusters:portion}
\begin{tabular}{lrrrrrrrr}
\toprule
cluster &     0 &     1 &    2 &    3 &     4 &    5 &    6 &     7 \\
\midrule
\% of regions in cluster & 14.42 & 28.74 & 0.88 & 8.82 & 29.16 & 2.13 & 0.77 & 15.07 \\
\bottomrule
\end{tabular}
\end{table}

\paragraph{9 clusters}

The 9th cluster is again separated from the cluster with transportation hubs. This time, regions with very high diversity of available directions are grouped in cluster \textbf{8} (see Figure \ref{fig:48:scatter-9}). This character of public transport availability matches to train stations, which have a high diversity of available directions but lower frequency of trips. As shown in Table \ref{tab:48:9-clusters:portion}, this cluster consists of very few regions. 

This cluster separation is similar to the last step of analysis in the previous experiment, which resulted in the separation of regions with a railway-like type of public transport offer. This was the last type from the previously selected typology, so division for 9 types will be discussed as an updated level 3 in the typology description.

\begin{table}[h]
\centering
\caption{Percentage of regions in each cluster - division for 9 clusters, 48 cities}
\label{tab:48:9-clusters:portion}
\begin{tabular}{lrrrrrrrrr}
\toprule
cluster &     0 &     1 &    2 &    3 &     4 &    5 &    6 &     7 &    8 \\
\midrule
\% of regions in cluster & 14.42 & 28.74 & 0.74 & 8.82 & 29.16 & 2.13 & 0.77 & 15.07 & 0.14 \\
\bottomrule
\end{tabular}
\end{table}

\subsection{Typology description}

This part will discuss the results of clustering analysis in the search for changes in previously defined typology. When possible, previous numbers of clusters will be upheld in pursuit to show, that the typology has not changed significantly and has been well defined using fewer cities.

\paragraph{Level 1:}

\begin{figure}[t]
    \centering
    \begin{subfigure}{0.32\textwidth}
        \centering
        \includegraphics[width=\textwidth]{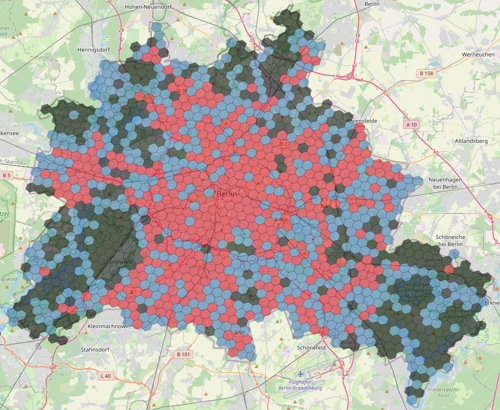}
        \caption{Berlin - 12 cities}
    \end{subfigure}
    \begin{subfigure}{0.34\textwidth}
        \centering
        \includegraphics[width=\textwidth]{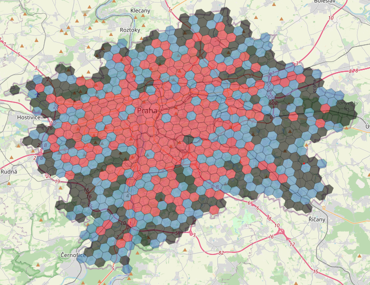}
        \caption{Prague - 12 cities}
    \end{subfigure}
    \begin{subfigure}{0.27\textwidth}
        \centering
        \includegraphics[width=\textwidth]{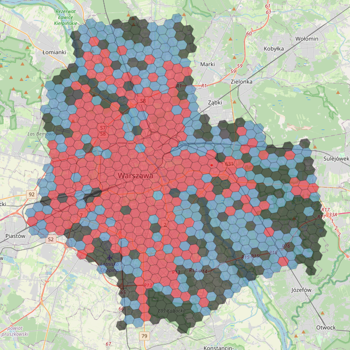}
        \caption{Warsaw - 12 cities}
    \end{subfigure}
    \begin{subfigure}{0.32\textwidth}
        \centering
        \includegraphics[width=\textwidth]{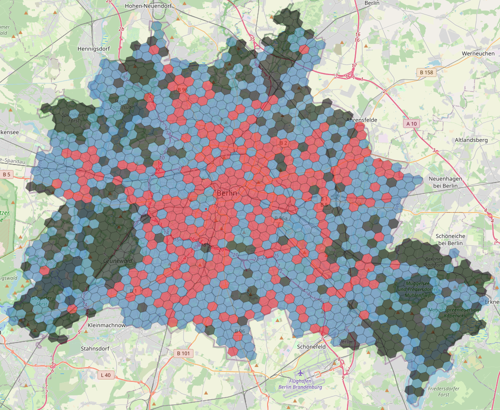}
        \caption{Berlin - 48 cities}
    \end{subfigure}
    \begin{subfigure}{0.34\textwidth}
        \centering
        \includegraphics[width=\textwidth]{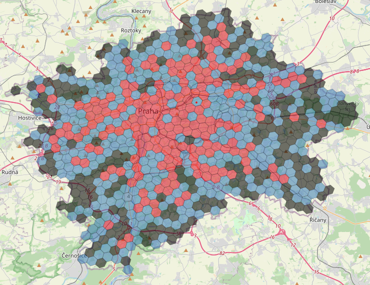}
        \caption{Prague - 48 cities}
    \end{subfigure}
    \begin{subfigure}{0.27\textwidth}
        \centering
        \includegraphics[width=\textwidth]{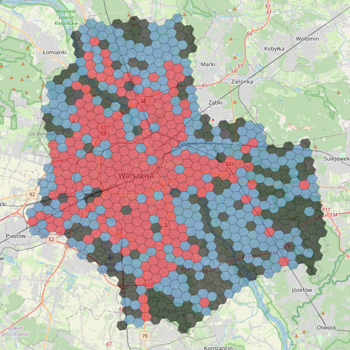}
        \caption{Warsaw - 48 cities}
    \end{subfigure}
    \caption{Difference on level 1 of typology after cities addition}
    \label{fig:48:2-typology:changes}
\end{figure}

The first level of typology was defined as a split between areas with the suburban type of traffic and the city center. With the addition of more cities, this split was upheld at this level. However, since with the addition of cities with a high frequency of public transport and diversity of public transport the boundary between them was moved higher. Therefore, for the cities which were included in the previous experiment, a drop in the number of regions in a city-center-type cluster was observed. It was illustrated in Figure \ref{fig:48:2-typology:changes}. This drop was more visible in cities with a very high share of this type of region (like Brussels in the previous experiments) than in cities that already had this share quite low (eg. Vilnius).

\paragraph{Level 2:}

The second level of typology is very similar compared to the experiment with 12 cities. The only difference is that those mid-city types (red and purple in this case) are separated differently. Previously, the general availability of public transport was a difference between those types. With an expanded list of cities used in this experiment, this division is between regions with a high number of trips and those with a high number of directions. It must be noted, that cluster with more directions (red) has values compared to the ones which would classify a region to hubs in the previous experiment. This is another evidence, that previously selected cities did not contain regions with such high public transport availability as some of those added. It does not mean, that previously selected cities had generally worse public transport offers because the share of regions in the type of a hub is significantly lower. It only means that there are regions with extreme values which were not present before. 

\begin{figure}[h]
    \centering
    \begin{subfigure}{0.37\textwidth}
        \centering
        \includegraphics[width=\textwidth]{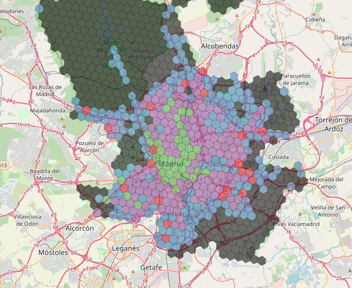}
        \caption{Madrid}
        \label{fig:48:4-typology:madryt}
    \end{subfigure}
    \begin{subfigure}{0.4\textwidth}
        \centering
        \includegraphics[width=\textwidth]{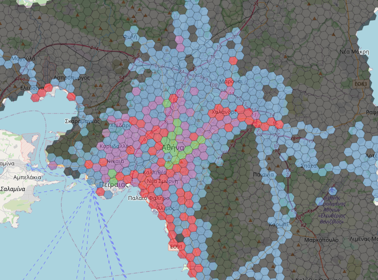}
        \caption{Athens}
        \label{fig:48:4-typology:ateny}
    \end{subfigure}
    \caption{Example of areas with better public transport availability in new cities}
    \label{fig:48:4-typology:examples}
\end{figure}

An example of this can be seen in Madrid and Athens (Figure \ref{fig:48:4-typology:examples}), where the majority of regions fall in green and purple types, which are those with a high number of trips each hour. In Madrid, this is very clearly visible, which may be a result of dense development and numerous metro lines in the city. 

This level of typology is still capable of very efficient identification of regions with better public transport availability in the cities. An example can be seen in Athens (Figure \ref{fig:48:4-typology:ateny}), where a tramline along the coast differentiates those regions from other parts of the city. Moreover, a hub type in Piraeus is probably a combination of metro and tram lines starting in close proximity. 

\paragraph{Level 3:}

The last level of typology can be based on the division for \textbf{9} clusters, which is more than in the previous experiment. This was caused by the fact, that the cluster with a railway-like type of transportation hubs was separated as 9th. It was very important in typology, therefore it was extended to 9 types. Obtained types are shortly characterized below. 

\begin{itemize}
    \item suburban areas
    
    \begin{itemize}
        \item cluster \textbf{1} - parts of suburbs with better variety of public transport - some sort of entry points for suburbs
        \item cluster \textbf{4} - regions with the least amount of public transport available
        \item cluster \textbf{7} - regions in suburbs with more trips but with small variety of public transport
    \end{itemize}
    
    \item mid-city 
    
    \begin{itemize}
        \item cluster \textbf{0} - mid-city regions with regular public transport availability
        \item cluster \textbf{3} - mid-city regions with a lot of trips every hour on limited number of directions
        \item cluster \textbf{5} - regions with very good public transport availability, in cities with worse public transport network they may serve as hubs, in other just as regional hubs
    \end{itemize}
    
    \item hubs
        
    \begin{itemize}
        \item cluster \textbf{2} - very high intensity hubs
        \item cluster \textbf{6} - regions with a lot of trips, but limited directions, most often regions with metro network in a city center
        \item cluster \textbf{8} - railway-type hubs with a lot of diversity and less trips per hour
    \end{itemize}
\end{itemize}

The suburban areas contain regions with high diversity and high quantity of available public transport (respectively clusters \textbf{1} and \textbf{7}). They can be treated as local transfer points in an area with poorer public transport availability and well-connected parts of suburbs (for example just one bus line but with a high frequency of trips). The remaining regions from suburban-type have significantly worse public transport availability. 

The mid-city type regions with a high frequency of public transport but with limited variety were separated (cluster \textbf{3}). Those are regions with very good availability of public transport. The second type separated regions with a relatively big variety of public transport (cluster \textbf{5}). Those can serve as transportation hubs and main transfer points in cities with worse quality of public transport network or as local hubs in districts. The remaining cluster includes regions with regular availability of public transport.

Hubs were separated differently. One of the clusters shows significantly better public transport availability (cluster \textbf{2}). It represents hubs with the highest number of trips and directions. Two remaining clusters maintain only one of them on a high level - either number of trips (cluster \textbf{6}) or a number of directions (cluster \textbf{8}). The first one would mean regions with a dense metro network. The other contains quite often train stations. 

\begin{figure}[h]
    \centering
    \includegraphics[width=0.8\textwidth]{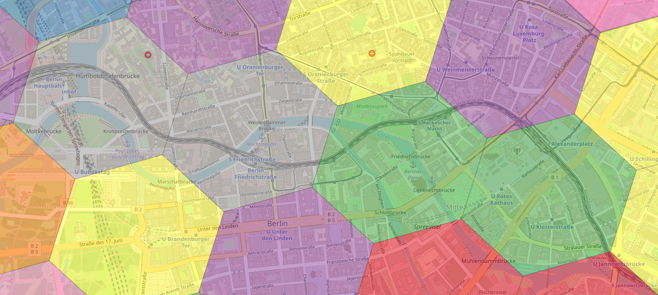}
    \caption{Examples different types of hubs in Berlin}
    \label{fig:48:9-typology:example-berlin}
\end{figure}

An example of that is presented in Figure \ref{fig:48:9-typology:example-berlin}. In gray cluster \textbf{8} there is the main railway station and in green one (cluster \textbf{2}) there is one of the main metro stations \textit{Alexanderplatz}.

\begin{figure}[h]
    \centering
    \includegraphics[width=0.8\textwidth]{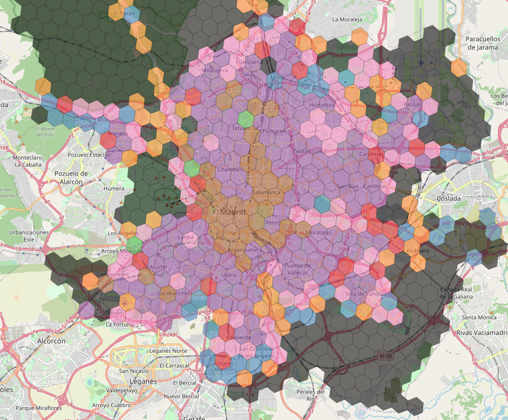}
    \caption{Very high availability of public transport in Madrid}
    \label{fig:48:9-typology:example-madrid}
\end{figure}

This typology allowed to identify a city with a very high availability of public transport - Madrid - which was presented in Figure \ref{fig:48:9-typology:example-madrid}. Most of the regions in this city fall into either cluster \textbf{7}, \textbf{3} or \textbf{6}, which all group regions with a high number of trips per hour. This is possibly a result of high urban density, which resulted in a lot of public transport stops in a small area.

\subsection{Research questions - answers}

\paragraph{RQ6}: Will the character of differences change with the addition of more cities?

Yes, the addition of more cities has resulted in the emergence of regions with greater public transport accessibility which has shifted the boundary between types. Also, the nature of some of the divisions has changed as exemplified by level 2 typologies where one type has many directions and the other has a high number of trips.

\paragraph{RQ7}: Will the typology be applicable for more cities?

Yes, although the criteria for belonging to each type have changed they have remained the same as for fewer cities. For example, there are still separate regions of suburban character. Additionally, hub regions are still divided into the same three types. 

\paragraph{RQ8}: What types of public transport offer was not present in previously selected 12 cities?

One change that was noticed when the list of cities was expanded was the appearance of a large number of regions with a high number of connections but a low number of destinations. An example is the regions from Madrid, where most of the hexagons were classified as these types. In addition, the suburban area was divided into three types separating areas with a high number of connections and a low variety of directions. The most likely reason for the appearance of such areas is the increase in the proportion of cities with metro networks in the dataset.

\chapter{Summary and conclusions}

This thesis had two main objectives. Firstly, propose an unsupervised embedding method to learn a representation of micro-regions in the city, which would make information about public transport offer useful. Secondly, perform an exploratory analysis of such obtained representation space to identify similarities between regions and define a typology of types of public transport offers. Both of them were successfully made. In order to achieve this, several sub-tasks were identified and solved.

The first step was to perform a literature review on representation learning for spatial data and utilization of public transport schedules. The representation learning approach was identified to be very effective when applied to tasks related to spatial data. An interesting research gap was identified in terms of using public transport schedules in this domain of problems. 
Moreover, the usage of public transport data was examined in the reviewed literature. Most of the works focus on a different problem - prediction of delays but papers that tackle a problem of public transport network analysis do not use representation learning. 

In the next step, a feature engineering process was carried out which involved defining variety and quantity features of public transport availability. Moreover, Uber's H3 spatial index was proposed as a method for dividing cities into micro-regions. Finally, normalization approaches were discussed as normalization was highly recommended.

Having defined a set of features for a given micro-region, an embedding method was proposed. To the best of author's knowledge, this is the first method of embedding micro-regions that utilizes static information from public transport timetables. It uses deep neural networks and is based on an autoencoder architecture. 

Two datasets were collected to test the proposed method. The first consisted of twelve European cities, which were selected based on their diverse characteristics. This was used to perform an in-depth analysis of the obtained representation space. The second dataset consisted of 48 cities and was selected with a couple of exclusion criteria based on city size and GTFS feed availability. This dataset was used to verify the scalability of the proposed solution and its ability to generate a general typology by comparing results on both datasets.

All cities were divided, as described above, into micro-regions, which were embedded using the proposed method. The exploratory analysis was based on a hierarchical clustering approach. Multiple numbers of clusters were tested to define a typology of types of public transport offer. Based on that, a multi-level typology was defined and successfully analyzed.

Despite the fact, that the proposed solution seems rather simple, usage of a wide selection of cities allowed to obtain a good quality, multi-level typology definition which can be used for comparisons \emph{between} cities and \emph{within} them. Surprisingly several unique functions of regions were identified in terms of their role in the public transport network.

\section{Future works}\label{sec:future}

The proposed solution proved, in the opinion of the author, that public transport timetables are a valuable source of information for representing regions in a city. The full methodology proposed in this thesis is a good start for further development. Based on the results from this thesis, the following areas for future works were identified. 

The first aspect which is worth examining is the usage of local normalization. The overview of this idea was presented in this thesis, but due to this thesis limitations were not tested.  This approach should allow for independence from the scale of available transportation in a city and allow for the identification of regions with similar functions despite different public transport availability. At the cost of being able to compare cities among themselves, this approach should allow comparing the styles of public transport organizations in different cities.

Another area worth analyzing is extending features that describe a region. For example, the introduction of travel times and taking into consideration the possibility of transfers would certainly allow for a better understanding of the public transport network of a city. 

Finally, other approaches to embedding could be tested. The literature review demonstrated that embedding models from other domains work well when applied to spatial data. Therefore, methods originated in the graph embedding domain or language processing could be incorporated into the problem of public transport-based embeddings.

\listoffigures

%\listof{listing}{Listings}

\listoftables

% \nocite{*}

\bibliographystyle{helpers/dyplom}
\bibliography{bibliography}

\begin{thebibliography}{10}

\bibitem{tabnet}
S.~{\"{O}}. Arik, T. Pfister, \emph{Tabnet: Attentive interpretable tabular
  learning}, CoRR. 2019, volume abs/1908.07442.

\bibitem{Barnes2020}
R. Barnes, S. Buthpitiya, J. Cook, A. Fabrikant, A. Tomkins, F. Xu,
  \emph{{BusTr: Predicting Bus Travel Times from Real-Time Traffic}},
  Proceedings of the ACM SIGKDD International Conference on Knowledge Discovery
  and Data Mining. 2020, pages 3243--3251.

\bibitem{cape}
B. Chang, Y. Park, D. Park, S. Kim, J. Kang, \emph{{Content-Aware Hierarchical
  Point-of-Interest Embedding Model for Successive POI Recommendation}}, w:
  \emph{Proceedings of the Twenty-Seventh International Joint Conference on
  Artificial Intelligence}, volume 2018-July (International Joint Conferences
  on Artificial Intelligence Organization, California, 2018), pages 3301--3307.

\bibitem{zone2vec}
J. Du, Y. Chen, Y. Wang, J. Pu, \emph{{Zone2Vec: Distributed Representation
  Learning of Urban Zones}}, w: \emph{2018 24th International Conference on
  Pattern Recognition (ICPR)}, volume 2018-Augus (IEEE, 2018), pages 880--885.

\bibitem{pytorch-lightning}
W. {Falcon et al.}, \emph{Pytorch lightning}, GitHub repository:
  \url{https://github.com/PyTorchLightning/pytorch-lightning}. 2019, volume~3.

\bibitem{FayyazS.2017}
S.~K. {Fayyaz S.}, X.~C. Liu, G. Zhang, \emph{{An efficient General Transit
  Feed Specification (GTFS) enabled algorithm for dynamic transit accessibility
  analysis}}, PLOS ONE. 2017, volume~12, 10, pages e0185333.

\bibitem{VonFerber2008}
C. von Ferber, T. Holovatch, Y. Holovatch, V. Palchykov, \emph{{Public
  transport networks: empirical analysis and modeling}}, European Physical
  Journal B. 2008, volume~68, 2, pages 261--275.

\bibitem{gtfs}
{Google Inc.}, \emph{{Reference | Static Transit | Google Developers}},
  \\\url{https://developers.google.com/transit/gtfs/reference}. Accessed:
  2021-04-05.

\bibitem{s2}
{Google Inc.}, \emph{{S2 Geometry}}, \\\url{https://s2geometry.io/}. Accessed:
  2021-04-23.

\bibitem{Kramer1991}
M.~A. Kramer, \emph{{Nonlinear principal component analysis using
  autoassociative neural networks}}, AIChE Journal. 1991, volume~37, 2, pages
  233--243.

\bibitem{doc2vec}
Q. Le, T. Mikolov, \emph{Distributed representations of sentences and
  documents}, w: \emph{Proceedings of the 31st International Conference on
  International Conference on Machine Learning - Volume 32}, ICML'14 (JMLR.org,
  2014), pages II–1188–II–1196.

\bibitem{word2vec}
T. Mikolov, K. Chen, G.~S. Corrado, J. Dean, \emph{Efficient estimation of word
  representations in vector space}. 2013.

\bibitem{openmobilitydata}
{MobilityData IO}, \emph{{Open Mobility Data}},
  \url{https://openmobilitydata.org/}. \\Accessed: 2021-04-16.

\bibitem{ward}
F. Murtagh, P. Legendre, \emph{{Ward's Hierarchical Agglomerative Clustering
  Method: Which Algorithms Implement Ward's Criterion?}}, Journal of
  Classification. 2014, volume~31, 3, pages 274--295.

\bibitem{pytorch}
A. Paszke, S. Gross, F. Massa, A. Lerer, J. Bradbury, G. Chanan, T. Killeen, Z.
  Lin, N. Gimelshein, L. Antiga, A. Desmaison, A. Kopf, E. Yang, Z. DeVito, M.
  Raison, A. Tejani, S. Chilamkurthy, B. Steiner, L. Fang, J. Bai, S. Chintala,
  \emph{Pytorch: An imperative style, high-performance deep learning library},
  w: \emph{Advances in Neural Information Processing Systems 32}, edited by
  H.~Wallach, H.~Larochelle, A.~Beygelzimer, F.~d\textquotesingle
  Alch\'{e}-Buc, E.~Fox, R.~Garnett (Curran Associates, Inc., 2019), pages
  8024--8035.

\bibitem{scikit-learn}
F. Pedregosa, G. Varoquaux, A. Gramfort, V. Michel, B. Thirion, O. Grisel, M.
  Blondel, P. Prettenhofer, R. Weiss, V. Dubourg, J. Vanderplas, A. Passos, D.
  Cournapeau, M. Brucher, M. Perrot, E. Duchesnay, \emph{Scikit-learn: Machine
  learning in {P}ython}, Journal of Machine Learning Research. 2011, volume~12,
  pages 2825--2830.

\bibitem{Raghothama2016}
J. Raghothama, V.~M. Shreenath, S. Meijer, \emph{{Analytics on public transport
  delays with spatial big data}}, w: \emph{Proceedings of the 5th ACM
  SIGSPATIAL International Workshop on Analytics for Big Geospatial Data,
  BigSpatial 2016} (Association for Computing Machinery, Inc, New York, New
  York, USA, 2016), pages 28--33.

\bibitem{gtfs-kit}
A. Raichev, \emph{{GTFS Kit}}. 2019. Github repository:
  \url{https://github.com/mrcagney/gtfs_kit}.

\bibitem{Shoman2020}
M. Shoman, A. Aboah, Y. Adu-Gyamfi, \emph{{Deep Learning Framework for
  Predicting Bus Delays on Multiple Routes Using Heterogenous Datasets}},
  Journal of Big Data Analytics in Transportation. 2020, volume~2, 3, pages
  275--290.

\bibitem{loc2Vec}
V. Spruyt, \emph{Loc2vec: Learning location embeddings with triplet-loss
  networks}, \url{https://www.sentiance.com/2018/05/03/venue-mapping/}.
  Accessed: 2021-03-21.

\bibitem{toblers-first-law}
W.~R. Tobler, \emph{A computer movie simulating urban growth in the detroit
  region}, Economic Geography. 1970, volume~46, sup1, pages 234--240.

\bibitem{h3}
{Uber Technologies Inc.}, \emph{{H3: Uber's Hexagonal Hierarchical Spatial
  Index}}, \\\url{https://eng.uber.com/h3/}. Accessed: 2021-04-20.

\bibitem{kbr2018}
{Urząd Miasta Wrocławia}, \emph{{Kompleksowe Badania Ruchu we Wrocławiu
  2018}},
  \\\url{https://www.wroclaw.pl/srodowisko/kompleksowe-badania-ruchu-we-wroclawiu-i-okolicach-kbr}.
  Accessed: 2021-04-23.

\bibitem{wiki:spatial-embeddings}
{Wikipedia contributors}, \emph{Spatial embedding --- {Wikipedia}{,} the free
  encyclopedia}. 2021. Accessed: 2021-06-06.

\bibitem{hrnr}
N. Wu, X.~W. Zhao, J. Wang, D. Pan, \emph{{Learning Effective Road Network
  Representation with Hierarchical Graph Neural Networks}}, w:
  \emph{Proceedings of the ACM SIGKDD International Conference on Knowledge
  Discovery and Data Mining} (Association for Computing Machinery, 2020), pages
  6--14.

\bibitem{venue2vec}
S. Xu, J. Cao, P. Legg, B. Liu, S. Li, \emph{{Venue2Vec: An Efficient Embedding
  Model for Fine-Grained User Location Prediction in Geo-Social Networks}},
  IEEE Systems Journal. 2020, volume~14, 2, pages 1740--1751.

\bibitem{ze-mob}
Z. Yao, Y. Fu, B. Liu, W. Hu, H. Xiong, \emph{{Representing Urban Functions
  through Zone Embedding with Human Mobility Patterns}}, w: \emph{Proceedings
  of the Twenty-Seventh International Joint Conference on Artificial
  Intelligence}, volume 2018-July (International Joint Conferences on
  Artificial Intelligence Organization, California, 2018), pages 3919--3925.

\end{thebibliography}

\end{document}